\documentclass[10pt,twocolumn,letterpaper]{article}

\usepackage{cvpr}              
\definecolor{cvprblue}{rgb}{0.21,0.49,0.74}
\usepackage[pagebackref,breaklinks,colorlinks,allcolors=cvprblue]{hyperref}

\usepackage[export]{adjustbox}
\usepackage{tcolorbox}
\tcbuselibrary{listings, breakable}

\usepackage{comment}

\usepackage{amsmath}
\usepackage{amssymb}
\usepackage{mathtools}
\usepackage{amsthm}
\usepackage{setspace}
\usepackage{tensor}
\usepackage{pifont}

\usepackage[capitalize,noabbrev]{cleveref}
\usepackage{multirow}

\theoremstyle{plain}
\newtheorem{theorem}{Theorem}
\newtheorem{proposition}{Proposition}
\theoremstyle{definition}
\newtheorem{assumption}{Assumption}
\theoremstyle{definition}
\newtheorem{remark}{Remark}

\usepackage{amsmath,amsfonts,bm}

\def\eqref#1{equation~\ref{#1}}

\def\1{\bm{1}}

\def\rve{{\mathbf{e}}}

\def\rvp{{\mathbf{p}}}

\def\rvx{{\mathbf{x}}}
\def\rvy{{\mathbf{y}}}
\def\rvz{{\mathbf{z}}}

\def\rmI{{\mathbf{I}}}

\def\rmW{{\mathbf{W}}}
\def\rmX{{\mathbf{X}}}

\def\rmZ{{\mathbf{Z}}}

\DeclareMathAlphabet{\mathsfit}{\encodingdefault}{\sfdefault}{m}{sl}
\SetMathAlphabet{\mathsfit}{bold}{\encodingdefault}{\sfdefault}{bx}{n}

\DeclareMathOperator*{\argmin}{arg\,min}

\usepackage{titletoc}
\usepackage{wrapfig} 
\usepackage{tabularx}
\usepackage{graphicx}
\usepackage{algorithm}
\usepackage{algpseudocode}
\algtext*{EndWhile}
\algtext*{EndIf}
\algtext*{EndFor}
\usepackage{lipsum}
\usepackage[T1]{fontenc}
\makeatletter
\newcommand*{\algrule}[1][\algorithmicindent]{%
  \hspace*{.2em}
  \vrule 
  \hspace*{\dimexpr#1-.2em-.4pt}%
}

\newcommand{\StatePar}[1]{%
  \State\parbox[t]{\dimexpr\linewidth-\ALG@thistlm}{\strut #1\strut}%
}
\renewcommand{\ALG@beginalgorithmic}{\offinterlineskip}

\newcount\ALG@printindent@tempcnta
\def\ALG@printindent{%
  \ifnum \theALG@nested > 0
    \ifx\ALG@text\ALG@x@notext
    \else
      \unskip
      \ALG@printindent@tempcnta=1
      \loop
        \algrule[\csname ALG@ind@\the\ALG@printindent@tempcnta\endcsname]%
        \advance \ALG@printindent@tempcnta 1
        \ifnum \ALG@printindent@tempcnta<\numexpr\theALG@nested+1\relax
      \repeat
        \fi
    \fi
}
\patchcmd{\ALG@doentity}{\noindent\hskip\ALG@tlm}{\ALG@printindent}{}{\errmessage{failed to patch}}
\makeatother

\algrenewcommand\algorithmicend{\strut\textbf{end}}
\algrenewcommand\algorithmicdo{\strut\textbf{do}}
\algrenewcommand\algorithmicwhile{\strut\textbf{while}}
\algrenewcommand\algorithmicfor{\strut\textbf{for}}
\algrenewcommand\algorithmicforall{\strut\textbf{for all}}
\algrenewcommand\algorithmicloop{\strut\textbf{loop}}
\algrenewcommand\algorithmicrepeat{\strut\textbf{repeat}}
\algrenewcommand\algorithmicuntil{\strut\textbf{until}}
\algrenewcommand\algorithmicprocedure{\strut\textbf{procedure}}
\algrenewcommand\algorithmicfunction{\strut\textbf{function}}
\algrenewcommand\algorithmicif{\strut\textbf{if}}
\algrenewcommand\algorithmicthen{\strut\textbf{then}}
\algrenewcommand\algorithmicelse{\strut\textbf{else}}

\algrenewcommand\algorithmicrequire{\strut\textbf{Input:}}
\algrenewcommand\algorithmicensure{\strut\textbf{Output:}}

\let\oldState\State
\renewcommand{\State}{\oldState\strut}

\def\paperID{*****} 
\def\confName{CVPR}
\def\confYear{2026}

\title{Efficient Weighted Sampling via Score-based Generative Models}

\author{Heasung Kim\thanks{Equal contribution. \\Source code: \href{https://github.com/Heasung-Kim/efficient-weighted-sampling-via-score-based-generative-models}{https://github.com/Heasung-Kim/efficient-weighted-sampling-via-score-based-generative-models}}, Taekyun Lee\footnotemark[1], Hyeji Kim, and Gustavo de Veciana\\
The University of Texas at Austin\\
Austin, TX 78712\\
{\tt\small \{heasung.kim, taekyun, hyeji, deveciana\}@utexas.edu}
}

\begin{document}
\maketitle

\begin{abstract}
    Weighted sampling—sampling from a probability density function (PDF) proportional to the product of a base PDF and a weight function—is a fundamental technique with wide-ranging applications in variance reduction, biased sampling, data augmentation, and more. Leveraging the increasing availability of pretrained score-based generative models (SGMs), we propose a training-free weighted sampling framework that approximates the backward diffusion process of the target distribution by augmenting the pretrained base score function with an auxiliary guidance term, in a principled and computationally efficient manner. Our approach builds on two key components: a lightweight approximation of the guidance that avoids costly higher-order derivatives of both the score and weight functions, and an uncertainty-aware scheduler that dynamically adjusts the guidance strength based on a temporal analysis of approximation error. Together, these components enable accurate and stable sampling without relying on particle-based resampling or Hessian evaluations commonly required by existing methods. We validate the effectiveness of our method from synthetic to large-scale settings such as Stable Diffusion XL, where our framework achieves $1.2\times$ to $4.7\times$ speedups while consistently matching or outperforming state-of-the-art baselines in task performance. These results position our method as a scalable and inference-efficient solution for task-adaptive, time-sensitive sampling in generative applications.
\end{abstract}


\section{Introduction}

Score-based Generative Models (SGMs) have demonstrated the ability to sample from high-dimensional distributions~\cite{ho2020ddpm, song2021ddim, song2021score}, with applications spanning image~\cite{dhariwal2021diffusion, pmlr-v139-ramesh21a}, audio~\cite{kong2021diffwave, chen2021wavegrad}, and industrial environment modeling~\cite{lee2024generating, zhang2024denoising}.
In many practical settings, weighted sampling of the learned distribution of SGMs—achieved by reweighting the density function using a weight function—is essential for a variety of objectives, including variance reduction in estimation, data augmentation, selective feature analysis, reward-based sampling, and addressing bias and fairness considerations~\cite{robert2004monte, byrd2019effect, cortes2010learning, kamiran2012data}.

Formally, the weighted sampling problem is defined as follows: consider a Probability Density Function (PDF) $p$ over the domain $\mathbb{R}^{d}$ of the random vector $\rmX^{p}$, and weight function $w: \mathbb{R}^{d} \mapsto \mathbb{R}_{+}$ where $\mathbb{R}_{+}$ denotes the set of positive real values. The concept of weighted sampling entails drawing samples from a modified PDF $q$, whose PDF is proportional to the product of the weight function $w$ and the original PDF $p$ as $q(\rvx) = \frac{ w(\rvx) p(\rvx) }{\int w(\rvx) p(\rvx) \,\mathrm{d}\rvx}$. 

While supporting a wide range of task-specific functions $w(\rvx)$ is desirable, retraining a large-scale generative model for each target distribution $q$, or fine-tuning it accordingly, is computationally infeasible in practice. To mitigate this challenge, recent efforts have explored leveraging the pre-trained score function of an existing SGM to approximate the generation process under $q(\rvx)$---a technique broadly referred to as \emph{guidance}~\cite{yu2023freedom, song2023pseudoinverse, chungdiffusion}. This approach circumvents the need for costly task-specific retraining, making it attractive for real-world deployment.

However, existing guidance-based methods often exhibit limited fidelity in approximating the score function of the reweighted distribution $q$, resulting in degraded sample quality or unstable generation dynamics~\cite{guo2024gradient, phillips2024particle}. To overcome these limitations, state-of-the-art approaches have introduced advanced inference-time mechanisms such as \emph{time-travel resampling} which involves repeatedly resampling intermediate latent states, typically aligned with semantically meaningful generative phases~\cite{yu2023freedom}, or \emph{particle-based resampling schemes}~\cite{phillips2024particle, kim2025das}, which maintain a diverse set of candidate samples and adaptively reweight and resample them during inference. While these techniques can substantially improve sampling fidelity, they introduce additional computational overhead due to increased memory usage and repeated score evaluations, limiting their applicability in computation- and latency-sensitive applications.

In this work, we address two core challenges in SGM-based weighted sampling:  
(i) accurately approximating the score function of the target distribution $q$, and  
(ii) achieving efficient, real-time generation with minimal computational overhead.  
To this end, our method offers both principled theoretical foundations and empirically validated improvements over existing techniques. Our contributions are summarized as follows.

\paragraph{Algorithmic Design and Theoretical Development.}  
We propose a two-stage design for training-free weighted sampling using pretrained SGMs. In the first stage, we derive a lightweight approximation of the guidance term that steers the diffusion trajectory of the base SGM toward that of the target sampling density. This approximation avoids second-order derivatives of both the weight function and the base score function, thereby eliminating the need for costly Hessian computations. 
This is particularly valuable for large pretrained diffusion models, where differentiation is expensive.
In the second stage, we analyze the time-dependent approximation error and introduce a principled scheduler that dynamically adjusts the guidance strength to reduce correction errors.
Together, these components yield efficient and effective sampling.

\paragraph{Empirical Evaluation and Performance Superiority.}  
We conduct comprehensive empirical studies across a range of challenging settings, including sampling in multimodal base distributions, weighting with neural network-based objectives, and high-dimensional generation tasks. The results demonstrate that our method consistently achieves superior task performance while reducing computational costs. 
Notably, we observe up to a \textbf{\emph{4.7$\times$ speedup along with superior performance across diverse metrics and tasks}} compared to existing state-of-the-art methods.

\section{Background on SGMs}
Score-based generative modeling aims to learn the score function $\nabla_{\rvx} \log q(\rvx)$ of a target data distribution using a parameterized model, typically a neural network.  
A key innovation in SGM is to learn the score function of noise-perturbed data rather than the original distribution directly. This enables generative modeling via a backward Stochastic Differential Equation (SDE) and improves training and inference stability~\cite{song2021score}. Consider the following continuous-time diffusion process.
\begin{equation}
\label{true_forward_sde}
    \mathrm{d} \rmX^{q}_{t} = f(\rmX^{q}_{t}, t) \, \mathrm{d}t + \sigma(t) \, \mathrm{d} \rmW_{t},
\end{equation}
where $\rmX^{q}_{0}\sim q$ is the data variable at diffusion time $t=0$, and
$q_t(\rvx)$ denotes the marginal density of $\rmX^{q}_t$ induced by the forward SDE (i.e., $\rmX^{q}_t$ is obtained by evolving $\rmX^{q}_0$ for time $t$ under \eqref{true_forward_sde}).

When the score function $\nabla_{\rvx} \log q_t(\rvx)$ is known, this framework enables sampling of $\rmX^{q}_{0}$ that follows PDF $q_{0} = q$ via a corresponding backward SDE \cite{anderson1982reverse} which is given by 
\begin{align}
\label{true_backward_sde}
&\mathrm{d}\rmX^{q}_t = b_{1}(\rmX^{q}_t, t) \,\mathrm{d}t + \sigma(t)\,\mathrm{d}\tilde{\rmW}_{t}, 
 \quad
\text{where\quad}\\ 
&b_{1}(\rvx, t) = f(\rvx, t) - \sigma(t)^2 \nabla_{\rvx} \log q_{t}(\rvx)
\label{true_backward_sde_2}
\end{align}
with  $\rmX^{q}_{t} \sim q_{t}$ and $\mathrm{d} \tilde{\rmW}_{t}$ denoting the Brownian motion associated with the reverse-time process.

For instance, given an initial sample $\rmX^{q}_{T} \sim q_{T}(\rvx)$, and running the backward SDE (\ref{true_backward_sde}) yields a sample $\rmX^{q}_{0}$ from the distribution $q_0$. Typically, for sufficiently large values of $T$, a realization $\rvx_{T}$ can be initialized as Gaussian noise.
Following convention \cite{ho2020ddpm, lugmayr2022repaint, nichol2021improved, choi2021ilvr}, we consider the following drift and diffusion coefficients: $f(\rvx,t) = -\frac{1}{2} \beta(t) \rvx  \quad \text{and} \quad  \sigma(t) = \sqrt{\beta(t)}$
where $\beta(t)$ is a predefined scalar-valued continuous function such that $0 \leq \beta(t) \leq 1$. Here, in defining $\sigma(t)$ and $f(\rvx,t)$, we apply a slight abuse of notation: the scalar-vector multiplication in this context represents the elementwise multiplication of the vector by the scalar.

In modern practice, many pretrained SGMs are publicly available and provide accurate realizations of the score function of the base density, $\nabla_{\rvx} \log p_t(\rvx)$. Here $p_t(\rvx)$ is the marginal distribution of $\rmX^{p}_{t}$ which is modeled by the same SDE dynamics as in (\ref{true_forward_sde}) but with a different initial condition as
\begin{align}
\label{true_forward_sde_base_density}
    \mathrm{d} \rmX^p_t = -\frac{1}{2} \beta(t) \rmX^p_t\, \mathrm{d}t + \sqrt{\beta(t)}\, \mathrm{d} \rmW_t, \quad \rmX^p_0 \sim p_0.
\end{align}
Throughout this work, we assume access to such a pretrained base SGM, $\nabla_{\rvx} \log p_t(\rvx)$, and focus on efficiently approximating the score function of the weighted sampling density $q_t$, without any additional training or fine-tuning of the given model.

\section{Efficient Weighted Sampling via SGM}
\label{section_importance_sampling_via_score_based_generative_models}

The objective of this work is to construct an efficient and practical approximation of the score function $\nabla_{\rvx} \log q_{t}(\rvx)$ where $q_{0}(\rvx) \propto w(\rvx)p_{0}(\rvx)$, by leveraging only known quantities based on the following decomposition:
\begin{align}
\label{guidance}
    \nabla_{\rvx} \log {q}_{t}(\rvx) &= \nabla_{\rvx} \log {p}_{t}(\rvx) + g(\rvx,t) 
    \\
    &\approx \nabla_{\rvx} \log {p}_{t}(\rvx) + \tilde{g}(\rvx, t) 
\end{align}
where $g(\rvx, t)$ denotes the ground-truth \emph{guidance term}, i.e., the direction required to adjust the base score $\nabla_{\rvx} \log p_t(\rvx)$ to match the score of the weighted sampling density $q_t$. Our goal is to approximate $g(\rvx, t)$ with a computationally lightweight, tractable surrogate $\tilde{g}(\rvx, t)$, thereby obtaining an efficient estimate of the target score function.

Existing methods~\cite{chungdiffusion, yu2023freedom, kim2025das} can be adapted to approximate the guidance term. However, they often rely on computationally intensive procedures such as differentiation of pretrained score models and/or repeated resampling steps. These approaches, while effective, introduce significant runtime overhead—especially when deployed with large-scale SGMs.

\paragraph{Design Overview.} 
To address the limitations of prior approaches, our method consists of two key components:  
(i)~\emph{Lightweight Approximation.} We derive a first-order approximation of the guidance term $g$, avoiding second-order derivatives of both the base density $p_t$ and the weight function $w$. This makes our method well-suited for large-scale pretrained SGMs, where Hessian-based methods are computationally expensive.  
(ii) \emph{Uncertainty-Adaptive Guidance Scheduling.} We analyze the temporal evolution of the approximation uncertainty, quantifying the confidence in the approximation at each diffusion timestep. Based on this analysis, we propose a dynamic scheduler that modulates the strength of the approximated guidance term during sampling. By downweighting uncertain guidance and emphasizing reliable estimates, the scheduler improves the overall quality of the sampled distribution and enhances robustness.

Together, these components form the basis of our proposed method, Lightweight Approximation with uncertainty-adaptive Guidance Scheduling (LAGS), enabling fast and accurate sampling without the need for retraining or fine-tuning.

\subsection{The First-order Approximation of Weighted Sampling Score Function}

\paragraph{Connection to Noise-Perturbed Base and Weighted Sampling Densities.}

We begin by examining the form of the weighted sampling distribution for the noise-perturbed samples, $q_{t}(\rvx)$, under the SDE defined in (\ref{true_forward_sde}).
Consider the transition probability density function $G: \mathbb{R}^{d} \times \mathbb{R}^{d} \times [0,T] \mapsto \mathbb{R}_{+}$, also referred to as the Green’s function or fundamental solution of the SDE \cite{beck1992heat, duffy2015green, economou2006green}. This function represents the probability density of the process transitioning from the given initial state $\rmX^{q}_{0} \! = \! \rvy$ to the state $\rvx$ at time $t$. Specifically, $G(\rvx, \rvy, t)$ denotes the conditional probability density of transitioning from $\rvy$ to $\rvx$ over time $t$. Using this Green’s function, we can express $q_t(\rvx)$ as
\begin{align}
    \vspace{-3mm}
    q_{t}(\rvx) &\!\!= \!\!\int \!\!G(\rvx, \rvy, t) q_{0}(\rvy) \mathrm{d} \rvy \!\!= \!\!
 \frac{1}{Z_0} \!\!\int \!\!G(\rvx, \rvy, t) w(\rvy) p_0(\rvy) \mathrm{d} \rvy \label{q_t_green_function_01}
\end{align}
where $q_{0} = q$, $p_{0} = p$, and $Z_0$ is the normalization constant, $Z_0 = \int  w(\rvx) p(\rvx) \, \mathrm{d} \rvx$.

To further investigate the relationship between $q_{t}$ and $p_{t}$, it should be noted that the corresponding Green’s function for SDE in (\ref{true_forward_sde_base_density}) is also $G$ in (\ref{q_t_green_function_01}), thereby the distribution of $\rmX^{p}$ at time $t$, denoted by $p_{t}$, is given as $p_{t}(\rvx) = \int G(\rvx, \rvy, t) p_{0}(\rvy) \, \mathrm{d} \rvy$. We then have 
\begin{align}
\label{q_t_intermsof_p_t_l_01}
    q_{t}(\rvx) &= \frac{p_{t}(\rvx)}{Z_0} \int w(\rvy) G(\rvx, \rvy, t) \frac{p_0(\rvy)}{p_{t}(\rvx)} \, \mathrm{d} \rvy \nonumber
    \\ &= \frac{p_{t}(\rvx)}{Z_0} \mathbb{E}_{\rmX^{p}_{0} \sim p_{\rmX^{p}_{0}|\rmX^{p}_{t}}(\cdot|\rvx)} [w(\rmX^{p}_{0})] 
\end{align}
where $p_{\rmX^{p}_{0}|\rmX^{p}_{t}}(\cdot|\rvx)$ is the conditional PDF of the initial state $\rmX^{p}_{0}$ for a given state $\rmX^{p}_{t}\!=\!\rvx$ at $t$ under the corresponding SDE of $\rmX^{p}_{t}$. Based on this, we can observe that $q_{t}$ can be represented in terms of $p_{t}$ and $w$ as shown in (\ref{q_t_intermsof_p_t_l_01}), which implies that the score function of the weighted sampling density can be represented as 
\begin{align}
\label{def_true_g}
 \nabla_{\rvx}\!\log q_{t}(\rvx) &= \nabla_{\rvx}\! \log p_{t}(\rvx)  + g(\rvx, t) \text{~~~where~~~}  \nonumber \\ g(\rvx, t) &\coloneqq \nabla_{\rvx}\! \log  \mathbb{E}_{\rmX^{p}_{0} \sim p_{\rmX^{p}_{0}|\rmX^{p}_{t}}(\cdot|\rvx)}\! [w(\rmX^{p}_{0})].
\end{align}

\paragraph{Taylor Approximation.} While Eq.~(\ref{def_true_g}) provides an exact decomposition, the term $g(\rvx, t)$ is generally intractable. To address this, we apply a first-order Taylor expansion of $w(\rmX^p_0)$ about its conditional mean, $\bar{\rvx}'_{0}\vert_{\rvx,t} \coloneqq \mathbb{E}[\rmX^{p}_{0} \vert \rmX^{p}_{t}=\rvx]$, which gives us $w(\rmX^{p}_{0}) \approx  w(\bar{\rvx}'_{0}\vert_{\rvx,t}) + \nabla w(\bar{\rvx}'_{0}\vert_{\rvx,t})^{\top}(\rmX^{p}_{0} - \bar{\rvx}'_{0}\vert_{\rvx,t})$ and
\begin{align}
    \label{score_qt_taylor}
    g(\rvx,t)  &\coloneqq \nabla_{\rvx} \log \mathbb{E}_{\rmX^{p}_{0} \sim p_{\rmX^{p}_{0}|\rmX^{p}_{t}}(\cdot|\rvx)} [w(\rmX^{p}_{0})] \nonumber \\
    &\approx \nabla_{\rvx} \log w(\bar{\rvx}'_{0}\vert_{\rvx,t}).
\end{align}
The conditional mean of the initial state for a given $\rvx$ at $t$ can be obtained through the Tweedie’s approach (\cite{efron2011tweedie, kim2021noise2score,chungdiffusion}) as $\bar{\rvx}'_{0}\vert_{\rvx,t}= \frac{1}{\sqrt{\bar{\alpha}(t)}}(\rvx + (1-\bar{\alpha}(t)) \nabla_{\rvx} \log p_{t}(\rvx))$
where $\bar{\alpha}(t) = \exp( -\int_0^{t} \beta(s) \, ds ).$ 
Importantly, the representation in (\ref{score_qt_taylor}) can be interpreted from a probabilistic perspective by considering an event $E$ and a random variable $Z$ that satisfy $p(Z\in E \mid \rvx) \propto w(\rvx)$. Under this formulation, the approximation in (\ref{score_qt_taylor}) is equivalent to applying the posterior sampling and the denoising Tweedie mean technique \cite{chungdiffusion}. See Appendix \ref{appendix_relation_to_second_order} for details.

\paragraph{Finite-Difference Approximation for Hessian-Vector Multiplication.}
Given that we now have a tractable form for $\nabla_{\rvx} \log w(\bar{\rvx}'_{0}\vert_{\rvx,t})$, we can apply the chain rule to represent $\nabla_{\rvx} \log w(\bar{\rvx}'_{0}\vert_{\rvx,t})$ as follows.
\begin{align}
        \label{true_derivative_log_x_0}
    &\nabla_{\rvx} \log w(\bar{\rvx}'_{0}\vert_{\rvx,t}) \nonumber \\
    &= \frac{\left(  \rmI +  (1-\bar{\alpha}(t)) H_{\log p_{t}}(\rvx) \right)^{\!\top}}{\sqrt{\bar{\alpha}(t)}} \nabla_{\bar{\rvx}'_{0}\vert_{\rvx,t}} \log  w(\bar{\rvx}'_{0}\vert_{\rvx,t})
\end{align}
where $H_{\log p_{t}}(\rvx)$ is the Hessian matrix of $\log p_{t}(\rvx)$, which can be computationally demanding in practice, particularly when the pre-trained score function is implemented as a high-capacity model. Since \(H_{\log p_t}(\rvx)=\nabla_\rvx^2\log p_t(\rvx)\) is symmetric, the transpose can be dropped.
To enhance the  practicality of our approach, we propose approximating the Hessian matrix-vector multiplication in (\ref{true_derivative_log_x_0}) by directional derivative approximation using finite differences with a sufficiently small $\epsilon > 0$ as
\begin{align}
    \label{hessian_approximation_01}
    &\nabla_{\rvx} \log \ p_{t}(\rvx) + \epsilon H_{\log p_{t}}(\rvx)  \nabla_{\bar{\rvx}'_{0}\vert_{\rvx,t}} \log w(\bar{\rvx}'_{0}\vert_{\rvx,t}) \nonumber \\
    &\approx {\nabla_{\rvx} \log p_{t}(\rvx + \epsilon \nabla_{\bar{\rvx}'_{0}\vert_{\rvx,t}} \log w(\bar{\rvx}'_{0}\vert_{\rvx,t})) }.
\end{align}

Substituting (\ref{hessian_approximation_01}) into the score decomposition (\ref{score_qt_taylor}) yields the following first-order approximation of the guidance term $g(\rvx, t) \approx \tilde{g}^{(1)}(\rvx, t)$:
\begin{tcolorbox}[
  colback=gray!5,
  colframe=black!70,
  boxrule=0.3pt,             
  boxsep=1pt,       
  left=0.5pt,         
  right=0.5pt,        
  top=-5.5pt,          
  bottom=2pt,       
  arc=2pt,          
]
\begin{align}
\label{first_order_approximation}
  &\tilde{g}^{(1)}(\rvx,t) \coloneqq \frac{\nabla_{\bar{\rvx}'_{0}\vert_{\rvx,t}} \log w(\bar{\rvx}'_{0}\vert_{\rvx,t})}{\sqrt{\bar{\alpha}(t)}}   + \nonumber \\
  &\frac{\nabla_{\rvx} \log p_{t}(\rvx + \epsilon \nabla_{\bar{\rvx}'_{0}\vert_{\rvx,t}} \log w(\bar{\rvx}'_{0}\vert_{\rvx,t})) - \nabla_{\rvx} \log p_{t}(\rvx)}{\epsilon (1-\bar{\alpha}(t))^{-1} \sqrt{\bar{\alpha}(t)}}.
\end{align}
\end{tcolorbox}

Note that the score approximation in (\ref{first_order_approximation}) does not involve any second-order derivatives of the base density $p$ or the weight function $w$. This property is particularly valuable when performing weighted sampling with large-scale diffusion models, where computing second-order derivatives of the score model is often prohibitive.


\subsection{Approximation Accuracy}
We now analyze the accuracy of the approximation $\tilde{g}^{(1)}$, which later will drive the design of a new approximation $\tilde{g}$, presented in Sec. \ref{section:uncertainty_aware_score}, that incorporates an uncertainty-aware, time-dependent scheduling policy. 
In particular, we establish Theorem \ref{theorem_1} providing an upper bound on the Euclidean norm discrepancy between the approximation and the true guidance, with a detailed proof provided in Appendix \ref{appendix_technical_results}.

\begin{assumption}
    \label{assumption_bounded_norm}
    (Bounded $w$ and its derivatives)
    For all $\rvx \in \mathbb{R}^{d}$, we have $w(\rvx) \ge m$ for some $m>0$ and
    $\Vert \nabla_{\rvx} \log w(\rvx) \Vert \le \eta$.
    Moreover, $w$ is twice differentiable and its Hessian is uniformly bounded as
    $\Vert H_{w}(\rvz) \Vert \le \eta_{2}$ for all $\rvz \in \mathbb{R}^{d}$,
    where $H_w(\rvx)$ is the Hessian matrix of $w(\rvx)$.
\end{assumption}

\noindent Assumption \ref{assumption_bounded_norm} ensures that $w$ is bounded away from zero and its norm of log-gradient and Hessian are uniformly bounded.

\begin{assumption}
    \label{assumption_lipschitz}
    (Bounded log PDF derivatives)
    Suppose that $\mathrm{supp}(p_0)$ is contained in a bounded set, 
    i.e., $\Vert \rvy \Vert \le B$ for all $\rvy \in \mathrm{supp}(p_0)$.
    For all $\rvx \in \mathbb{R}^{d}$, 
    $\rvy \in \mathrm{supp}(p_0)$, and $t\in[0,T]$, we have 
    $\Vert \nabla_{\rvx} \log  p_{\rmX^{p}_{0}|\rmX^{p}_{t}}(\rvy|\rvx) \Vert \le \gamma_{t}$,  
    $\Vert H_{\log p_{t}}(\rvx)\Vert \le \zeta_{t}$, and 
    $\Vert H_{\log p_{t}}(\rvx) - H_{\log p_{t}}(\rvy) \Vert \le L_{t} \Vert \rvx-\rvy\Vert$.
\end{assumption}
\noindent The bounded norm of the score function and Hessian of the log-likelihood 
are standard assumptions in the analysis of SGMs 
\cite{chen2023improved, de2022convergence, chensampling, de2021diffusion}. 
The compact support condition is mild in practice, as data distributions 
such as images are inherently bounded (e.g., pixel values in $[0,1]^d$).
Under the variance preserving (VP) SDE, the conditional score is bounded whenever $\rvy$ lies in a compact set. We additionally assume Lipschitz continuity of the Hessian. 

\begin{theorem}
    \label{theorem_1}
    \text{\normalfont(Approximation gap)}
    Suppose Assumptions~\ref{assumption_bounded_norm}–\ref{assumption_lipschitz} hold.
    Then, for all $\rvx \in \mathbb{R}^d$ and $t \in [0, T]$, and for any $\epsilon > 0$, the deviation between the true guidance term $g(\rvx, t)$ and the approximation $\tilde{g}^{(1)}(\rvx, t)$, denoted by $\Delta^{(1)}(\rvx,t)$, is upper-bounded as follows.
    \begin{align}
    \label{theorem_1_inequality}
   &\Delta^{\!(1)}(\rvx,t) \!=\! g(\rvx, t) \!-\! \tilde{g}^{(1)}(\rvx, t)  \text{~~and~~}   \Vert \Delta^{\!(1)}(\rvx,t) \Vert \!\le\! u(\rvx, t), \nonumber \\
   &\text{where~~~~~} u(\rvx, t) \coloneqq \frac{(1-\bar{\alpha}(t))L_{t}\eta^{2}\epsilon}{2\sqrt{\bar{\alpha}(t)}}  + \big( \frac{\gamma_{t}\eta_{2}}{2m} \nonumber \\
    &+ \frac{ \big({1+(1-\bar{\alpha}(t))\zeta_{t}} \big)\eta\eta_{2} }{2m \sqrt{\bar{\alpha}(t)}} \big) \mathbb{E}\! \left[\Vert \rmX^{p}_{0}- \bar{\rvx}'_{0}\vert_{\rvx,t}\Vert^2 \vert \rmX^{p}_{t}\!=\!\rvx \right].
    \end{align}
\end{theorem}

\begin{remark}[Bounded approximation gap by variance]
\label{remark_1}
The norm discrepancy between the true guidance term $g(\rvx, t)$ and its approximation $\tilde{g}^{(1)}(\rvx, t)$ is bounded in terms of $\bar{\alpha}(t)$ and $\mathbb{E} \left[\Vert \rmX^{p}_{0}- \bar{\rvx}'_{0}\vert_{\rvx,t}\Vert^2 \vert \rmX^{p}_{t}=\rvx \right]$, which is the \emph{trace of the covariance matrix of $\rmX^{p}_{0}$ whose PDF is conditioned on the observation $\rvx$ at $t$}. As $  t\rightarrow 0$, this conditional variance converges to zero, resulting in the second term of $u(\rvx,t)$ vanishing, making the gap negligible. 
Additionally, the first term also vanishes as $\bar{\alpha}(t)\rightarrow 1$, provided $L_t$, $\gamma_t$, $\zeta_t$ remain bounded as $t \rightarrow 0$---a mild condition satisfied by common noise schedules.

\end{remark}

\subsection{Uncertainty-Adaptive Scheduling}
\label{section:uncertainty_aware_score}
Theorem~\ref{theorem_1} and Remark~\ref{remark_1} show that the approximation error decreases as $t \to 0$,
i.e., toward the end of the reverse-time sampling trajectory, which explains why early reverse steps are more unstable.
This behavior aligns with prior observations in various existing methods for image generation, where early-stage instability has led to resampling heuristics or manually tuned guidance strengths \cite{yu2023freedom, kim2025das}. However, such approaches may incur significant computational cost, particularly in high-dimensional settings.

To address this issue systematically, we propose a time-dependent function $\tau : [0, T] \to [0, 1]$ that modulates the contribution of the guidance term  $\tilde{g}^{(1)}$ in the approximated score function as $\nabla_{\rvx} \log q_{t}(\rvx) \approx \nabla_{\rvx} \log p_{t}(\rvx) + \tau(t) \tilde{g}^{(1)}(\rvx, t)$. That is, we consider the form of $ \tilde{g}(\rvx, t)=\tau(t) \tilde{g}^{(1)}(\rvx, t)$. The magnitude of $\tau(t)$ is selected to minimize the mean squared error between the true score function and the approximation
\begin{align}
    \label{risk}
    r(t; \tau) = \mathbb{E}[ \Vert &\nabla_{\rmX^{q}_{t}} \log {q}_{t}(\rmX^{q}_{t}) \nonumber \\ - &\nabla_{\rmX^{q}_{t}} \log p_{t}(\rmX^{q}_{t}) - \tau(t) \tilde{g}^{(1)}(\rmX^{q}_{t}, t) \Vert^2],
\end{align}
where the expectation is taken over $\rmX^q_t \sim q_t$, the state at time $t$ under the weighted sampling diffusion process.
By Theorem~\ref{theorem_1}, the approximation error $\Delta^{(1)}(\rvx,t)$ is controlled by the conditional variance term 
$\mathbb{E}\!\left[\|\rmX^p_{0}-\mathbb{E}[\rmX^p_{0}\mid \rmX^p_{t}]\|^{2}\mid \rmX^p_{t}=\rvx\right]$, 
which admits a deterministic upper bound expressible as a finite sum of $\frac{(1-\bar{\alpha}(t))^{i}}{\bar{\alpha}(t)^{j}}$ terms under the VP SDE (see Appendix~\ref{proof:proposition_scheduling}).
We then have the following Proposition \ref{proposition_scheduling} with a detailed proof provided in Appendix \ref{proof:proposition_scheduling}.

\begin{assumption}
\label{assumption_uncorrelated}
(Simplifying orthogonality condition)
The approximation error $\Delta^{(1)}(\rvx,t)$ satisfies 
$\mathbb{E}[\Delta^{(1)}(\rmX^{q}_{t},t)] = \mathbf{0}$ and is uncorrelated 
with $g$, i.e., 
$\mathbb{E}[g(\rmX^{q}_{t},t)^{\top} \Delta^{(1)}(\rmX^{q}_{t},t)] = 0$.
This is adopted for analytical tractability; see Remark~\ref{remark_scheduler} for an alternative without this assumption.
\end{assumption}

\begin{proposition}[Uncertainty-Adaptive Scheduling]
\label{proposition_scheduling}
Suppose that Assumptions \ref{assumption_bounded_norm}--\ref{assumption_uncorrelated} hold.
Then, there exists a deterministic upper envelope $r'(t;\tau)$ such that $r(t;\tau) \le r'(t;\tau)$, where $r'(t;\tau) = (c_{1} + \sum_{(i,j) \in \mathcal{I}} c_{(i,j)} \cdot \frac{(1 - \bar{\alpha}(t))^i}{\bar{\alpha}(t)^j}) \tau(t)^2 - 2c_{1} \tau(t) + c_{1}$
with constants $c_1 > 0$, $c_{(i,j)} \ge 0$ and a finite index set 
$\mathcal{I} \subset \mathbb{R}_{> 0} \times \mathbb{R}_{\ge 0}$.
When $c_1 > 0$, the minimizer $\tau^*(t)$ of $r'(t;\tau)$ is given as
\vspace{-3mm}
\begin{align}
\label{tau_scheduling}
    \tau^*(t) \!=\! \argmin_{\tau} r'(t; \tau) \!=\! \left(\!1 \!+\!\!\!\sum_{(i,j) \in \mathcal{I}} \frac{c_{(i,j)}}{c_{1}} \cdot \frac{(1 - \bar{\alpha}(t))^i}{\bar{\alpha}(t)^j} \!\!\right)^{\!\!\!-\!1}\!\!\!.
\end{align}
\end{proposition}

\begin{remark}[Increasing Confidence and Practical Scheduler Design]
\label{remark_scheduler}
To minimize $r'(t;\tau)$, which is the upper bound of the score function gap (\ref{risk}), the corresponding optimal scheduling function $\tau^*(t)$ is given by (\ref{tau_scheduling}).
Note that as $t \to 0$, $\bar{\alpha}(t) \to 1$ thereby $\tau^{*}(t)$ converges to $1$. This behavior is expected since the uncertainty $u(\rvx, t)$ in Theorem \ref{theorem_1} vanishes as $t \to 0$, reflecting that the guidance approximation error becomes negligible towards the end of the diffusion process and thus justifying full reliance on the approximation.
Although computing $\tau(t)$ exactly requires access to unknown problem-specific quantities (see Appendix~\ref{proof:proposition_scheduling}), each term of the form $\frac{(1-\bar{\alpha}(t))^{i}}{\bar{\alpha}(t)^{j}}$ vanishes at $t \to 0$ and increases monotonically with $t$.
This motivates a practical single-parameter approximation
$
\tau^{*}(t) \approx \left(1 + c {(1-\bar{\alpha}(t))^{2}}/{\bar{\alpha}(t)^{2}}\right)^{\!-\!1},
$
where $c > 0$ is a tunable constant.
This approximation implies that $\tau^{*}(t)$ is monotone \emph{decreasing} in the forward diffusion time $t$ (and hence \emph{increasing along the reverse-time sampling direction}), for any positive variance schedule $\beta(t)$, offering a simple and effective control mechanism.

Without Assumption~\ref{assumption_uncorrelated}, minimizing a tractable upper bound of equation (\ref{risk}) yields a hard gating scheduler $\tau^*(t)\in\{0,1\}$ (a step function in $t$; see Appendix~\ref{proof:proposition_scheduling}).
For empirical evaluation of different choices of $c$ and the approximation form $\tau^*$, see our ablation study in Appendix \ref{appendix_experiments_details_text_image_alignment}.
\end{remark}

\paragraph{Relation to Existing Methods.}
Our scheduler parallels heuristic strategies such as guidance learning rate control~\cite{yu2023freedom} and tempering~\cite{kim2025das}, but derives directly from uncertainty analysis with a single tunable parameter, generalizing well across tasks and SGMs.

Finally, combining $\tilde{g}^{(1)}(\rvx, t)$ in (\ref{first_order_approximation}) and the parameterized $\tau(t)$ in Remark \ref{remark_scheduler}, we suggest using the following approximated score function for weighted sampling backward diffusion process with only two predefined hyperparameters $\epsilon$ and $c$ as follows.
\begin{tcolorbox}[
  colback=gray!5,
  colframe=black!70,
  boxrule=0.3pt,             
  boxsep=0.0pt,       
  left=0.5pt,         
  right=0.5pt,        
  top=-7.0pt,          
  bottom=1.5pt,       
  arc=2pt,          
]
\begin{align}
    \label{our_approach}
    \nabla_{\rvx} \log q_{t}(\rvx)  &\approx \nabla_{\rvx} \log p_{t}(\rvx) + \tilde{g}(\rvx, t) \text{~~where~~} \nonumber \\ \tilde{g}(\rvx,t)&= \frac{\bar{\alpha}(t)^{2}}{\bar{\alpha}(t)^{2} + c {(1-\bar{\alpha}(t))^{2}}} \tilde{g}^{(1)}(\rvx, t). 
\end{align}
\end{tcolorbox}

\begin{remark}[Training-Free Weighted Sampling]\label{remark:training_free}
The proposed approximation form in (\ref{our_approach}) demonstrates that, for a given score function \( \nabla_{\mathbf{x}} \log p_{t}(\mathbf{x}) \) and weight function \( w(\mathbf{x}) \), {\emph{weighted sampling can be approximated without additional training}}. This presents a substantial advantage over existing diffusion model fine-tuning methods or cross-entropy methods for weighted sampling, which require learning the density of $q$ through training for each specific $w$.
\end{remark}

\begin{remark}[High Scalability and Computational Efficiency]\label{remark:scalability}
The additional computational cost introduced by our approach, compared to the base score-based sampling, is minimal. Calculating $\bar{\rvx}'_{0}\vert_{\rvx,t}$ involves only a linear transformation of $\rvx$ and the precomputed score function. Aside from this negligible cost, the potential overhead is the computation of $\nabla_{\bar{\rvx}'_{0}\vert_{\rvx,t}} \log w(\bar{\rvx}'_{0}\vert_{\rvx,t})$ in (\ref{first_order_approximation}). This is efficiently achieved with a single gradient backpropagation step on $w$ with respect to the realization $\bar{\rvx}'_{0}\vert_{\rvx,t}$ and one inference step of the score function, i.e., evaluating
$\nabla_{\rvx}\log p_t(\rvx+\epsilon \Delta)$ in addition to $\nabla_{\rvx}\log p_t(\rvx)$.
Overall, this additional cost is \emph{significantly less demanding} than methods requiring high-order derivatives of densities or resampling.
\end{remark}

\section{Experiments}
\label{sec:experiments}

\begin{figure*}[htp]
    \centering
    \includegraphics[width=1.0\linewidth]{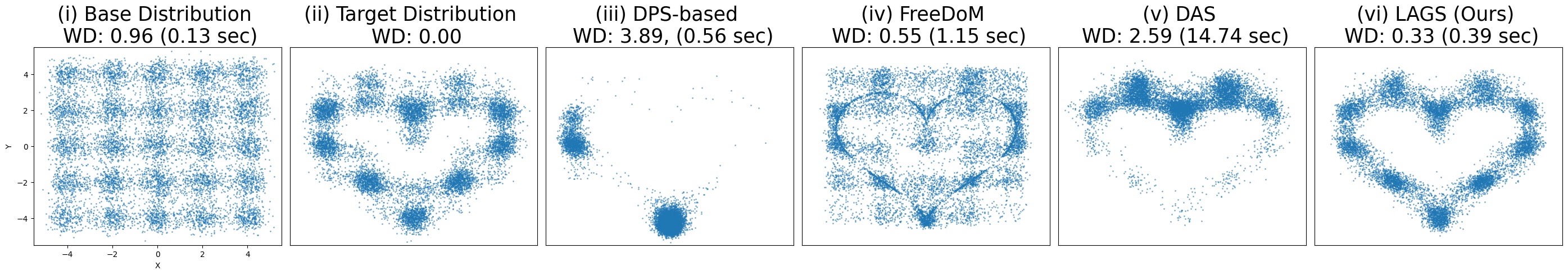}
    \vspace{-6mm}
    \caption{
    \textbf{\emph{Left} to \emph{Right}}: Base distribution, target distribution, and estimated sample densities. Wasserstein Distance and runtime (in seconds) are reported for DPS, FreeDoM, DAS, and LAGS (ours). Our method achieves the \textbf{\emph{lowest Wasserstein Distance (WD) and fastest runtime}}.
    }
    \label{fig:gm_heart_main}
    \vspace{-5mm}
\end{figure*}
We conducted comprehensive experiments to evaluate LAGS across a range of tasks. Specifically, we aimed to: \emph{(i) assess its ability to approximate complex target distributions, especially multimodal structures and exponential behavior weights; (ii) demonstrate its effectiveness on high-dimensional weighted sampling tasks; (iii) quantify the computational efficiency gains enabled by our method.}

Across various settings, our method consistently achieves the lowest Wasserstein Distance to the target distribution or yields the highest average weights. Especially in high-dimensional scenarios, it delivers a roughly 4.7$\times$ speedup over prior state-of-the-art methods while maintaining or improving task performance.

\paragraph{Baselines.} We compare our approach against the following representative training-free SGM guidance methods. DPS \cite{chungdiffusion}, originally designed for inverse problems, can be adapted for weighted sampling by leveraging its denoising mean estimator. FreeDoM \cite{yu2023freedom}, which stabilizes the reverse-time SDE sampling via a time-travel strategy; DAS \cite{kim2025das}, which uses sequential Monte Carlo with adaptive particle weighting, and has shown good performance on tasks involving high-modality base distributions. We also tested a single-particle version of DAS requiring less computational power for comparison. Pretrained SGMs such as \cite{rombach2022stablediffusion, podell2024sdxl} are also utilized as baselines.

\subsection{Training-Free Weighted Sampling over Multimodal Densities}
\label{experiments:gaussian_mixture}

\paragraph{Setup.} We evaluate the proposed method on a 2D weighted sampling task in which the base distribution is a mixture of 25 Gaussians (Fig.~\ref{fig:gm_heart_main} (i)), and the weight concentrates along a heart-shaped manifold, defining a structured target distribution (Fig.~\ref{fig:gm_heart_main} (ii)). Implementation details are provided in Appendix~\ref{appendix_experiments_toy_examples}. The task is analytically well-defined, enabling quantitative evaluation via the Wasserstein Distance (WD) between the generated samples and the ground-truth target distribution. We report both WD and wall-clock time required to generate $10^4$ samples along with the generated samples by each algorithm in Fig.~\ref{fig:gm_heart_main} (iii)–(vi).

\paragraph{Results.} As shown in Fig.~\ref{fig:gm_heart_main}, our method (LAGS) achieves both the lowest WD and fastest runtime. It generates all samples in 0.33 seconds without computing any second-order derivatives or resampling. In contrast, baselines such as FreeDoM and DAS incur higher computational cost due to iterative refinement or generating particles for sequential Monte Carlo methods.

FreeDoM provides broad coverage but often produces samples outside the support of the target distribution (where $q(\rvx) \approx 0$). DAS improves precision by concentrating on high-density regions, but this comes at substantial computational expense. Our method achieves both high coverage and high precision, accurately approximating the support of $q(\rvx)$ with markedly improved efficiency. We provide additional examples in Appendix~\ref{appendix_experiments_toy_examples}.

\subsection{Application to Training-Free Text-to-Image Alignment with Human Preference Scores}
\label{subsection:reward_maximization}

\begin{figure*}[t]
    \centering
    \includegraphics[width=1.0\linewidth]{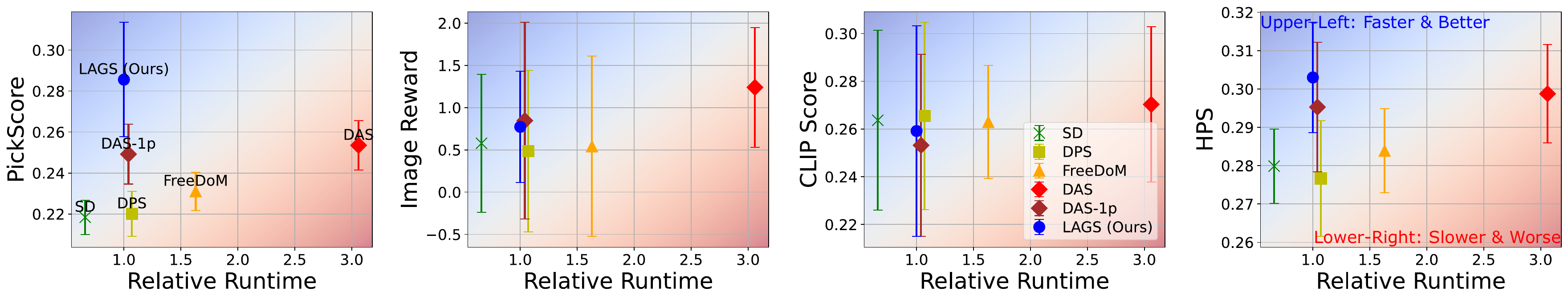}
    \vspace{-7mm}
    \caption{
    \textbf{Results on SD}: Comparative performance across PickScore, ImageReward, CLIP Score, and HPS (left to right), with runtime normalized to LAGS. Our method achieves the best performance in PickScore and HPS with the lowest runtime among the guidance methods. We adopt the benchmark set of prompts from \cite{kim2025das}.
    }
    \label{fig:performance_sd}
    \includegraphics[width=1.0\linewidth]{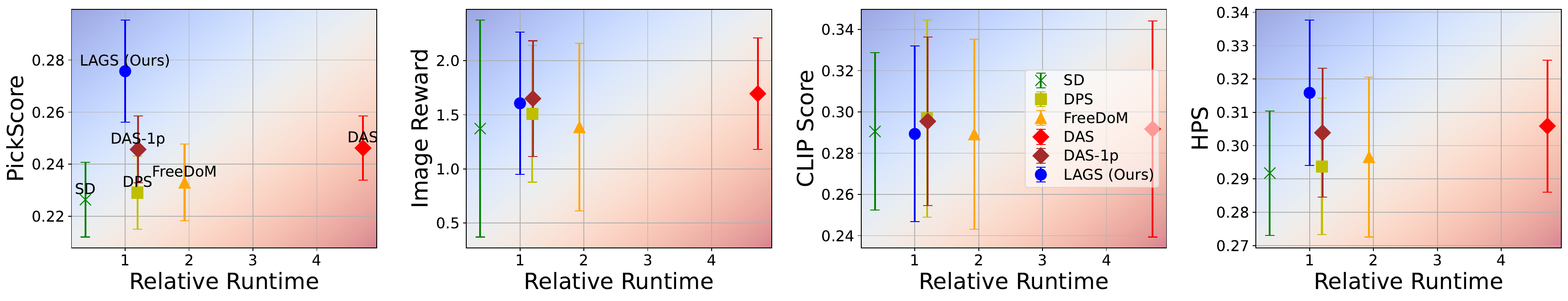}
    \vspace{-7mm}
    \caption{
    \textbf{Results on SDXL}: Evaluation on the same metrics as in Fig.~\ref{fig:performance_sd}. LAGS consistently achieves the best performance in PickScore (target metric) and HPS with the lowest runtime among the guidance methods, breaking the typical trade-off between high score and computation complexity.
    }
    \label{fig:performance_sdxl}
    \vspace{-5mm}
\end{figure*}

We further evaluated our method on high-dimensional generative tasks, focusing on text-to-image synthesis guided by human preference scores. In this setting, the base SGM is a prompt-conditioned image generator, and the weight function $w(\rvx)$ corresponds to a predefined human preference score for a given text-image pair. The objective is to maximize the expected preference score—i.e., average weight—by prioritizing samples that are better aligned with human preference. 
Such scenarios demand both high alignment accuracy and computational efficiency, as they directly impact user experience and system scalability in real-world deployments.

\paragraph{Limitations of the experimental scope.}
We focus on training-free guidance with a target weight function, distinct from approaches based on text embedding manipulation~\cite{kang2025rare}, model fine-tuning~\cite{blacktraining}, or Large Language Models~\cite{chen2024gentron}. For fairness, we restrict comparisons to training-free guidance methods.

\paragraph{Setup, Metrics, and Fairness.}
We follow the evaluation protocol where $w(\rvx)$ is defined via PickScore~\cite{kirstain2023pick}, a learned alignment metric trained from human preferences. All methods use the same pretrained SGMs—Stable Diffusion (SD)~\cite{rombach2022stablediffusion} and Stable Diffusion XL (SDXL)~\cite{podell2024sdxl}—with a fixed number of backward diffusion steps ($T$). 

For SD, we adopt a prompt set from~\cite{kim2025das}. For SDXL, we combine prompts from~\cite{wu2023better},~\cite{kim2025das}, and additional manually curated samples to assess generalization. Evaluation includes expected weight $\mathbb{E}[w(\rmX^{q})]$, PickScore, and runtime. We also report secondary metrics: CLIP Score~\cite{hessel2021clipscore}, ImageReward~\cite{xu2023imagereward}, and Human Preference Score (HPS)~\cite{wu2023hps}.

Figures~\ref{fig:performance_sd} and~\ref{fig:performance_sdxl} show the results on SD and SDXL, respectively. Each subplot presents one metric (PickScore, ImageReward, CLIP Score, HPS from left to right), with runtime shown on the horizontal axis relative to our method (normalized to $1\times$). For SDXL, our method completes sampling in 85 seconds per image, while DAS requires approximately 4.72$\times$ more time.

\paragraph{Results.}
The proposed method, LAGS, consistently achieves the \emph{highest PickScore} and \emph{lowest runtime} across both SD and SDXL settings, as shown in the leftmost columns of Figures~\ref{fig:performance_sd} and~\ref{fig:performance_sdxl}. This represents a substantial improvement over existing baselines, which typically exhibit a trade-off between target performance and computational cost. In contrast, LAGS breaks this trend by achieving both higher weight, i.e., PickScore, and faster sampling.

Among baselines, DAS delivers strong performance but incurs a higher computational cost. FreeDoM and DPS improve upon the raw SD and SDXL outputs in terms of PickScore, but remain notably less effective than LAGS in both performance and efficiency.

\begin{figure*}[t]
    \centering
    \begin{tabular}{@{}c@{}c@{}}

            \includegraphics[width=1.0\linewidth]{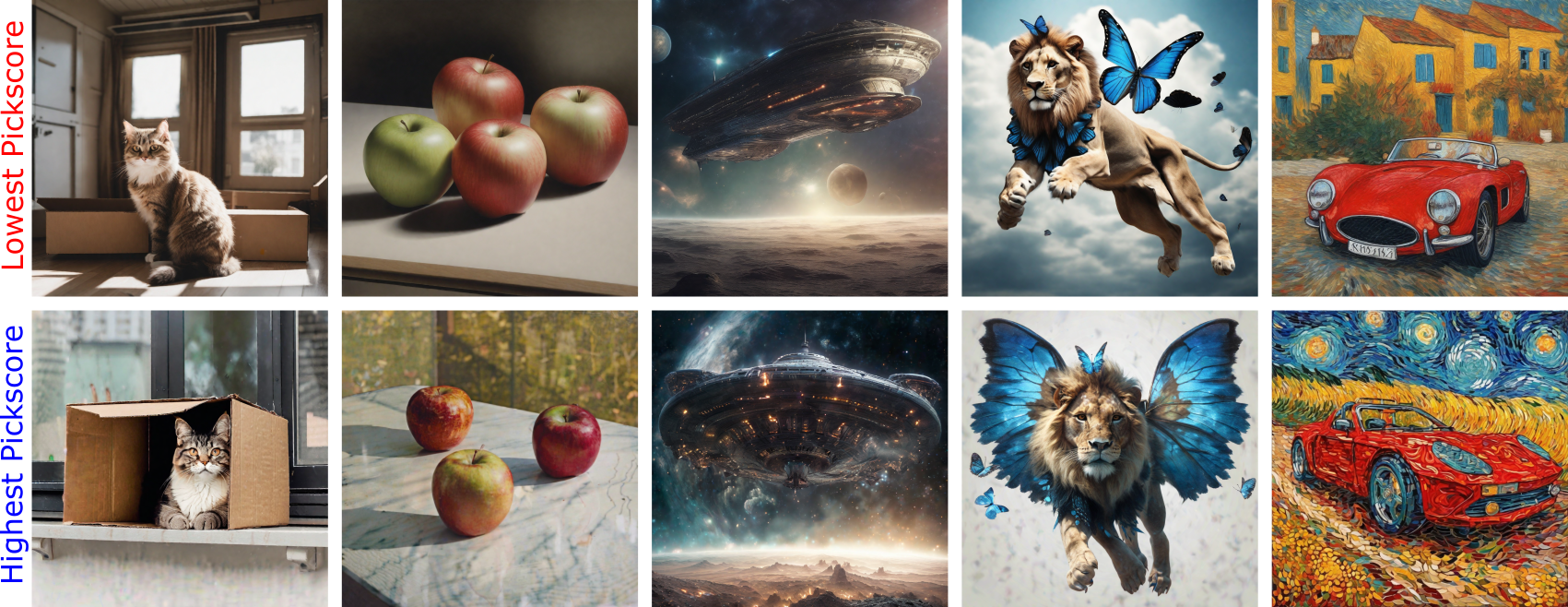}
            \\
    \end{tabular}\vspace{-3mm}
    \caption{\textbf{\emph{Sampling results with the proposed lightweight guidance $\tilde{g}(\rvx, t)$ on a high-dimensional domain.}} \emph{Top row:} Samples with the \textbf{lowest PickScore}; \emph{Bottom row:} those with the \textbf{highest PickScore}, selected from all generations.
Prompts (left to right):  
\emph{``A cat sitting inside a cardboard box next to a window'', ``Three apples on a table'', ``A majestic, photorealistic alien spaceship drifting before a vast galaxy'', ``A photo-realistic image of flying lion with blue butterfly wings'', ``A red sports car in the style of Vincent van Gogh''}.  %
{Notably, for all prompts, the highest PickScore samples are achieved by our method}. Additional qualitative results are provided in Appendix~\ref{appendix_text_image_alignment_additional_results}.
}
\label{fig:minmax_pickscore_examples}
\vspace{-1mm}
\end{figure*}

To qualitatively illustrate the impact of PickScore on the generated outputs, we present visual examples in Fig.~\ref{fig:minmax_pickscore_examples}. The top row depicts samples with the \emph{lowest} PickScores for a given prompt, while the bottom row shows those with the \emph{highest} PickScores among outputs generated by each method. Samples with low PickScores often display semantic or stylistic failures, such as omission of positional cues (e.g., ``next to a window''), incorrect object counts (e.g., four apples instead of three), or reduced representational fidelity (the associated text prompts are shown in Fig.~\ref{fig:minmax_pickscore_examples}). Consistent with the trends shown in Figures~\ref{fig:performance_sd}-\ref{fig:performance_sdxl}, the highest PickScores for the selected prompts are attained by our method. Additional qualitative examples are provided in Appendix~\ref{appendix_text_image_alignment_additional_results}.

\paragraph{Cross-Metric Generalization.}
To assess whether improvements in PickScore extend to other alignment metrics, we evaluate ImageReward~\cite{xu2023imagereward}, CLIP Score~\cite{hessel2021clipscore}, and HPS~\cite{wu2023hps}. As shown in Fig.~\ref{fig:performance_sd}, LAGS achieves the highest HPS on SD, while DAS performs better on ImageReward and CLIP Score. On SDXL (Fig.~\ref{fig:performance_sdxl}), LAGS still outperforms all baselines on PickScore and HPS, where the margin against DAS on ImageReward and CLIP score is minor (less than 0.1 and 0.06, respectively), despite being faster. These results indicate that LAGS’s reward-aligned sampling generalizes well to diverse human-preference proxies with high efficiency.

We further report image quality metrics in Table~\ref{tab:sdxl_runtime_brisque_maniqa} using the widely adopted no-reference measure BRISQUE~\cite{mittal2012no} and MANIQA metrics~\cite{yang2022maniqa}. While DAS achieves the best overall image quality, our method attains comparable scores with only a modest gap—within the variability across guidance methods—despite its substantially lower runtime.

\paragraph{Hyperparameter Sensitivity.}
Our method introduces only a small number of hyperparameters. In all experiments, we fix the confidence parameter to $c\!=\!10$ across models and tasks, yet LAGS consistently outperforms baseline methods.
By construction, the scheduler in (\ref{tau_scheduling}) smoothly modulates the guidance strength between $0$ and $1$ (with $\tau(0) = 1$), and we observe that performance remains stable over a wide range of $c$ values (approximately $0.1 \le c \le 20$).
Appendix~\ref{appendix_text_image_alignment_ablation} provides a detailed ablation study on the effect of varying $c$ and further confirms the robustness of the proposed scheduler.

\paragraph{Computational Efficiency.}
A key advantage of LAGS lies in its computational efficiency achieved by avoiding second-order derivative computations and eliminating the need for resampling techniques. Notably, the efficiency gain becomes more pronounced as model size increases. On SD, LAGS achieves a 6\% speedup over DAS-1p and DPS. With the larger SDXL model, the improvement grows to 16.7\%, demonstrating the method’s scalability with model complexity. Compared to resampling-based approaches, the gap is even more substantial, for example, LAGS is $1.9\times$ and $4.7\times$ faster than FreeDoM and DAS, respectively, on SDXL.

\begin{table}[t]
\centering
\scriptsize
\begin{tabular}{lccc}
\toprule
Method & Runtime [$\times$ 85s] $\downarrow$ & BRISQUE $\downarrow$ & MANIQA $\uparrow$ \\
\midrule
SDXL (default)                 & 0.377 & 21.48 $\pm$ 12.45 & 0.733 $\pm$ 0.139 \\
DPS                  & 1.193 & 24.52 $\pm$ 13.03 & 0.720 $\pm$ 0.139 \\
FreeDoM      & 1.930 & 34.49 $\pm$ 15.20 & 0.711 $\pm$ 0.144 \\
DAS                  & 4.722 & \bf{19.55 $\pm$ 12.75} & 0.720 $\pm$ 0.145 \\
DAS-1P               & 1.202 & 22.56 $\pm$ 11.97 & \bf{0.728 $\pm$ 0.130} \\
LAGS (ours)     & \bf{1.000} & 26.40 $\pm$ 12.99 & 0.720 $\pm$ 0.134 \\
\bottomrule
\end{tabular}
\caption{Runtime and no-reference image quality measures (BRISQUE, MANIQA).
Values are mean $\pm$ standard deviation. Bold indicates best among guidance methods.}
\label{tab:sdxl_runtime_brisque_maniqa}
\vspace{-5mm}
\end{table}

\subsection{Additional Applications}
\label{sec:additional_applications}

Owing to its efficiency and scalability, the proposed method extends naturally to a range of practical scenarios with diverse weight functions $w(\rvx)$. Below, we summarize additional applications: \textbf{Sampling for Fairness.} Using classifier outputs as weights enables our method to target underrepresented or misclassified groups, serving as a training-free tool for mitigating class bias. We demonstrate this in Appendix~\ref{appendix_experiments_details_celeba} and~\ref{appendix_experiments_details_mnist} across multiple datasets; \textbf{Sampling Control Beyond Prompts.} Beyond prompt conditioning, LAGS supports rapid fine-grained control via custom $w(\rvx)$. In Appendix~\ref{appendix_experiments_details_stablecascade}, we showcase frequency- and color-sensitive sampling by emphasizing spectral features through $w(\rvx)$ \emph{without prompt control}, highlighting LAGS’s versatility for controllable generation.

\section{Conclusion}
We first propose a lightweight first-order approximation that avoids second-order gradient computations of the generative model. To mitigate potential approximation gaps, we introduce an uncertainty-adaptive scaling mechanism. Together, these components enable LAGS to achieve both strong empirical performance and excellent scalability with a fraction of the computational cost of state-of-the-art baselines. As larger SGMs continue to be developed and released, we anticipate that the proposed Hessian-free, lightweight formulation will become increasingly relevant for efficient and stable weighted sampling.

\newpage

\section*{Acknowledgements}
This work was supported in part by the National Science Foundation under Grant 2148224 and Grant 2443857; in part by the Army Research Office (ARO) under Award W911NF2310062; in part by the Office of Naval Research (ONR) under Award N000142412542; CNS-2212202; in part by Office of the Under Secretary of Defense for Research and Engineering (OUSD R\&E), National Institute of Standards and Technology (NIST), and industry partners as specified in the Resilient and Intelligent NextG Systems (RINGS) Program; and in part by Wireless Networking and Communications Group (WNCG)/6G@UT.

{
    \small
    \bibliographystyle{ieeetr}
    \bibliography{references_sde, references_importance_sampling, references_generative_model, references_csi}
}


\newpage
\appendix
\onecolumn

{\small
\begin{spacing}{0.7}
\tableofcontents
\end{spacing}
}

\newpage

\section*{Impact Statement}
\label{section:Impact_Statements}
The proposed method facilitates weighted sampling for a wide range of existing pretrained score-based models and weight functions. We explicitly clarify that \emph{{the primary objective of this study is to advance methodological rigor and the field of Machine Learning}}; this work neither intends to advocate nor promote any particular social perspective through the experimental results presented. We do acknowledge the potential for this approach to be misused in reinforcing biased sampling practices, though, and thus strongly encourage careful consideration of the implications associated with its application.

\section{Related Work}\label{sec:related}

\paragraph{Importance sampling.}
Importance sampling \cite{rubinstein2016simulation} and weighted sampling play a key role in various foundational tasks such as variance reduction, data augmentation, root cause analysis, and reliability assessment \cite{grover2019bias, antoniou2017data, li2024efficient}. However, directly estimating the weighted PDF $q$ via Monte Carlo methods using samples drawn from $p$ is challenging, primarily due to instability in accurately computing the normalization constant. 

To address this, conventional techniques often leverage Cross-Entropy methods \cite{rubinstein1999cross, rubinstein2001combinatorial}, which iteratively optimize a parameterized PDF to minimize the Kullback-Leibler (KL) divergence from the target weighted sampling PDF. However, for complex, high-dimensional datasets—such as natural images with intractable distributions—the selection of a suitable class of PDFs becomes highly challenging. Recent developments in generative modeling offer a solution via bijective mappings, where a tractable initial distribution is progressively transformed through the change-of-variables formula, resulting in a valid end distribution known as normalizing flow \cite{rezende2015variational, kobyzev2020normalizing}. Normalizing flow-based methods have been applied to weighted sampling, employing efficient invertible functions as in \cite{muller2019neural} or interpretable, shape-constrained networks \cite{xiang2024importance}.

Nevertheless, Cross-Entropy or normalizing flow-based methods often exhibit limited representational flexibility and can struggle with mapping selection \cite{zhang2021diffusion, cornish2020relaxing}. Additionally, training separate generative models for distinct weight functions is computationally prohibitive, particularly for applications requiring diverse importance criteria or multiple estimations across varied weighting functions.

\paragraph{Score-based generative models.}

Recent advances in generative modeling have focused on score-based models, which learn the score function of a target distribution instead of estimating its PDF directly \cite{vincent2011connection, song2021score, hyvarinen2005estimation}. This approach enables efficient sampling via stochastic differential equations or iterative methods \cite{song2021score, ho2020ddpm, song2021ddim}, achieving notable success in high-resolution image generation \cite{croitoru2023diffusion}. By eliminating the need for shape-constrained bijective mappings required by normalizing flow-based models, score-based models allow for greater design flexibility. 
Direct applications of existing score-based methods to weighted sampling, however, present challenges, as they require samples from the target distribution, limiting adaptability. Also, training distinct score-based generative models for various weight functions is inefficient and impractical. 

To address these challenges, recent techniques \cite{kim2025das} model weighted sampling as a diffusion process and approximate its score function using the original PDF’s score combined with the weight function, rather than learning the weighted-sampling score directly. This avoids training separate generative models for different importance weight functions while still fully exploiting the expressive power of score-based models.

\paragraph{Connections and differences between conditional and weighted sampling.} 
 Advances in generative modeling have enabled conditional generation across various modalities, including text-to-image synthesis \cite{mansimov2015generating, reed2016generative, li2019controllable, yu2022scaling, nichol2022glide, podell2024sdxl}, image-to-image translation \cite{isola2017image, richardson2021encoding, liu2017unsupervised}, and text-to-audio generation \cite{liu2023audioldm}. These approaches typically rely on explicit conditioning during training to enforce alignment between input modalities and output distributions.

More recently, score-based generative frameworks have introduced a new class of methods for \emph{conditional sampling} that do not require conditioning during training. Instead, conditioning is applied at inference time by modifying the sampling dynamics of pretrained, unconditional models \cite{mengsdedit, trippediffusion, wu2024practical}. This contrasts with conventional methods that train diffusion models explicitly with conditioning \cite{dhariwal2021diffusion}.

While conceptually related, weighted sampling differs from conditional sampling as conditional sampling typically enforces constraints derived from partial observations or degraded inputs, where the conditioning variable is inherently tied to the instance being sampled (e.g., an incomplete image in inverse problems \cite{rout2024beyond, rout2024solving}) and sometimes such noisy observation can be modelled with known quantities. In contrast, weighted sampling reweights samples drawn from a base distribution using an externally defined weight function. This function need not be derived from or coupled with the original distribution, thereby enabling flexible sampling objectives that extend beyond pointwise conditioning.

Nonetheless, several frameworks originally proposed for conditional generation can be adapted for weighted sampling, as they share the common goal of steering the sampling trajectory toward a desired distribution. For example, the guidance mechanisms in \cite{chungdiffusion, yu2023freedom}, while designed for solving inverse problems, can be naturally repurposed for importance-weighted sampling due to their structural similarity. As such, we include these representative diffusion guidance methods in our performance comparison.

Additionally, recent efforts to enhance sample quality and estimation accuracy have focused on improving the sampling dynamics in diffusion models. Techniques such as stochastic gradient-based refinement \cite{kim2023refining, doucet2022score} and adaptive MCMC approaches \cite{deng2020contour, deng2022adaptively, deng2022interacting} have demonstrated strong performance in sampling from complex, multimodal distributions. In particular, the sequential Monte Carlo (SMC) strategy proposed in \cite{kim2025das} provides an effective mechanism for reward-aware sampling through recursive estimation and adaptive resampling. This method has shown substantial improvements over finetuning-based alternatives and is included in our evaluation as a high-performing baseline.

\paragraph{Scope of This Work.}
The primary objective of this work is to develop an efficient and scalable algorithm for performing weighted sampling using score-based generative models (SGMs). A key motivation stems from the observation that existing guidance-based sampling methods often incur substantial computational overhead due to repeated resampling or multi-step refinement procedures, particularly when applied to large-scale diffusion models. These approaches typically require accurate estimation of the score function associated with the target density, which can be prohibitively expensive in practice.

To address this limitation, we propose a lightweight approximation framework for constructing the target guidance function. Building on this, we further introduce an adaptive scheduling strategy derived from the theoretical analysis of guidance accuracy, enabling efficient control of the sampling dynamics. The resulting algorithm is well-suited for practical deployment across diverse generative weighted sampling tasks.

\section{Technical Results}
\label{appendix_technical_results}

\begin{table}[t]
\caption{Notation and Description}
\label{tab:notation_description}
\vskip 0.15in
\begin{center}
\begin{small}
\begin{tabular}{lcccr}
\toprule
Notation & Description & Note \\
\midrule
$\rmX^{q}_{t}$    & A random vector of the weighted sampling diffusion process at time $t$  & $\forall t, \rmX^{q}_{t} \in \mathbb{R}^{d}$  \\
$\rmX^{p}_{t}$    & A random vector of the original (base) diffusion process at time $t$  & $\forall t, \rmX^{p}_{t} \in \mathbb{R}^{d}$  \\
$\rmZ$    & A Gaussian random vector & $\rmZ \sim \mathcal{N}(\mathbf{0}, \mathbf{I})$  \\
\hline
$w(\rvx)$    & a weight function &  $0 < m < w(\rvx) < \infty$\\
$q(\rvx)$    & The weighted sampling PDF   & $q(\rvx) = q_{0}(\rvx)$\\
$q_{t}(\rvx)$    & The PDF of $\rmX^{q}_{t}$  & \\
$p(\rvx)$    & The original (base) PDF &  $p(\rvx) = p_{0}(\rvx)$\\
$p_{t}(\rvx)$    & The PDF of $\rmX^{p}_{t}$  & \\
$\nabla_{\rvx} \log q_{t}(\rvx)$ & The score function of $q_{t}$ &\\
$\nabla_{\rvx} \log p_{t}(\rvx)$ & The score function of $p_{t}$ &\\
$\mathcal{N}(\mu, \sigma^2)$ & A Gaussian distribution with mean $\mu$ and variance $\sigma^2$ &\\
$\mathcal{U}(\mathcal{S})$ & A uniform distribution over space $\mathcal{S}$ &\\
$\bar{\rvx}'_{0}\vert_{\rvx,t}$ & The conditional mean of $\rmX^{p}_{0}$ for a given observation $\rvx$ at $t$ & \\

\hline
$[a,b]$    & A closed interval from $a\in \mathbb{R}$ to $b\in \mathbb{R}$ &  \\
$[v]_{i} $    & The $i$-th element of a vector $v$ & \\
$[M]_{i,j}$    & The $(i,j)$-th element of a matrix $M$ & \\
$\mathbb{N}$ & The set of natural numbers & \\
$\mathbb{R}$ & The set of real numbers & \\

\bottomrule
\end{tabular}
\end{small}
\end{center}
\vskip -0.1in
\end{table}

\subsection{Proof of Theorem \ref{theorem_1}}
\label{proof_theorem_1}
\begin{proof}

Fix $t\in[0,T]$ and $\rvx\in\mathbb{R}^{d}$.
Recall that the weighted-sampling diffusion marginal can be written as
\[
q_t(\rvx)\ \propto\ p_t(\rvx)\,
\mathbb{E}\!\left[w(\rmX^{p}_{0})\mid \rmX^{p}_{t}=\rvx\right],
\]
where the proportionality constant does not depend on $\rvx$.
Therefore,
\begin{equation}
\label{eq:score_decomposition_qt}
\nabla_{\rvx}\log q_t(\rvx)-\nabla_{\rvx}\log p_t(\rvx)
=
\nabla_{\rvx}\log \mathbb{E}\!\left[w(\rmX^{p}_{0})\mid \rmX^{p}_{t}=\rvx\right].
\end{equation}

Let $\mu(\rvx)\coloneqq \bar{\rvx}'_{0}\vert_{\rvx,t}$ and define
\[
v(\rvx,t)\ \coloneqq\ \nabla_{\mu(\rvx)}\log w(\mu(\rvx)).
\]
By the definition of the first-order approximated guidance (finite-difference HVP),
\begin{equation}
\label{eq:def_tilde_g1_for_proof}
\tilde g^{(1)}(\rvx,t)
\coloneqq
\frac{1}{\sqrt{\bar{\alpha}(t)}}\,v(\rvx,t)
+
\frac{1-\bar{\alpha}(t)}{\sqrt{\bar{\alpha}(t)}}\,
\frac{\nabla_{\rvx}\log p_t(\rvx+\epsilon v(\rvx,t))-\nabla_{\rvx}\log p_t(\rvx)}{\epsilon}.
\end{equation}

Using (\ref{eq:score_decomposition_qt}), the approximation error satisfies
\begin{align}
\label{eq:score_gap_start}
\left\Vert g(\rvx,t)-\tilde g^{(1)}(\rvx,t)\right\Vert
&=
\left\Vert
\nabla_{\rvx}\log \mathbb{E}\!\left[w(\rmX^{p}_{0})\mid \rmX^{p}_{t}=\rvx\right]
-\tilde g^{(1)}(\rvx,t)
\right\Vert \nonumber\\
&\le
\underbrace{
\Bigl\Vert
\nabla_{\rvx}\log \mathbb{E}\!\left[w(\rmX^{p}_{0})\mid \rmX^{p}_{t}=\rvx\right]
-\nabla_{\rvx}\log w(\mu(\rvx))
\Bigr\Vert}_{\eqqcolon\,T_1(\rvx,t)}
+
\underbrace{
\Bigl\Vert
\nabla_{\rvx}\log w(\mu(\rvx))-\tilde g^{(1)}(\rvx,t)
\Bigr\Vert}_{\eqqcolon\,T_2(\rvx,t)}.
\end{align}
By the chain rule and $\mu(\rvx)=\frac{1}{\sqrt{\bar{\alpha}(t)}}\bigl(\rvx+(1-\bar{\alpha}(t))\nabla_{\rvx}\log p_t(\rvx)\bigr)$, we have
\begin{equation}
\label{eq:nabla_logw_mu_exact}
\nabla_{\rvx}\log w(\mu(\rvx))
=
\frac{\left(\rmI+(1-\bar{\alpha}(t))H_{\log p_t}(\rvx)\right)}{\sqrt{\bar{\alpha}(t)}}\,v(\rvx,t).
\end{equation}
Therefore, subtracting (\ref{eq:def_tilde_g1_for_proof}) from (\ref{eq:nabla_logw_mu_exact}) yields
\begin{align}
\label{eq:T2_hvp_gap}
T_2(\rvx,t)
&=
\Bigl\Vert
\nabla_{\rvx}\log w(\mu(\rvx))-\tilde g^{(1)}(\rvx,t)
\Bigr\Vert \nonumber\\
&=
\frac{1-\bar{\alpha}(t)}{\sqrt{\bar{\alpha}(t)}}\,
\left\Vert
H_{\log p_t}(\rvx)\,v(\rvx,t)
-
\frac{\nabla_{\rvx}\log p_t(\rvx+\epsilon v(\rvx,t))-\nabla_{\rvx}\log p_t(\rvx)}{\epsilon}
\right\Vert.
\end{align}

We now bound the two terms $T_1(\rvx,t)$ and $T_2(\rvx,t)$ separately.
By applying the chain rule, the term $T_1(\rvx,t)$ can be further reformulated as follows.

\paragraph{Bounding $T_1(\rvx,t)$.}
Define the Jensen gap
\[
J_t(\rvx)\ \coloneqq\ \mathbb{E}\!\left[w(\rmX^{p}_{0})\mid \rmX^{p}_{t}=\rvx\right]-w(\mu(\rvx)).
\]
Then
\begin{align}
\label{eq:T1_fraction_form}
T_1(\rvx,t)
&=
\left\Vert
\nabla_{\rvx}\log \mathbb{E}\!\left[w(\rmX^{p}_{0})\mid \rmX^{p}_{t}=\rvx\right]
-\nabla_{\rvx}\log w(\mu(\rvx))
\right\Vert \notag\\
&=
\left\Vert
\frac{\nabla_{\rvx}\mathbb{E}[w(\rmX^{p}_{0})\mid \rmX^{p}_{t}=\rvx]}
{\mathbb{E}[w(\rmX^{p}_{0})\mid \rmX^{p}_{t}=\rvx]}
-
\frac{\nabla_{\rvx}w(\mu(\rvx))}{w(\mu(\rvx))}
\right\Vert \notag\\
&=
\left\Vert
\frac{w(\mu(\rvx))\,\nabla_{\rvx}J_t(\rvx) - J_t(\rvx)\,\nabla_{\rvx}w(\mu(\rvx))}
{\mathbb{E}[w(\rmX^{p}_{0})\mid \rmX^{p}_{t}=\rvx]\; w(\mu(\rvx))}
\right\Vert \notag\\
&\le
\frac{\Vert \nabla_{\rvx}J_t(\rvx)\Vert}{\mathbb{E}[w(\rmX^{p}_{0})\mid \rmX^{p}_{t}=\rvx]}
+
\frac{|J_t(\rvx)|\,\Vert \nabla_{\rvx}w(\mu(\rvx))\Vert}
{\mathbb{E}[w(\rmX^{p}_{0})\mid \rmX^{p}_{t}=\rvx]\; w(\mu(\rvx))} \notag\\
&\le
\frac{\Vert \nabla_{\rvx}J_t(\rvx)\Vert}{m}
+
\frac{|J_t(\rvx)|\,\Vert \nabla_{\rvx}\log w(\mu(\rvx))\Vert}{m},
\end{align}
where we used $w(\cdot)\ge m$ so that $\mathbb{E}[w(\rmX^{p}_{0})\mid \rmX^{p}_{t}=\rvx]\ge m$ and
$\Vert \nabla_{\rvx}w(\mu)\Vert = w(\mu)\Vert \nabla_{\rvx}\log w(\mu)\Vert$.

\paragraph{Bounding $J_t(\rvx)$ and $\nabla_{\rvx}J_t(\rvx)$.}
Let $e(\rvy,\rvx)\coloneqq \rvy-\mu(\rvx)$.
Since $\Vert H_w(\rvz)\Vert\le \eta_2$ for all $\rvz$ (Assumption~\ref{assumption_bounded_norm}),
Taylor's theorem implies the smoothness inequality
\begin{align}
\label{eq:w_smoothness}
\Bigl| w(\rvy)-w(\mu(\rvx))-\nabla w(\mu(\rvx))^{\top}(\rvy-\mu(\rvx)) \Bigr|
\le \frac{\eta_{2}}{2}\,\|e(\rvy,\rvx)\|^{2}.
\end{align}
Taking conditional expectation w.r.t.\ $\rmX^{p}_{0}\sim p_{\rmX^{p}_{0}\mid \rmX^{p}_{t}}(\cdot\mid \rvx)$ and using
$\mathbb{E}[e(\rmX^{p}_{0},\rvx)\mid \rmX^{p}_{t}=\rvx]=0$, we obtain
\begin{align}
\label{eq:J_bound}
|J_t(\rvx)|
\le \frac{\eta_{2}}{2}\,
\mathbb{E}\!\left[\left\|e(\rmX^{p}_{0},\rvx)\right\|^{2}\ \Big|\ \rmX^{p}_{t}=\rvx\right].
\end{align}

Define the conditional score
\[
S(\rvy,\rvx,t)\ \coloneqq\ \nabla_{\rvx}\log p_{\rmX^{p}_{0}\mid \rmX^{p}_{t}}(\rvy\mid \rvx).
\]
By differentiation under the integral sign,
\[
\nabla_{\rvx}\mathbb{E}\!\left[w(\rmX^{p}_{0})\mid \rmX^{p}_{t}=\rvx\right]
=
\mathbb{E}\!\left[w(\rmX^{p}_{0})\,S(\rmX^{p}_{0},\rvx,t)\mid \rmX^{p}_{t}=\rvx\right],
\]
and since $\int p_{\rmX^{p}_{0}\mid \rmX^{p}_{t}}(\rvy\mid \rvx)\,d\rvy=1$, we have
$\mathbb{E}[S(\rmX^{p}_{0},\rvx,t)\mid \rmX^{p}_{t}=\rvx]=0$.
Using the chain rule $\nabla_{\rvx}w(\mu(\rvx))=(\nabla_{\rvx}\mu(\rvx))^\top \nabla w(\mu(\rvx))$
and $\nabla_{\rvx}\mu(\rvx)=\mathbb{E}[e(\rmX^{p}_{0},\rvx)S(\rmX^{p}_{0},\rvx,t)^\top\mid \rmX^{p}_{t}=\rvx]$,
we obtain
\begin{align}
\label{eq:gradJ_expression}
\nabla_{\rvx}J_t(\rvx)
=
\mathbb{E}\!\Bigl[
\Bigl(w(\rmX^{p}_{0})-w(\mu(\rvx))-\nabla w(\mu(\rvx))^{\top}e(\rmX^{p}_{0},\rvx)\Bigr)\,
S(\rmX^{p}_{0},\rvx,t)
\ \Big|\ \rmX^{p}_{t}=\rvx
\Bigr],
\end{align}
where we used $\mathbb{E}[S(\rmX^{p}_{0},\rvx,t)\mid \rmX^{p}_{t}=\rvx]=0$ to insert the term $-w(\mu(\rvx))$.
Hence, by Jensen's inequality, $\|S(\cdot,\rvx,t)\|\le \gamma_t$ (Assumption~\ref{assumption_lipschitz}),
and (\ref{eq:w_smoothness}),
\begin{align}
\label{eq:gradJ_bound}
\|\nabla_{\rvx}J_t(\rvx)\|
&\le
\gamma_t\,
\mathbb{E}\!\left[
\Bigl|w(\rmX^{p}_{0})-w(\mu(\rvx))-\nabla w(\mu(\rvx))^{\top}e(\rmX^{p}_{0},\rvx)\Bigr|
\ \Big|\ \rmX^{p}_{t}=\rvx
\right] \notag\\
&\le
\frac{\eta_{2}\gamma_{t}}{2}\,
\mathbb{E}\!\left[\left\|e(\rmX^{p}_{0},\rvx)\right\|^{2}\ \Big|\ \rmX^{p}_{t}=\rvx\right].
\end{align}

\paragraph{Bounding $\|\nabla_{\rvx}\log w(\mu(\rvx))\|$.}
From (\ref{eq:nabla_logw_mu_exact}) and $\|H_{\log p_t}(\rvx)\|\le \zeta_t$ (Assumption~\ref{assumption_lipschitz}),
and $\|v(\rvx,t)\|\le \eta$ (Assumption~\ref{assumption_bounded_norm}),
\begin{align}
\label{eq:nabla_logw_mu_bound}
\left\Vert \nabla_{\rvx}\log w(\mu(\rvx)) \right\Vert
\le
\frac{1+(1-\bar{\alpha}(t))\zeta_t}{\sqrt{\bar{\alpha}(t)}}\,\eta.
\end{align}

\paragraph{Putting together the bound for $T_1(\rvx,t)$.}
Substituting (\ref{eq:J_bound}), (\ref{eq:gradJ_bound}), and (\ref{eq:nabla_logw_mu_bound}) into (\ref{eq:T1_fraction_form}) yields
\begin{align}
\label{eq:T1_final_bound}
T_1(\rvx,t)
\le
\left(
\frac{\gamma_t\eta_2}{2m}
+
\frac{(1+(1-\bar{\alpha}(t))\zeta_t)\,\eta\,\eta_2}{2m\sqrt{\bar{\alpha}(t)}}
\right)
\mathbb{E}\!\left[\left\|\rmX^{p}_{0}-\mu(\rvx)\right\|^{2}\ \Big|\ \rmX^{p}_{t}=\rvx\right].
\end{align}

\paragraph{Bounding $T_2(\rvx,t)$.}
Using Taylor's theorem with integral remainder for $\nabla_{\rvx}\log p_t$,
\[
\nabla_{\rvx}\log p_t(\rvx+\epsilon v)
=
\nabla_{\rvx}\log p_t(\rvx)
+
\int_{0}^{\epsilon} H_{\log p_t}(\rvx+s v)\,v\,ds,
\]
we obtain
\[
\frac{\nabla_{\rvx}\log p_t(\rvx+\epsilon v)-\nabla_{\rvx}\log p_t(\rvx)}{\epsilon}
-
H_{\log p_t}(\rvx)\,v
=
\frac{1}{\epsilon}\int_{0}^{\epsilon}\bigl(H_{\log p_t}(\rvx+s v)-H_{\log p_t}(\rvx)\bigr)\,v\,ds.
\]
Taking norms and using the Hessian Lipschitz property
$\|H_{\log p_t}(\rvx)-H_{\log p_t}(\rvy)\|\le L_t\|\rvx-\rvy\|$ gives
\begin{align}
\label{eq:T2_bound}
T_2(\rvx,t)
&=
\frac{1-\bar{\alpha}(t)}{\sqrt{\bar{\alpha}(t)}}\,
\left\Vert
H_{\log p_t}(\rvx)\,v
-
\frac{\nabla_{\rvx}\log p_t(\rvx+\epsilon v)-\nabla_{\rvx}\log p_t(\rvx)}{\epsilon}
\right\Vert \notag\\
&\le
\frac{1-\bar{\alpha}(t)}{\sqrt{\bar{\alpha}(t)}}\,
\frac{1}{\epsilon}\int_{0}^{\epsilon} L_t\,s\,\|v(\rvx,t)\|^{2}\,ds
\le
\frac{(1-\bar{\alpha}(t))\,L_t\,\eta^{2}\,\epsilon}{2\sqrt{\bar{\alpha}(t)}}.
\end{align}

\paragraph{Conclusion.}
Combining (\ref{eq:score_gap_start}), (\ref{eq:T1_final_bound}), and (\ref{eq:T2_bound}), we obtain
\[
\left\Vert g(\rvx,t)-\tilde g^{(1)}(\rvx,t)\right\Vert
\le
\frac{(1-\bar{\alpha}(t))\,L_t\,\eta^{2}\,\epsilon}{2\sqrt{\bar{\alpha}(t)}}
+
\left(
\frac{\gamma_t\eta_2}{2m}
+
\frac{(1+(1-\bar{\alpha}(t))\zeta_t)\,\eta\,\eta_2}{2m\sqrt{\bar{\alpha}(t)}}
\right)
\mathbb{E}\!\left[\left\|\rmX^{p}_{0}-\mu(\rvx)\right\|^{2}\ \Big|\ \rmX^{p}_{t}=\rvx\right],
\]
which matches the definition of $u(\rvx,t)$ in Theorem~\ref{theorem_1}. This completes the proof.
\end{proof}

\subsection{Proof of Proposition \ref{proposition_scheduling}}
\label{proof:proposition_scheduling}
We have the mean squared error risk measure which is given as follows.
\begin{align}
    \label{mse_risk}
    r(t;\tau) &= \mathbb{E}[ \Vert \nabla_{\rmX^{q}} \log {q}_{t}(\rmX^{q}) -\nabla_{\rmX^{q}} \log \tilde{q}_{t}(\rmX^{q}) \Vert^2] =  \mathbb{E}[ \Vert{g}(\rmX^{q},t) - \tau(t)\tilde{g}^{(1)}(\rmX^{q},t) \Vert^2] \\
     &=  (1-\tau(t))^2  \mathbb{E}[\Vert{g}(\rmX^{q},t)\Vert^2] + (\tau(t))^2  \mathbb{E}[\Vert \Delta^{(1)}(\rmX^{q},t)\Vert^2].
\end{align}

Recall the definition of the ground-truth guidance term, 
\begin{align}
\label{def_g_fractional_form}
    g(\rvx,t) = \frac{ \nabla_{\rvx} \mathbb{E}_{\rmX^{p}_{0} \sim p_{\rmX^{p}_{0}|\rmX^{p}_{t}}(\cdot|\rvx)}[w(\rmX^{p}_{0})] }{  \mathbb{E}_{\rmX^{p}_{0} \sim p_{\rmX^{p}_{0}|\rmX^{p}_{t}}(\cdot|\rvx)}[w(\rmX^{p}_{0})] }. 
\end{align}
Then, the numerator of (\ref{def_g_fractional_form}) can be represented as follows.
\begin{align}
  &\nabla_{\rvx} \mathbb{E}_{\rmX^{p}_{0} \sim p_{\rmX^{p}_{0}|\rmX^{p}_{t}}(\cdot|\rvx)}[w(\rmX^{p}_{0})]
   =\int w(\mathbf{y})\,
        \nabla_{\mathbf{x}}
        p_{\rmX^{p}_{0}\mid\rmX^{p}_{t}}(\mathbf{y}\mid\mathbf{x})
        \,d\mathbf{y} \nonumber\\
    &=\int w(\mathbf{y})\,
        p_{\rmX^{p}_{0}\mid\rmX^{p}_{t}}(\mathbf{y}\mid\mathbf{x})\,
        \nabla_{\mathbf{x}}
        \log p_{\rmX^{p}_{0}\mid\rmX^{p}_{t}}(\mathbf{y}\mid\mathbf{x})
        \,d\mathbf{y}
   =\mathbb{E}_{\rmX^{p}_{0}\mid\rmX^{p}_{t}=\mathbf{x}}
      \bigl[w(\rmX^{p}_{0})\,S(\rmX^{p}_{0},\mathbf{x},t)\bigr],
  \label{eq:nabla_f}
\end{align}
where \(S(\rmX^{p}_{0},\mathbf{x},t)
      \triangleq
      \nabla_{\mathbf{x}}\log
      p_{\rmX^{p}_{0}\mid\rmX^{p}_{t}}(\rmX^{p}_{0}\mid\mathbf{x})\)
is the conditional score.

Let
\(h(\mathbf{y},\mathbf{x})
   =w(\mathbf{y})
     p_{\rmX^{p}_{0}\mid\rmX^{p}_{t}}(\mathbf{y}\mid\mathbf{x})\).
Using \(\nabla_{\mathbf{x}}p = p \nabla_{\mathbf{x}}\log p\), we have
\[
  \bigl\Vert\nabla_{\mathbf{x}}h(\mathbf{y},\mathbf{x})\bigr\Vert
  = w(\mathbf{y})\,
     \bigl\Vert\nabla_{\mathbf{x}}
        p_{\rmX^{p}_{0}\mid\rmX^{p}_{t}}(\mathbf{y}\mid\mathbf{x})
     \bigr\Vert
  = w(\mathbf{y})\,p_{\rmX^{p}_{0}\mid\rmX^{p}_{t}}(\mathbf{y}\mid\mathbf{x})
    \bigl\Vert
      \nabla_{\mathbf{x}}\log
      p_{\rmX^{p}_{0}\mid\rmX^{p}_{t}}(\mathbf{y}\mid\mathbf{x})
     \bigr\Vert
  \le \gamma_t\, w(\mathbf{y})\,
        p_{\rmX^{p}_{0}\mid\rmX^{p}_{t}}(\mathbf{y}\mid\mathbf{x}).
\]
Since \(\mathbb{E}[w(\rmX^{p}_{0})\mid \rmX^{p}_{t}=\mathbf{x}]
      = \int w(\mathbf{y})p_{\rmX^{p}_{0}\mid\rmX^{p}_{t}}(\mathbf{y}\mid\mathbf{x})d\mathbf{y}<\infty\)
and \(\gamma_t<\infty\), we obtain
\[
  \int \bigl\Vert\nabla_{\mathbf{x}}h(\mathbf{y},\mathbf{x})\bigr\Vert d\mathbf{y}
  \le \gamma_t\,\mathbb{E}[w(\rmX^{p}_{0})\mid \rmX^{p}_{t}=\mathbf{x}] <\infty.
\]
Therefore, differentiation under the integral sign is justified (Leibniz rule), giving
\(\nabla_{\mathbf{x}}\!\int h = \int\nabla_{\mathbf{x}}h\).

Dividing (\ref{eq:nabla_f}) by $\mathbb{E}[\,w(\rmX^{p}_{0}) \mid \rmX^{p}_{t}=\mathbf{x}\,]$
gives
\begin{align}
\label{eq:g_ratio}
  \Vert g(\mathbf{x},t) \Vert
  \;=\;
    \frac{\Vert \mathbb{E}[w(\rmX^{p}_{0})\,S(\rmX^{p}_{0},\mathbf{x},t)\mid \rmX^{p}_{t}=\mathbf{x}\,] \Vert}
       {\Vert \mathbb{E}[w(\rmX^{p}_{0})\mid \rmX^{p}_{t}=\mathbf{x}\,] \Vert} \le
    \frac{ \mathbb{E}[w(\rmX^{p}_{0})\, \Vert S(\rmX^{p}_{0},\mathbf{x},t)\Vert\mid \rmX^{p}_{t}=\mathbf{x}\,] }
       { \mathbb{E}[w(\rmX^{p}_{0})\mid \rmX^{p}_{t}=\mathbf{x}\,] }
    \le
    \frac{ \gamma_t \mathbb{E}[w(\rmX^{p}_{0})\mid \rmX^{p}_{t}=\mathbf{x}\,] }{ \mathbb{E}[w(\rmX^{p}_{0})\mid \rmX^{p}_{t}=\mathbf{x}\,] }
    = \gamma_t.
\end{align}
By Assumption \ref{assumption_lipschitz}, we have $\Vert g(\mathbf{x},t) \Vert \le \gamma_{t}$.

We now examine $\mathbb{E}[\Vert \Delta^{(1)}(\rmX,t)\Vert^2]$. Recall the definition of the upper bound of the uncertainty $u$ and for all $\rvx$ and $t$, we have
\begin{align}
    \label{upperbound_recall}
    u(\rvx, t) \le \frac{(1-\bar{\alpha}(t))L\eta^{2}\epsilon}{2\sqrt{\bar{\alpha}(t)}}  + \left(\frac{\gamma \eta_{2}}{2m} + \frac{ \left({1+(1-\bar{\alpha}(t))\zeta} \right)\eta\eta_{2} }{2m \sqrt{\bar{\alpha}(t)}} \right) \mathbb{E} \left[\Vert \rmX^{p}_{0}- \bar{\rvx}'_{0}\vert_{\rvx,t}\Vert^2 \vert \rmX^{p}_{t}\!=\!\rvx \right],
\end{align}
where $\gamma = \max_{t\in[0,T]} \gamma_{t}$, $\zeta = \max_{t\in[0,T]} \zeta_{t}$, $L = \max_{t\in[0,T]} L_{t}$. This replacement converts the time-dependent bound in Theorem~\ref{theorem_1} into a time-uniform (and hence looser) deterministic envelope.
The conditional covariance admits the matrix identity
\begin{align}
\mathrm{Cov}(\rmX^{p}_{0}\mid \rmX^{p}_{t}=\rvx)
=
\frac{1-\bar{\alpha}(t)}{\bar{\alpha}(t)}
\Bigl(
\rmI + (1-\bar{\alpha}(t)) H_{\log p_t}(\rvx)
\Bigr).
\end{align}
Hence, it can be represented as \cite{rout2024beyond}
\begin{align}
\label{upperbound_recall_2}
\mathbb{E}\!\left[\|\rmX^{p}_{0}-\bar{\rvx}'_{0}\vert_{\rvx,t}\|^2 \mid \rmX^{p}_{t}=\rvx \right]
&=
\mathrm{Tr}\!\left(\mathrm{Cov}(\rmX^{p}_{0}\mid \rmX^{p}_{t}=\rvx)\right) \\
&=
d\frac{1-\bar{\alpha}(t)}{\bar{\alpha}(t)}
+
\frac{(1-\bar{\alpha}(t))^2}{\bar{\alpha}(t)}
\mathrm{Tr}(H_{\log p_t}(\rvx)).
\end{align}
Since $\mathrm{Cov}(\rmX^{p}_{0}\mid \rmX^{p}_{t}=\rvx)\succeq 0$ and
$\frac{1-\bar{\alpha}(t)}{\bar{\alpha}(t)}>0$, it follows that
\begin{align}
\rmI+(1-\bar{\alpha}(t))H_{\log p_t}(\rvx)\succeq 0,
\end{align}
equivalently,
\begin{align}
\lambda_{\min}(H_{\log p_t}(\rvx))\ge -(1-\bar{\alpha}(t))^{-1}.
\end{align}

Under the assumptions $\Vert H_{\log p_{t}}(\rvx) \Vert \le \zeta$, $\mathrm{Tr}(H_{\log p_{t}}(\rvx)) \le d \zeta$, and substituting (\ref{upperbound_recall_2}) to the upper bound in (\ref{upperbound_recall}), we have $u(\rvx, t) \le u'(t)$ where
\begin{align}
\label{noise_upper_bound}
u'(t)=
\frac{L\eta^{2}\epsilon}{2}\,\frac{1-\bar{\alpha}(t)}{\bar{\alpha}(t)^{1/2}}
+\frac{d(\gamma\eta_{2})}{2m}\,
 \frac{1-\bar{\alpha}(t)}{\bar{\alpha}(t)}
+\frac{d\zeta(\gamma\eta_{2})}{2m}\,
 \frac{(1-\bar{\alpha}(t))^{2}}{\bar{\alpha}(t)} \nonumber \\
 \quad+\frac{d\eta\eta_{2}}{2m}\,
 \frac{1-\bar{\alpha}(t)}{\bar{\alpha}(t)^{3/2}}
+\frac{d\zeta\eta\eta_{2}}{m}\,
 \frac{(1-\bar{\alpha}(t))^{2}}{\bar{\alpha}(t)^{3/2}}
+\frac{d\zeta^{2}\eta\eta_{2}}{2m}\,
 \frac{(1-\bar{\alpha}(t))^{3}}{\bar{\alpha}(t)^{3/2}}.
\end{align}

Define \(\gamma \coloneqq \max_{t\in[0,T]} \gamma_t\). Set \(c_1 \coloneqq \gamma^2\).
Since \(\Vert g(\mathbf{x},t)\Vert \le \gamma_t\) for all \(\mathbf{x}\),
we have \(\mathbb{E}[\Vert g(\rmX^{q}_{t},t)\Vert^{2}] \le \gamma_t^{2} \le \gamma^{2}\).
Moreover, \(\Vert \Delta^{(1)}(\rvx,t)\Vert \le u(\rvx,t) \le u'(t)\) implies
\(\mathbb{E}[\Vert \Delta^{(1)}(\rmX^{q}_{t},t)\Vert^{2}] \le (u'(t))^{2}\).
Therefore,
\begin{align}
\label{risk_upper_bound}
r(t;\tau) \le r'(t;\tau)
\coloneqq (1-\tau(t))^2  \gamma^{2} + (\tau(t))^2  (u'(t))^{2}.
\end{align}

For each fixed \(t\), \(r'(t;\tau)\) is a strictly convex quadratic in \(\tau\), and expanding~(\ref{risk_upper_bound}) gives
\[
r'(t;\tau)=\gamma^2-2\gamma^2\tau+\bigl(\gamma^2+(u'(t))^2\bigr)\tau^2.
\]
Hence its unique minimizer over \(\tau\in\mathbb{R}\) is
\begin{equation}
\label{eq:tau_star_exact}
\tau^*(t)=\frac{\gamma^2}{\gamma^2+(u'(t))^2}.
\end{equation}

It remains to write \((u'(t))^2\) in closed form.  Define \(c_1\coloneqq \gamma^2\) and rewrite~(\ref{noise_upper_bound}) as
\begin{equation}
\label{eq:uprime_compact}
u'(t)=\sum_{k=1}^{6} A_k\,\phi_k(t),\qquad
\phi_k(t)=\frac{(1-\bar\alpha(t))^{i_k}}{\bar\alpha(t)^{j_k}},
\end{equation}
with exponent pairs
\[
(i_k,j_k)\in\Bigl\{(1,\tfrac12),(1,1),(2,1),(1,\tfrac32),(2,\tfrac32),(3,\tfrac32)\Bigr\},
\]
and coefficients
\[
\begin{aligned}
A_1&=\frac{L\eta^2\epsilon}{2},&
A_2&=\frac{d\,\gamma\eta_2}{2m},&
A_3&=\frac{d\,\zeta\,\gamma\eta_2}{2m},&
A_4&=\frac{d\eta\eta_2}{2m},&
A_5&=\frac{d\zeta\eta\eta_2}{m},&
A_6&=\frac{d\zeta^2\eta\eta_2}{2m}.
\end{aligned}
\]
Therefore, \((u'(t))^2\) expands exactly as
\begin{equation}
\label{eq:uprime_square_exact}
(u'(t))^2
=\sum_{k=1}^{6}\sum_{\ell=1}^{6} A_kA_\ell\,
\frac{(1-\bar\alpha(t))^{i_k+i_\ell}}{\bar\alpha(t)^{j_k+j_\ell}}
=\sum_{(i,j)\in\mathcal I} c_{(i,j)}\,
\frac{(1-\bar\alpha(t))^{i}}{\bar\alpha(t)^{j}},
\end{equation}
where the (finite) index set is
\[
\mathcal I=
\Bigl\{(2,1),(2,\tfrac32),(2,2),(2,\tfrac52),(2,3),
(3,\tfrac32),(3,2),(3,\tfrac52),(3,3),
(4,2),(4,\tfrac52),(4,3),
(5,\tfrac52),(5,3),(6,3)\Bigr\},
\]
and the coefficients are obtained by collecting equal exponent pairs:
\[
\begin{aligned}
c_{(2,1)}&=A_1^2, \qquad
c_{(2,\tfrac32)}=2A_1A_2, \qquad
c_{(2,2)}=A_2^2+2A_1A_4,\\
c_{(2,\tfrac52)}&=2A_2A_4, \qquad
c_{(2,3)}=A_4^2, \qquad
c_{(3,\tfrac32)}=2A_1A_3,\\
c_{(3,2)}&=2A_1A_5+2A_2A_3, \qquad
c_{(3,\tfrac52)}=2A_2A_5+2A_3A_4, \qquad
c_{(3,3)}=2A_4A_5,\\
c_{(4,2)}&=A_3^2+2A_1A_6, \qquad
c_{(4,\tfrac52)}=2A_2A_6+2A_3A_5, \qquad
c_{(4,3)}=A_5^2+2A_4A_6,\\
c_{(5,\tfrac52)}&=2A_3A_6, \qquad
c_{(5,3)}=2A_5A_6, \qquad
c_{(6,3)}=A_6^2.
\end{aligned}
\]
Substituting~(\ref{eq:uprime_square_exact}) into~(\ref{eq:tau_star_exact}) yields
\[
\tau^*(t)
=\frac{c_1}{c_1+\sum_{(i,j)\in\mathcal I} c_{(i,j)}
\frac{(1-\bar\alpha(t))^{i}}{\bar\alpha(t)^{j}}}
=\left(1+\sum_{(i,j)\in\mathcal I}\frac{c_{(i,j)}}{c_1}
\frac{(1-\bar\alpha(t))^{i}}{\bar\alpha(t)^{j}}\right)^{-1}.
\]

\begin{figure}[htbp]
    \centering
    \includegraphics[width=0.5\columnwidth]    {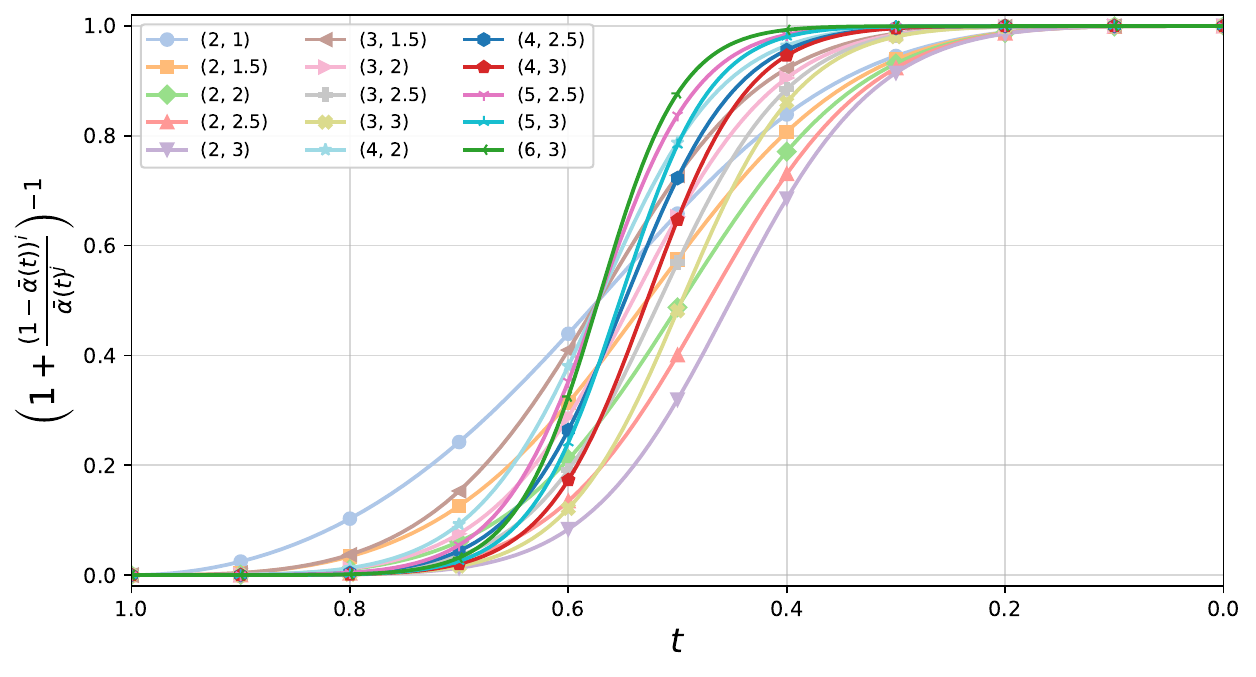}
     \caption{Sigmoid-shaped curves defined by $\left(1 + \tfrac{(1 - \bar{\alpha})^i}{\bar{\alpha}^j} \right)^{-1}$ for various $(i,j) \in \mathcal{I}$, under the cosine-based cumulative schedule $\bar{\alpha}(t)$. For practical implementation, we approximate the behavior of $\tau^*(t)$ using a single-parameter surrogate.}
    \label{fig:scheduler}
\end{figure}

\paragraph{Approximation of $\tau^*(t)$.}
Figure~\ref{fig:scheduler} illustrates the behavior of the function
$
\left(1 + \frac{(1 - \bar{\alpha}(t))^i}{\bar{\alpha}(t)^j} \right)^{-1},
$
for various index pairs $(i, j) \in \mathcal{I}$, under the cosine-based variance schedule. Here, $\bar{\alpha}(t)$ denotes the cumulative product of a scheduling function applied to diffusion coefficients, as typically derived from variance-preserving score-based generative models.

It is noteworthy that, due to the decreasing nature of the term $\frac{(1 - \bar{\alpha}(t))^i}{\bar{\alpha}(t)^j}$ in $t$, the composite function above exhibits a monotonic increase over time. Rather than precisely estimating the weight coefficients $c_{(i,j)}$ for each index pair $(i, j)$, we propose to approximate the overall effect with a single representative parameter. This simplification is motivated by the empirical observation that the variation across terms in $\mathcal{I}$ often follows a coherent monotonic trend that can be effectively captured using a single-parameter approximation.

\paragraph{Scheduling without Assumption \ref{assumption_uncorrelated}.}
Recall the mean squared error between the score functions $\nabla_{\rmX} \log q_t(\rmX^{q})$ and $\nabla_{\rmX} \log \tilde{q}_t(\rmX^{q})$, which is expressed as
\begin{align}
    \label{mse_risk_without_assumption}
    &r(t;\tau) = \mathbb{E}[ \Vert \nabla_{\rmX} \log {q}_{t}(\rmX^{q}) -\nabla_{\rmX^{q}} \log \tilde{q}_{t}(\rmX^{q}) \Vert^2] =  \mathbb{E}[ \Vert{g}(\rmX^{q},t) - \tau(t)\tilde{g}^{(1)}(\rmX^{q},t) \Vert^2] \\
     &=  (1-\tau(t))^2  \mathbb{E}[\Vert{g}(\rmX^{q},t)\Vert^2] + \mathbb{E}[2(1\!-\!\tau(t))\tau(t) g(\rmX^{q},t) ^{\!\top} \!\!\Delta^{\!(1)}(\rmX^{q},t)] + (\tau(t))^2  \mathbb{E}[\Vert \Delta^{(1)}(\rmX^{q},t)\Vert^2]
\end{align}

Applying the Cauchy–Schwarz inequality and upper bounds on the second and third terms yields
\begin{align}
     &r(t;\tau) \le (1-\tau(t))^2  \gamma^2 +  2(1-\tau(t))\tau(t) \sqrt{\mathbb{E}[ \Vert g(\rmX^{q},t) \Vert^2 ]} \sqrt{ \mathbb{E} [\Vert \Delta^{(1)}(\rmX^{q}, t) \Vert^2]}+ (\tau(t))^2  (u'(t))^2 \nonumber \\
     &\le (1-\tau(t))^2  \gamma^2 +  2(1-\tau(t))\tau(t) \gamma u'(t)+ (\tau(t))^2  (u'(t))^2\\
&= r_{\text{ub}}(t;\tau) \coloneqq \gamma^{2}+2\gamma \bigl(u'(t)-\gamma\bigr)\tau+\bigl(\gamma-u'(t)\bigr)^{2}\tau^{2}.
\end{align}
Here, we denote $\tau(t)$ by $\tau$ since $t$ is fixed. The upper bound $r_{\mathrm{ub}}(t;\tau)$ is a quadratic function of $\tau$ with a non-negative leading coefficient $(\gamma - u'(t))^2$, and hence is convex.

We now examine the minimizer of $r_{\mathrm{ub}}(t;\tau)$ over $\tau \in [0,1]$. Taking derivative yields
$$
\frac{\partial r_{\text{ub}}(t;\tau)}{\partial\tau}=2\gamma \bigl(u'(t)-\gamma\bigr)+2\bigl(\gamma-u'(t)\bigr)^{2}\tau
$$
which vanishes at
$$
\tau_{\mathrm{crit}}=\frac{\gamma}{\gamma-u'(t)}.
$$
\smallskip
\noindent%
\textbf{Case $u'(t)<\gamma$.}  Then $\gamma-u'(t)>0$, which implies $\tau_{\mathrm{crit}}>1$.  Convexity forces the minimum on $[0,1]$ to occur at the right endpoint, that is, $\tau^*(t)=1$.  

\smallskip
\noindent%
\textbf{Case $u'(t)>\gamma$.}  Now $\gamma-u'(t)<0$, hence $\tau_{\mathrm{crit}}<0$.  The minimum is attained at the left endpoint, $\tau^*(t)=0$.  

\smallskip
\noindent%
\textbf{Case $u'(t)=\gamma$.} Every $\tau$ in the interval therefore minimizes the function.  

Importantly, the scheduler $\tau^*(t)$ is monotonic in $t$ due to the behavior of $u'(t)$, which denotes the upper bound of uncertainty as defined in~(\ref{noise_upper_bound}). As $t \to 0$, we have $u'(t) \to 0$ and thus $u'(t) < \gamma$ for any $\gamma > 0$, leading to $\tau^*(t) = 1$. Conversely, at large $t$, the uncertainty increases and $u'(t) > \gamma$, resulting in $\tau^*(t) = 0$.
This implies that there exists a critical threshold $t' \in [0, T]$ such that $\tau^*(t)$ behaves as a step function, switching from $1$ to $0$ at $t = t'$. In our experiments (Appendices~\ref{appendix_experiments_details_celeba}–\ref{appendix_experiments_details_stablecascade}), we adopt this scheduler using a fixed threshold of $t' = 0.7T$ for heuristic estimation of the strategy.

\subsection{Discretization of the SDE for weighted sampling}
\label{appendix_sde_disrectization}
To generate samples following the weighted sampling PDF $q$, the proposed SDE can be implemented via discretization methods. For example, the Euler-Maruyama scheme discretizes the time interval $[0,T]$ with step size $\Delta t$, allowing iterative solutions to the SDE. In this section, we consider $T \in \mathbb{N}$ (a positive integer) and adopt a unit time discretization, i.e., $\Delta t = 1$. Recall the proposed SDE which is given as follows.
\begin{align}
\label{propoed_sde_to_be_discretized}
     &\mathrm{d} \rmX^{\tilde{q}}_{t} = -\frac{\beta(t)}{2} \left[  \rmX^{\tilde{q}}_{t} +2  \nabla_{\rmX^{\tilde{q}}_{t}} \log \tilde{q}_{t}(\rmX^{\tilde{q}}_{t}) \right]\, \mathrm{d}t + \sqrt{\beta(t )} \, \mathrm{d} \tilde{\rmW}_{t}.
 \end{align}

Let $\{\beta(t)\}_{t\in[0,T]}$ denote the variance schedule of the forward diffusion
process, and define the discrete instants
$
t_k = \frac{k}{N}T,\; k=0,\dots,N,
$
with unit–step discretisation $\Delta t = T/N =1$ for ease of exposition.
Denote by $\rvx_k \triangleq \rvx^{\tilde{q}}_{t_k}$ the discretised state.
Starting from the reverse–time SDE proposed in~(\ref{propoed_sde_to_be_discretized}),
the Euler–Maruyama scheme yields
\begin{align}
\label{eq:euler}
\rvx_{k-1}
&= \rvx_k
+ \frac{\beta_k}{2}\bigl(\rvx_k + 2\nabla_{\rvx_k}\log\tilde{q}_{t_k}(\rvx_k)\bigr)
+ \sqrt{\beta_k}\,\boldsymbol{\xi}_k,
\qquad
\boldsymbol{\xi}_k\sim\mathcal{N}(\mathbf{0},\mathbf{I}),
\end{align}
where $\beta_k = \beta(t_k)$.
To connect~(\ref{eq:euler}) to the standard diffusion notation,
define the single–step and cumulative attenuation factors
$
\alpha_k = 1-\beta_k$, and $
\bar{\alpha}_k = \prod_{j=1}^{k}\alpha_j.
$
Using the representation of the SGM, we denote the denoising model with the following representation
$
\nabla_{\rvx_k}\log\tilde{q}_{t_k}(\rvx_k)
\approx -\frac{1}{\sqrt{1-\bar{\alpha}_k}}\,
\hat{\epsilon}_\theta(\rvx_k,k)
$. Based on these, we adopt the discrete DDPM \cite{ho2020ddpm} update
\begin{align}
\label{eq:ddpm}
\rvx_{k-1}
&=
\frac{\sqrt{\bar{\alpha}_{k-1}}}{\sqrt{\bar{\alpha}_k}}
\bigl(\rvx_k
-\sqrt{1-\bar{\alpha}_k}\,
\hat{\epsilon}_\theta(\rvx_k,k)\Bigr)
+
\sqrt{1-\bar{\alpha}_{k-1}-\sigma_k^{2}}\,
\hat{\epsilon}_\theta(\rvx_k,k)
+
\sigma_k\boldsymbol{\xi}_k,
\end{align}
with $\sigma_k\in[0,1]$ the noise level, $\sigma_k=\eta_{s} \sqrt{(1-\bar{\alpha}_{k-1})(1-\alpha_k)/\,(1-\bar{\alpha}_k)}$, $\eta_{s}$ the parameter controlling the stochasticity.
When we use DDIM, we have $\eta_{s}=0$ in~(\ref{eq:ddpm}) suppressing the stochastic component and
transforms~(\ref{eq:ddpm}) into the following update rule
\begin{align}
\label{eq:ddim}
\rvx_{k-1}
=
\sqrt{\bar{\alpha}_{k-1}}\,
\hat{\mathbf{x}}_{0,k}
+
\sqrt{1-\bar{\alpha}_{k-1}}\,
\hat{\epsilon}_\theta(\rvx_k,k),
\quad
\hat{\mathbf{x}}_{0,k}
=
\frac{\rvx_k-\sqrt{1-\bar{\alpha}_k}\,\hat{\epsilon}_\theta(\rvx_k,k)}
{\sqrt{\bar{\alpha}_k}}.
\end{align}

For experiments in Appendix \ref{appendix_experiments_toy_examples}, \ref{appendix_experiments_details_text_image_alignment}, and \ref{appendix_experiments_details_stablecascade}, we adopt DDIM sampling. For those in Appendix \ref{appendix_experiments_details_mnist}, we use DDPM sampling.

\subsection{Relation to Existing Methods}
\label{appendix_relation_to_second_order}
In the main text, we adopt a first-order Taylor approximation of the weighted score function based on the conditional mean of the initial sample:
\begin{align}
    \nabla_{\rvx} \log q_{t}(\rvx) 
    &= \nabla_{\rvx} \log p_{t}(\rvx) 
    + \nabla_{\rvx} \log \mathbb{E}_{\rmX^{p}_{0} \sim p(\cdot \mid \rvx)}[w(\rmX^{p}_{0})] \\
    &\approx \nabla_{\rvx} \log p_{t}(\rvx) + \nabla_{\rvx} \log w(\bar{\rvx}'_{0} \vert_{\rvx,t}),
\end{align}
where $\bar{\rvx}'_{0} \vert_{\rvx,t}$ denotes the conditional expectation of $\rmX^{p}_{0}$ given $\rvx$ under the reference diffusion process.

This approximation is closely related to the denoising-based posterior sampling technique introduced in~\cite{chungdiffusion, kim2025das}, which was originally proposed for solving inverse problems via diffusion guidance. Consider a binary random variable $C \in \{0,1\}$ such that $p(C=1 \mid \rvx) \propto w(\rvx)$. Then, by Bayes' rule, the posterior over $\rvx$ conditioned on $C=1$ becomes
\begin{align}
    p(\rvx \mid C = 1) = \frac{p(\rvx)\, p(C = 1 \mid \rvx)}{p(C = 1)} \propto p(\rvx)\, w(\rvx),
\end{align}
yielding the desired reweighted distribution. Taking the gradient of the log posterior gives
\begin{align}
    \nabla_{\rvx} \log p_{t}(\rvx \mid C = 1) 
    = \nabla_{\rvx} \log p_{t}(\rvx) + \nabla_{\rvx} \log w(\rvx).
\end{align}
Following~\cite{chungdiffusion}, one can replace $\rvx$ in the second term with the estimated conditional mean $\bar{\rvx}'_{0} \vert_{\rvx,t}$, resulting in the approximation
\[
\nabla_{\rvx} \log w(\rvx) \approx \nabla_{\rvx} \log w(\bar{\rvx}'_{0} \vert_{\rvx,t}).
\]
This formulation has been well investigated in~\cite{kim2025das}.

Compared to existing methods~\cite{chungdiffusion, yu2023freedom, bansal2023universal, song2023loss, kim2025das}, which compute the full gradient $\nabla_{\rvx} \log w(\bar{\rvx}'_{0} \vert_{\rvx,t})$ by backpropagating through the score network, our approach avoids differentiating through the composition of the weight function and the score model. This distinction reduces computational overhead, especially when the score function is parameterized by a large-scale neural network.

Specifically, we approximate the directional update using a finite-difference estimate of the Hessian-vector product, requiring only one gradient evaluation of the task-specific loss. This results in substantially lower computational cost, particularly in settings where the guidance model is significantly larger than the task model, as demonstrated in Sec.~\ref{sec:experiments}.

\section{Implementations and Experiments}
\label{appendix_all_experiments}

\subsection{Experiments with Synthetic Datasets}
\label{appendix_experiments_toy_examples}

\paragraph{Base density.} We evaluate the proposed method and baselines on a 2D weighted sampling task where the base distribution is multimodal, and the importance weight concentrates along a heart-shaped manifold. Specifically, the base distribution is a mixture of 25 isotropic Gaussians with variance $0.2$, arranged on a $5 \times 5$ grid over $[-4, 4]^2$. The score model is a 2-layer U-Net trained via DDIM \cite{song2021ddim}.

For clarity, we slightly abuse notation in this subsection; symbols are local to this subsection.

\paragraph{Heart shape.}
The heart-shaped importance region is defined by the parametric curve $u(\phi) \in \mathbb{R}^2$, where 
$u(\phi) = \left[ 16 \sin^3 \phi,\ 13 \cos \phi - 5 \cos(2\phi) - 2 \cos(3\phi) - \cos(4\phi) \right]^\top$ 
for $\phi \in [0, 2\pi]$. We discretize the curve into $N = 1000$ points as $\phi_n = 2\pi(n-1)/(N-1)$, $n = 1, \dots, N$, and let $\mathcal{U} = \{ u(\phi_n) \}_{n=1}^{N}$. The importance weight is defined as  
$w(\rvx) = \exp\left( - \frac{1}{s} \min_n \| \rvx - u_n \|^2 \right)$,  
where $u_n \in \mathcal{U}$ and $s = 0.05$. The corresponding target distribution is $q(\rvx) \propto w(\rvx) p(\rvx)$.

\begin{figure}[t]
    \centering
    \includegraphics[width=1.0\linewidth]{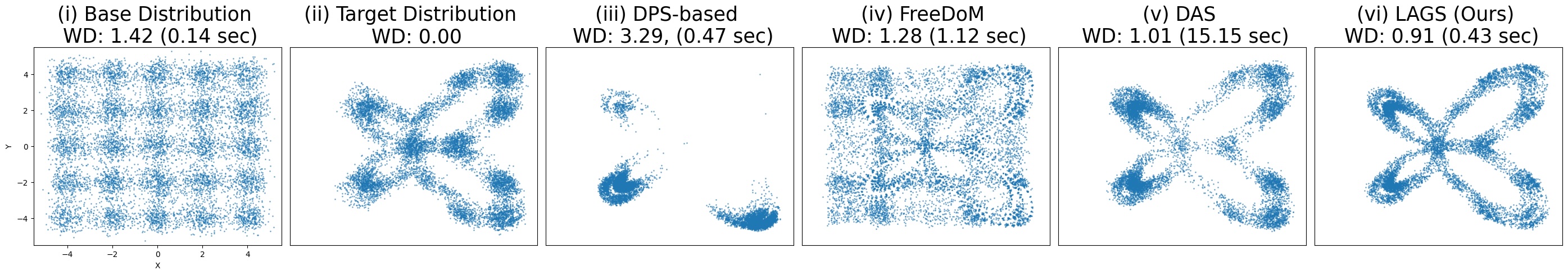}
    \includegraphics[width=1.0\linewidth]{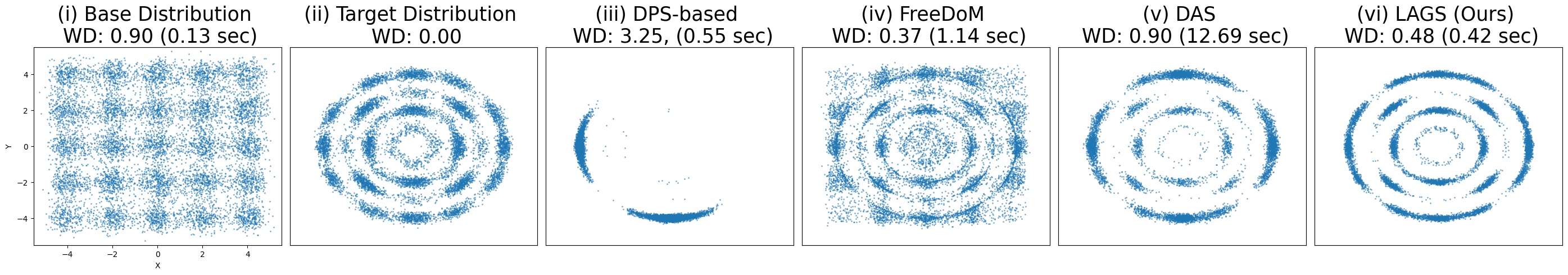}
    \includegraphics[width=1.0\linewidth]{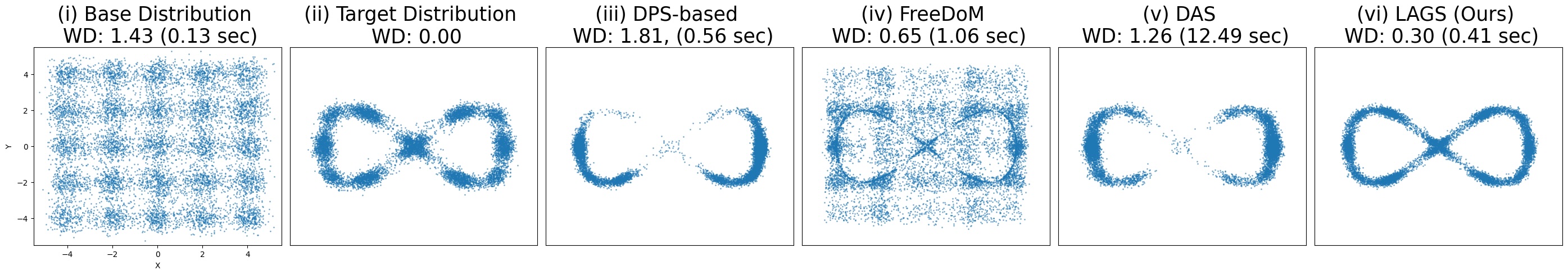}
    \caption{
    \textbf{\emph{Left} to \emph{Right}}: Base distribution, target distribution, and estimated sample densities. Wasserstein Distance and runtime (in seconds) are reported for DPS, FreeDoM, DAS, and LAGS (ours). Our method achieves the \textbf{\emph{lowest Wasserstein Distance (WD) and fastest runtime}}.
    }
    \label{fig:gm_appendix}
\end{figure}

\paragraph{Butterfly shape.}
For the experiment setup of the top row in Figure~\ref{fig:gm_appendix}, the butterfly-shaped importance region is defined by the parametric curve
\[
u(\phi) = a \cdot \rho(\phi)
\begin{bmatrix}
\cos \phi \\
\sin \phi
\end{bmatrix}, \quad
\rho(\phi) = \exp(\cos \phi) - 2 \cos(4\phi) + \sin^5\left( \frac{\phi}{12} \right),
\]
for $\phi \in [0, 12\pi]$. We set $a = 1.2$ and discretize the curve into $N = 1000$ points: $\phi_n = 12\pi(n-1)/(N-1)$, $n = 1, \dots, N$. Let $\mathcal{U} = \{ u(\phi_n) \}_{n=1}^{N}$. The importance weight is defined as  
$w(\rvx) = \exp\left( - \frac{1}{s} \min_n \| \rvx - u_n \|^2 \right)$,  
with $s = 0.08$.

\paragraph{Concentric rings.}
For the experiment setup of the second row in Figure~\ref{fig:gm_appendix}, the ring-shaped importance region consists of $K = 4$ concentric circles centered at the origin. The radius of the $k$-th ring is $\delta_k = k \cdot \delta$ with $\delta = 1.0$ and $k = 1, \dots, K$. Each ring is parameterized as $u_k(\phi) = \left[ \delta_k \cos \phi,\ \delta_k \sin \phi \right]^\top$ for $\phi \in [0, 2\pi]$ and discretized into $N = 300$ points. Let $\mathcal{U} = \bigcup_{k=1}^{K} \{ u_k(\phi_n) \}_{n=1}^{N}$. The importance weight is defined as  
$w(\rvx) = \exp\left( - \frac{1}{s} \min_n \| \rvx - u_n \|^2 \right)$,  
with $s = 0.05$.

\paragraph{Infinity shape.}
For the experiment setup of the bottom row in Figure~\ref{fig:gm_appendix}, the infinity-shaped importance region is defined by 
$u(\phi) = \left[ \sin \phi,\ \sin \phi \cos \phi \right]^\top$ 
for $\phi \in [0, 2\pi]$. We set $a = 4.0$ and discretize the curve into $N = 1000$ points as $u_n = a \cdot u(\phi_n)$, where $\phi_n = 2\pi(n-1)/(N-1)$ for $n = 1, \dots, N$. Let $\mathcal{U} = \{ u_n \}_{n=1}^{N}$. The importance weight is defined as  
$w(\rvx) = \exp\left( - \frac{1}{s} \min_n \| \rvx - u_n \|^2 \right)$,  
with $s = 0.1$.

\paragraph{Generative model.}
We consider a lightweight score-based network with (64,64,2) dense layers with ReLU activations for the first two layers. We employ a standard linear schedule for the forward diffusion process:
$\beta_{k} = \beta_{\text{start}} + \frac{k}{K - 1} (\beta_{\text{end}} - \beta_{\text{start}})$,
for $k = 0, \ldots, K-1$, with $\beta_{\text{start}} = 10^{-4}$ and $\beta_{\text{end}} = 0.02$. We adopt the DDIM sampler \cite{song2021ddim} and the optimization is performed over $10^4$ epochs using the Adam optimizer. The number of forward diffusion steps is set to $10^3$. For inference, we subsample $10^2$ timesteps for accelerated generation.

\paragraph{Sampling from the Optimal Weighted Sampling Distribution.} When the true distribution $p(\rvx)$ is explicitly known and sampling from the base probability density function is possible, we can construct a weighted sampling distribution $q(\rvx)$. This is feasible under the assumption that there exists a known constant $M$ satisfying  $    w(\rvx) \leq M, \quad \forall \rvx \in \mathcal{X}$.

Under these conditions, weighted sampling can be performed via the acceptance-rejection method as follows:
\begin{enumerate}
    \item {Sampling from $p(\rvx)$:} Draw $\rvx \sim p(\rvx)$.
    \item {Computing Acceptance Probability:} Compute  
        $a(\rvx) = \frac{w(\rvx)}{M}$,    ensuring $0 \leq a(\rvx) \leq 1$.
    \item {Acceptance/Rejection Step:} Draw $u \sim \mathcal{U}(0,1)$. Accept $\rvx$ if $u \leq a(\rvx)$; otherwise, reject and repeat.
\end{enumerate}

This acceptance-rejection method for weighted sampling is infeasible in high-dimensional scenarios in general. Specifically, in cases such as high-resolution image generation, the exact domain $\mathcal{X}$ may be unknown, or the acceptance probability $a(\rvx)$ may be extremely small, leading to prohibitively high computational costs. 

However, this method remains feasible for the 2D synthetic datasets considered in this section, where the exact domain, base distribution, and weight function are known in closed form. This controlled setting allows us to quantitatively measure the accuracy of weighted sampling and serves as a benchmark for evaluating training-free importance weighting methods.

We evaluate performance using the Wasserstein Distance between the generated samples and the ground-truth target density, as well as the wall-clock time to generate $10^4$ samples.

\subsection{Experiments of Text-Image Alignment Maximization}
\label{appendix_experiments_details_text_image_alignment}

\subsubsection{Baseline Implementation}
\label{baseline_implementation_text_image_alignment}
We implement several competitive baselines under a unified evaluation framework. DPS method is adapted by directly applying posterior sampling to the weighted sampling formulation, following the derivation presented in Appendix~\ref{appendix_relation_to_second_order}, consistent with the approach in \cite{chungdiffusion}. The guidance control mechanism is integrated based on the FreeDoM framework proposed in \cite{yu2023freedom}, but \emph{without any resampling} steps;
FreeDoM baseline is used as released, with no modifications, and follows the official hyperparameter setting—including learning rate control—as described in Appendix 2-(3) of \cite{yu2023freedom}; For DAS \cite{kim2025das}, we employ the released implementation and settings designed for Stable Diffusion, without modification. To further assess computational efficiency, we also use a variant denoted as DAS-1p, where the number of particles is set to 1 and sampling is executed using a single batch. This represents the fastest and most lightweight version of DAS.

\paragraph{Fairness.}
To ensure a fair comparison across all methods, the weight function is fixed to $8 \times \text{PickScore}$, following the scaling scheme used in \cite{kim2025das}. Furthermore, the backward diffusion process is consistently configured to use $100$ discrete timesteps for all baselines.
This facilitates an evaluation by isolating the effect of each algorithm's guidance strategy. Note that the guidance methods differ in how they approximate the true importance weight or posterior score function, and then they are all constrained to operate under a shared target backward diffusion process. 

For evaluations on SD, we adopt prompts used in \cite{kim2025das}. For SDXL experiments, we construct a comprehensive prompt set by merging samples from \cite{wu2023better} and \cite{kim2025das}, along with additional manually curated prompts designed to test generalization to diverse semantic configurations.

\subsubsection{Additional Results}
\label{appendix_text_image_alignment_additional_results}

\paragraph{Weight--diversity trade-off.}
We examine the trade-off between weight, i.e., reward optimization and sample diversity, focusing on whether weighted sampling degrades the diversity of generated images. Diversity is quantified using the CLIP-based mean pairwise distance, where higher values indicate greater diversity.

\begin{table}[t]
\centering
\scriptsize
\caption{CLIP-based diversity (Mean Pairwise Distance). Higher is better. Based on Fig.~\ref{fig:performance_sd}.}
\begin{tabular}{c c c c c c}
\hline
SD & DPS & FreeDoM & DAS & DAS-1P & LAGS (Ours) \\
\hline
0.406 & 0.375 & 0.387 & \textbf{0.424} & 0.360 & 0.381 \\
\hline
\end{tabular}
\label{tab:clip_diversity_horizontal}
\end{table}

As shown in Table~\ref{tab:clip_diversity_horizontal}, most guidance-based methods incur a reduction in diversity relative to the base model, SD. DAS is a notable exception, achieving the highest diversity, likely due to its multi-particle resampling at each diffusion step.

LAGS preserves diversity with a $6.2\%$ relative decrease compared to SD, while still improving reward. These results indicate that LAGS achieves a favorable reward--diversity trade-off without relying on costly multi-particle resampling.

\paragraph{Weight-realism tradeoff.}
Weighted sampling with human preference scores may exhibit a potential tradeoff between \emph{alignment}, i.e., elevating samples that maximize the target weight function, and quality measures, often translated into realism or perceptual naturalness. When the guidance signal is strong, either due to a sharp curvature of the weight function or large scheduler coefficients, the resulting refinement can occasionally over-amplify semantically meaningful but visually brittle directions, leading to reduced image naturalness. This phenomenon has also been reported in prior analyses of score amplification in diffusion models.

Our method already achieves the highest target metric (PickScore) with the fastest inference time; nevertheless, it is desirable to examine whether realism can be further improved without compromising alignment much. To this end, we adopt a lightweight adaptation of the \emph{Adaptive Projected Guidance} (APG) mechanism proposed by Sadat et al. 2025 \cite{sadat2025eliminating}, originally developed to suppress over-saturation artifacts in diffusion guidance. 

Consider the deterministic DDIM update and its corresponding drift direction as
\begin{equation}
    \boldsymbol{\mu}_k
    =
    \sqrt{\bar{\alpha}_{k-1}}\,
    \hat{\mathbf{x}}_{0,k}
    +
    \sqrt{1-\bar{\alpha}_{k-1}}\,
    \hat{\boldsymbol{\epsilon}}_k, \qquad
    \mathbf{u}_k
    =
    \frac{\boldsymbol{\mu}_k - \rvx_k}
         {\lVert \boldsymbol{\mu}_k - \rvx_k\rVert_2}.
\end{equation}
Based on this, we can decompose the guidance into a component parallel to this direction and a component orthogonal to it as
\begin{equation}
\label{apg_formula}
    \mathbf{g}_k^{\parallel}
    =
    \frac{\langle \tilde{g}^{(1)}(\rvx_k,t_k),\, \mathbf{u}_k\rangle}
         {\lVert \mathbf{u}_k\rVert_2^2 }\,
    \mathbf{u}_k,
    \qquad
    \mathbf{g}_k^{\perp}
    =
    \tilde{g}^{(1)}(\rvx_k,t_k) - \mathbf{g}_k^{\parallel}, \qquad
    \hat{\mathbf{g}}_k
    =
    \mathbf{g}_k^{\perp}
    +
    \varsigma\,\mathbf{g}_k^{\parallel}.
\end{equation}
We set $\varsigma\!=\!0$ and the computational overhead of this projection and scaling is negligible with below $0.5\%$ of sampling time.

\paragraph{Results.}
\begin{figure*}[t]
    \centering
    \includegraphics[width=1.0\linewidth]{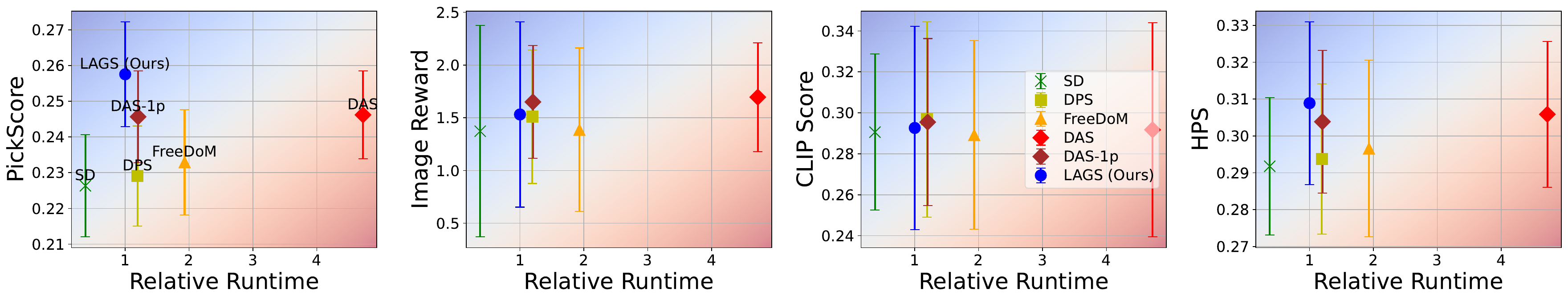}
    \vspace{-7mm}
    \caption{
    \textbf{Results on SDXL}: Evaluation on the same metrics as in Figure~\ref{fig:performance_sd}. LAGS is implemented with the projected guidance (\ref{apg_formula}).
    }
    \label{fig:performance_sdxl_apg}
\end{figure*}
Using $c\!=\!10$, we compare LAGS against existing guidance methods in
Figure~\ref{fig:performance_sdxl_apg}.  
LAGS continues to achieve the highest PickScore and HPS among all methods, while
maintaining competitive ImageReward and CLIP Score.  
Importantly, LAGS remains on the Pareto frontier of \emph{both} alignment (PickScore, HPS) and
efficiency: it simultaneously attains the strongest alignment performance and the lowest runtime
among all guidance-based sampling approaches.

\begin{figure}[t]
    \centering
    \includegraphics[width=0.24\linewidth]{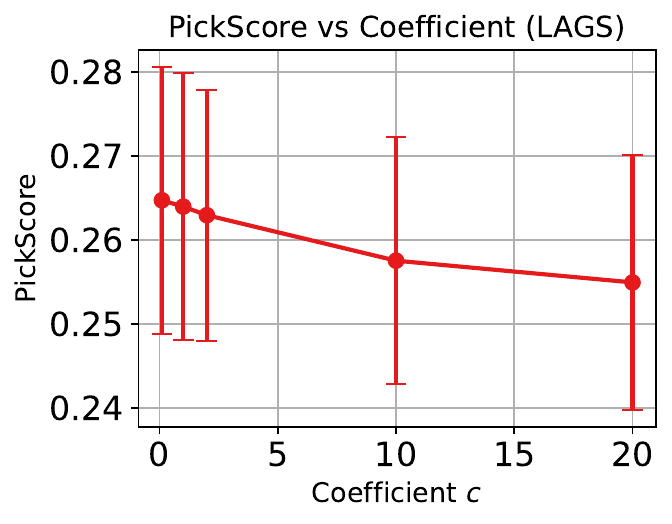}
    \includegraphics[width=0.24\linewidth]{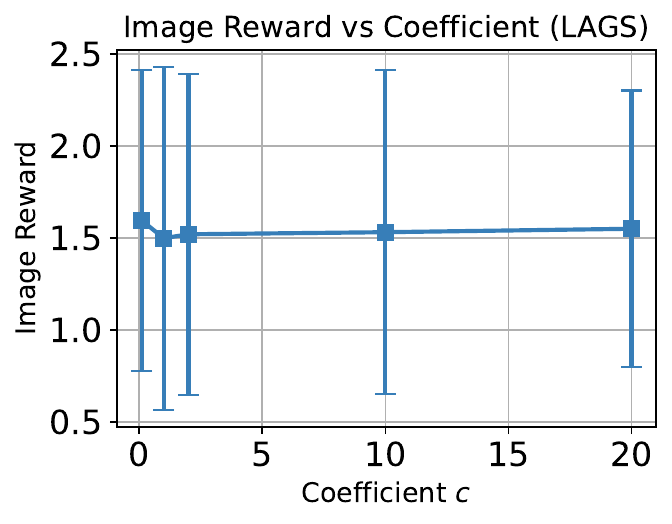}
    \includegraphics[width=0.24\linewidth]{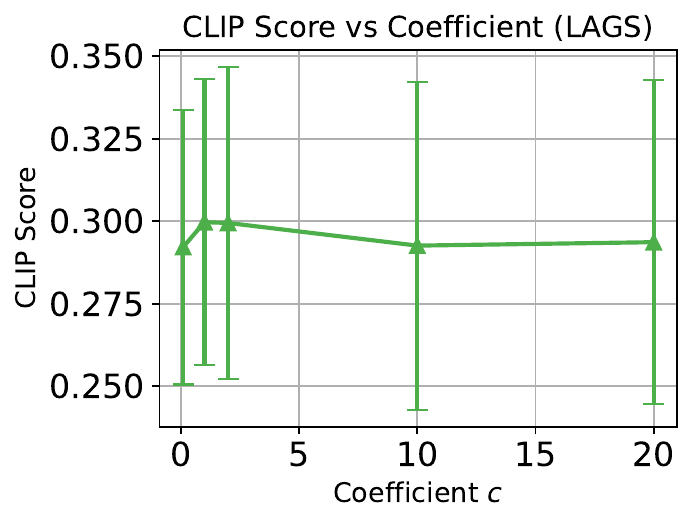}
    \includegraphics[width=0.24\linewidth]{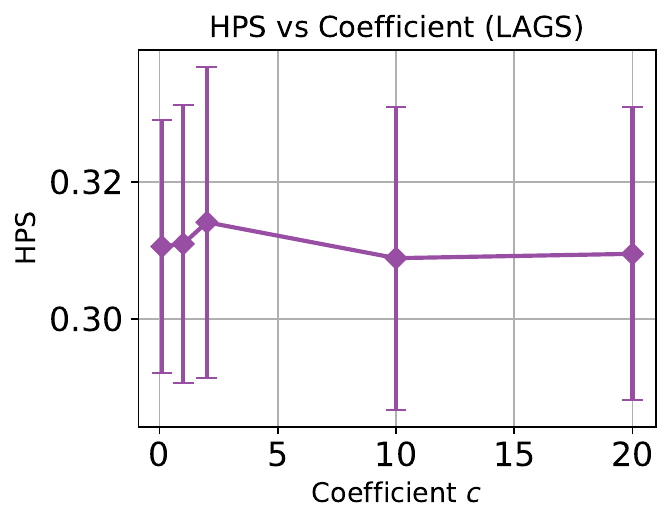}
    \caption{ Performance over the choice of $c$.
    \textbf{Columns (left to right):} PickScore, Image Reward, CLIP Score, and HPS. 
    Each subplot plots metric values against scheduler coefficient $c \in [0.1, 20]$.
    }
    \label{fig:coefficient_ablation_apg}
\end{figure}
As shown in Figure~\ref{fig:coefficient_ablation_apg}, increasing the scheduler parameter $c$ reduces the magnitude of the guidance and thus decreases PickScore monotonically. The other metrics such as ImageReward, CLIP Score, and HPS exhibit non-monotonic behavior.
Remarkably, with the wide range of $c\in[0.1, 20]$, our method remains on the Pareto frontier across PickScore, HPS, and runtime.

\begin{table*}[t]
\centering
\scriptsize
\begin{tabular}{lccccc}
\toprule
Metric 
& SDXL (base) 
& DPS 
& FreeDoM 
& DAS 
& DAS-1P \\
\midrule
Runtime  $\downarrow$
& 0.377 
& 1.193 
& 1.930 
& 4.722 
& 1.202 \\
BRISQUE $\downarrow$
& 21.48 $\pm$ 12.45
& 24.52 $\pm$ 13.03
& 34.49 $\pm$ 15.20
& \textbf{19.55 $\pm$ 12.75}
& 22.56 $\pm$ 11.97 \\
MANIQA $\uparrow$
& 0.733 $\pm$ 0.139
& 0.720 $\pm$ 0.139
& 0.711 $\pm$ 0.144
& 0.720 $\pm$ 0.145
& 0.728 $\pm$ 0.130 \\
\bottomrule
\end{tabular}
\begin{tabular}{lccccc}
\toprule
Metric 
& LAGS ($c{=}0.1$) 
& LAGS ($c{=}1$) 
& LAGS ($c{=}2$) 
& LAGS ($c{=}10$) 
& LAGS ($c{=}20$) \\
\midrule
Runtime $\downarrow$
& \textbf{1.000} 
& \textbf{1.000} 
& \textbf{1.000} 
& \textbf{1.000} 
& \textbf{1.000} \\
BRISQUE $\downarrow$
& 25.04 $\pm$ 15.56
& 21.88 $\pm$ 12.50
& 21.24 $\pm$ 12.18
& 19.87 $\pm$ 13.97
& 20.12 $\pm$ 14.39 \\
MANIQA $\uparrow$
& \textbf{0.745 $\pm$ 0.119}
& 0.737 $\pm$ 0.131
& 0.744 $\pm$ 0.132
& 0.732 $\pm$ 0.131
& 0.728 $\pm$ 0.132 \\
\bottomrule
\end{tabular}
\caption{Runtime and no-reference image quality measures (BRISQUE, MANIQA)
for LAGS (ours) with different guidance strengths $c$. Values are mean $\pm$ standard deviation.}
\label{tab:sdxl_runtime_brisque_maniqa_lags_aps}
\end{table*}
We further analyze the image quality measured by BRISQUE and MANIQA in Table \ref{tab:sdxl_runtime_brisque_maniqa_lags_aps}. Notably, despite achieving the fastest runtime, the proposed method attains the highest MANIQA score among all guidance methods. This indicates that, although there exists a trade-off between perceptual quality and the target PickScore, our method achieves the best overall performance across three axes: runtime, image quality, and alignment with the target weighting function.

For BRISQUE, DAS achieves the lowest score (corresponding to the highest perceptual quality). Nevertheless, the proposed method yields competitive performance, with BRISQUE values ranging from 19.87 to 25.04, indicating consistently acceptable image quality.

\begin{figure}[htbp]
    \centering
    \includegraphics[width=1.0\columnwidth]{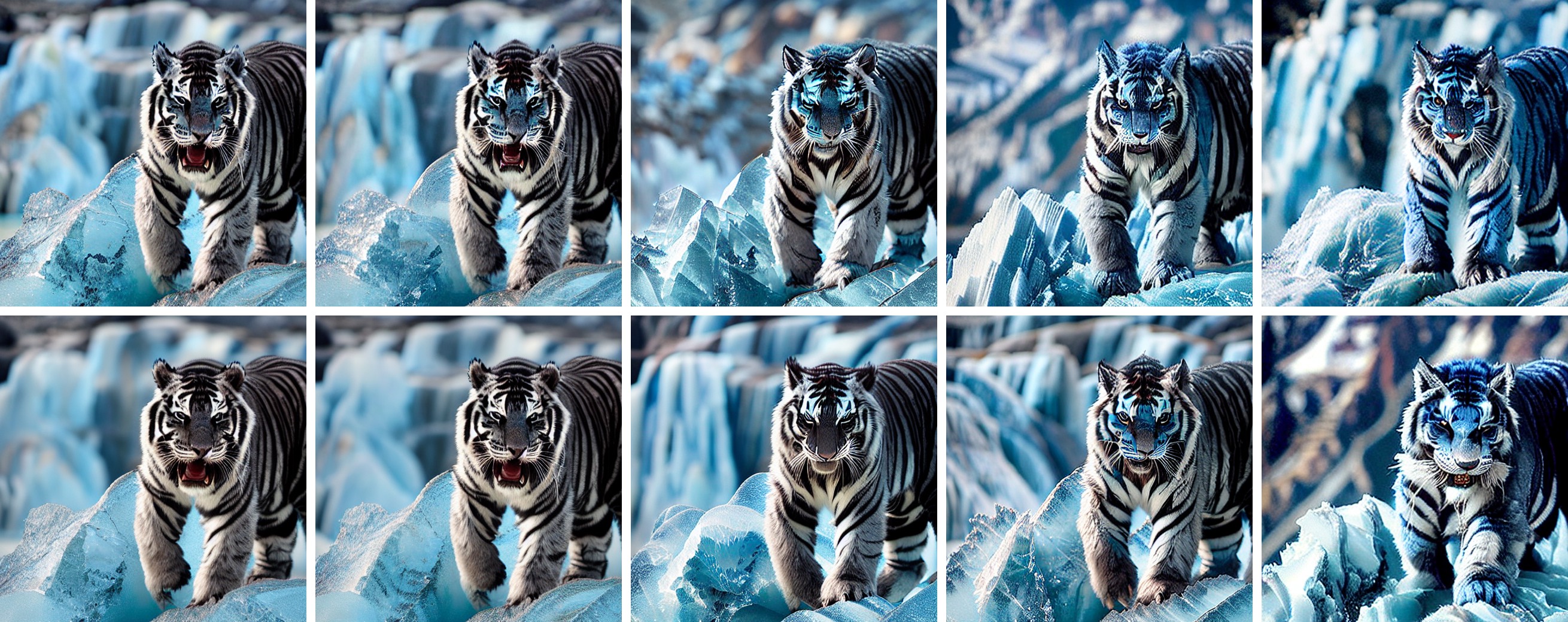}
    \caption{
    \textbf{Prompt:} \textit{“A blue-furred tiger on a glacier.”} 
    Each column shows images generated by LAGS with scheduler coefficients 
    $c \in \{20,10,1,0.1,0.01\}$ (left to right). Smaller values of $c$ increase 
    reliance on the approximated guidance. 
    \textbf{Top row:} standard LAGS. 
    \textbf{Bottom row:} projected guidance variant using $\mathbf{g}_k^{\perp}$, 
    which removes the DDIM direction from the guidance term. As $c$ decreases, 
    standard LAGS amplifies the blue-toned weighting effect, while projected 
    guidance produces softer and more natural textures.
    }
    \label{fig:blue_tiger_LAGS}
\end{figure}

\begin{figure}[htbp]
    \centering
    \includegraphics[width=1.0\columnwidth]{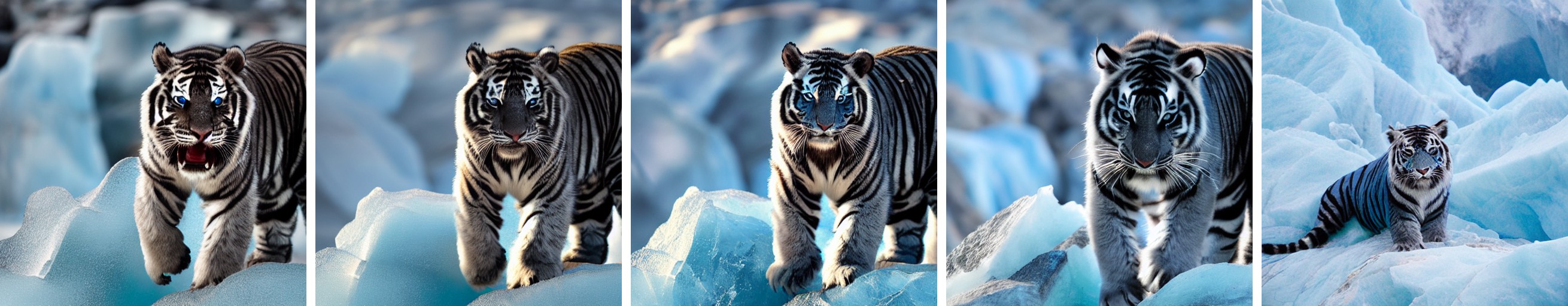}
    \caption{
    Images generated by baseline methods—SD, DPS, FreeDoM, DAS-1p, and DAS 
    (left to right)—using the same prompt and seed as Figure~\ref{fig:blue_tiger_LAGS}. 
    Among the baselines, DAS (rightmost column) achieves the highest PickScore (0.256) but requires 
    more than $3\times$ the computational cost of our method.
    }
    \label{fig:blue_tiger_baselines}
\end{figure}

In the top row of Figure~\ref{fig:blue_tiger_LAGS}, we show images generated by LAGS for different scheduler coefficients $c \in \{20, 10, 1, 0.1, 0.01\}$ (left to right). As $c$ decreases and the sampler relies more heavily on the approximated guidance, the model progressively emphasizes the weight function, leading to a stronger presence of the blue-toned features encouraged by the target score.

In the bottom row, we apply the projected guidance variant using $\mathbf{g}_k^{\perp}$, which removes the original DDIM direction from the guidance term. This modification reduces the influence of the color-specific weighting while producing noticeably softer and more natural textures. For example, at $c = 0.01$, the projected version yields smoother and more coherent structures, whereas the standard LAGS output exhibits stronger but slightly harsher color contrast.

Figure~\ref{fig:blue_tiger_baselines} presents results from the baseline methods (SD, DPS, FreeDoM, DAS-1p, DAS) generated with the same prompt and random seed. Among these baselines, DAS achieves the highest PickScore (0.256) and correctly captures the ``blue-furred'' appearance, but requires more than $3\times$ the computational cost of our method, highlighting the efficiency advantage of LAGS.

\subsubsection{Ablation Studies}
\label{appendix_text_image_alignment_ablation}
We introduced an uncertainty-based scheduler parameterized by $c$, which modulates the strength of the guidance signal across the diffusion trajectory. Importantly, the scheduler converges to $1$ regardless of the specific choice of $c$, ensuring that the model increasingly relies on the approximated guidance term near the terminal phase of the diffusion process.

The role of the parameter $c$ is to control the steepness and onset of the scheduler. A smaller $c$ yields a stronger guidance signal and an earlier rise in the scheduler curve  (See Appendix \ref{proof:proposition_scheduling}), whereas a larger $c$ results in a weaker signal and delays its saturation, resembling a sharper sigmoid. In this subsection, we conduct ablations across a range of $c$ values to understand their influence on generation quality and alignment.

\begin{figure}[htbp]
    \centering
    \includegraphics[width=1.0\columnwidth]{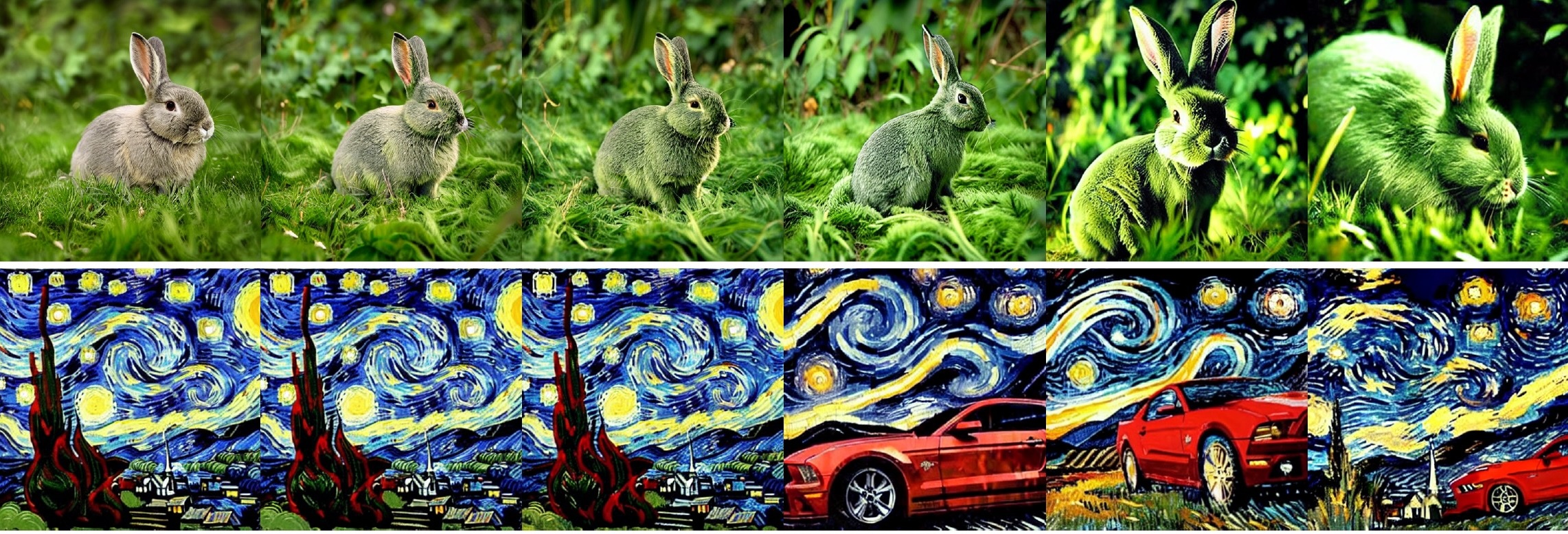} 
        \caption{
\textbf{Prompts (top to bottom):} \textit{“A green colored rabbit”, “A red Ford Mustang in the style of Starry Night by Van Gogh”}. 
\textbf{Each column:} Images generated by LAGS with different scheduler parameters $c = 100, 10, 1, 0.1, 0.01, 0.001$ (left to right). Smaller $c$ indicates high reliance on the approximated guidance.
}
    \label{fig:rabit_mustang}
\end{figure}

\paragraph{Visualization with varying $c$.}
Figure~\ref{fig:rabit_mustang} visualizes the effect of varying $c$. As expected, smaller values of $c$ produce stronger guidance, resulting in stronger alignment between the image and the prompt semantics. In contrast, large $c$ values (e.g., $c=100$) often fail to capture key prompt attributes. For instance, the leftmost column does not preserve the color cue (“green” rabbit) or the object category (“red car”). Conversely, the rightmost column ($c = 0.001$) yields images that are semantically faithful to the prompts and achieve the highest PickScore among the variants.

As $c$ increases, the generated images tend to gradually emphasize visual features, particularly evident in color fidelity for the first row. The leftmost column shows the failure mode of generated images with a missing color property, semantic entity. As $c$ increases, with higher probability, the generated samples show high alignment with the given texts. This behavior may vary depending on the prompt and the content of the generated image or the given conditional texts, pointing to potential discrepancies between quantitative alignment metrics and visual perception \cite{wu2023hps}, especially for some specific set of prompts or samples.

\paragraph{Performance cross validation with varying $c$.}
We test the entire prompt set with various $c$ values and measure the target PickScore along with the different alignment metrics to examine the performance transition.

\begin{figure}[t]
    \centering
    \includegraphics[width=0.24\linewidth]{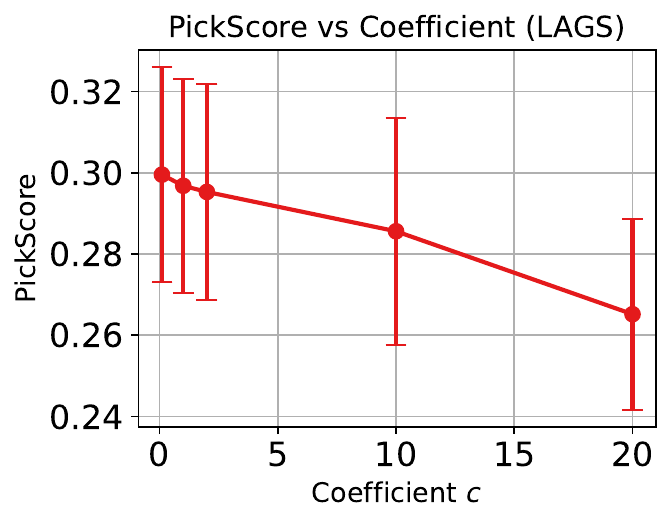}
    \includegraphics[width=0.24\linewidth]{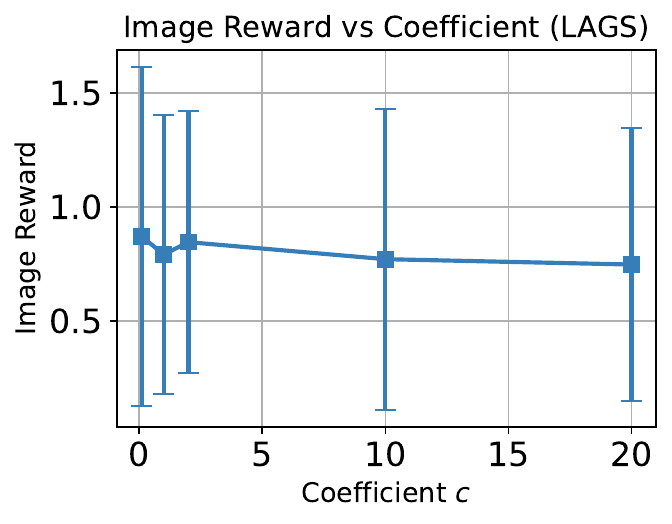}
    \includegraphics[width=0.24\linewidth]{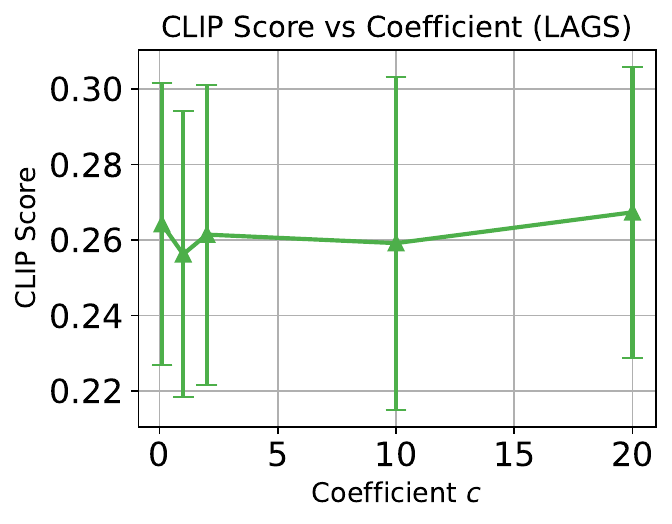}
    \includegraphics[width=0.24\linewidth]{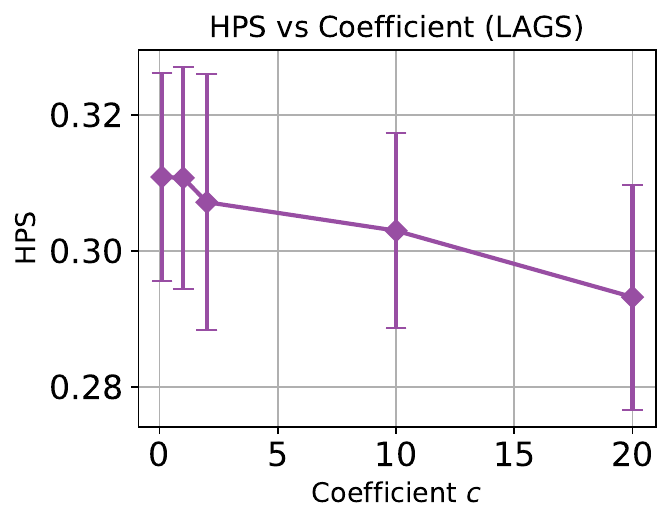}
    \includegraphics[width=0.24\linewidth]{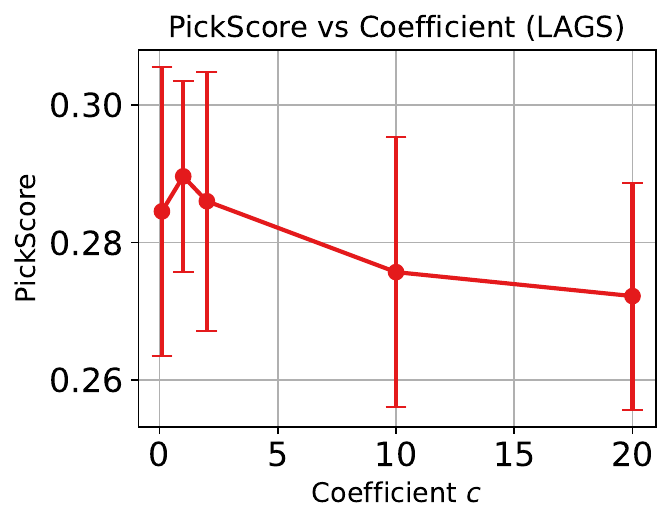}
    \includegraphics[width=0.24\linewidth]{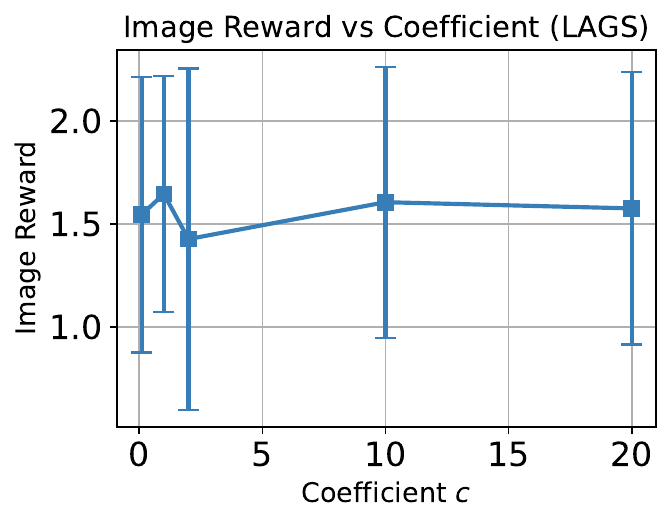}
    \includegraphics[width=0.24\linewidth]{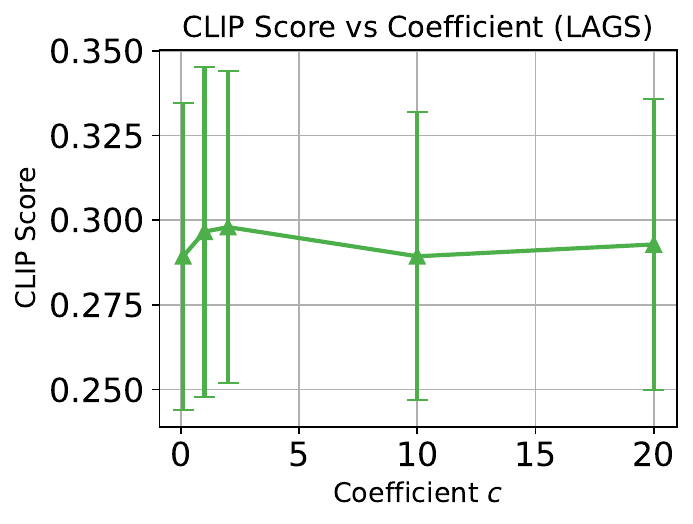}
    \includegraphics[width=0.24\linewidth]{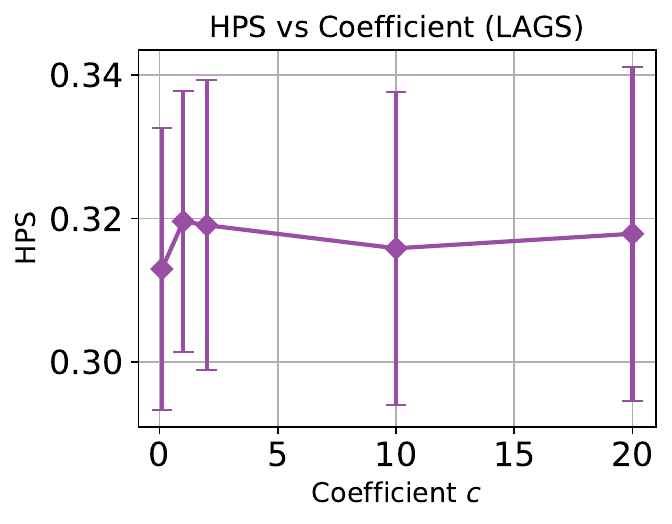}
    \caption{ Performance over the choice of $c$.
    \textbf{Columns (left to right):} PickScore, Image Reward, CLIP Score, and HPS. 
    \textbf{Rows:} Top – Stable Diffusion (SD), Bottom – SDXL. Each subplot plots metric values against scheduler coefficient $c \in [0.1, 20]$.
    }
    \label{fig:coefficient_ablation}
\end{figure}

Figure~\ref{fig:coefficient_ablation} summarizes the empirical behavior of PickScore under different choices of the scheduler coefficient $c$. Since larger $c$ attenuates the contribution of the guidance term $\tilde{g}(\rvx,t)$, increasing $c$ naturally reduces the influence of the weighted-sampling direction during the backward diffusion process. This manifests as a consistent downward trend in PickScore, i.e., the average target weight $\mathbb{E}[w(\mathbf{x})]$. For SD, this decay is strictly monotonic. For SDXL, the trend is similar, with $c=1$ achieving the highest PickScore before gradually declining for larger values of $c$.

This monotonic behavior in PickScore does not always mirror movement in the other alignment metrics such as CLIP Score, HPS, or ImageReward. Although guidance-based methods, including ours, reliably improve the target-specific objective (PickScore), these metrics occasionally exhibit misalignment. This reflects inherent metric heterogeneity rather than a limitation of the proposed method and baselines, and suggests that more general alignment-oriented weight functions may be needed to simultaneously elevate all alignment metrics.

Across a broad range of scheduler strengths, the proposed method consistently achieves the highest PickScore among all training-free guidance approaches, reaffirming its ability to reliably increase the expected weight $\mathbb{E}[w(\mathbf{x})]$. Empirically, we observe that even moderate values of $c$ retain strong performance, demonstrating robustness of the lightweight approximation and validating the effectiveness of the uncertainty-adaptive scheduling strategy.

\subsection{Experiments on CelebA Dataset}
\label{appendix_experiments_details_celeba}

This section examines an application of the proposed weighted sampling method in scenarios where the weight function is parameterized by a neural classifier.
Consider a setting in which an SGM is trained without any class supervision, for example, without labels related to gender in the CelebA dataset. Although the SGM has no inherent notion of class, we assume access to an external binary classifier that predicts whether an image is labeled as ``man’’ or ``not man.’’

The goal is to increase the sampling probability of a particular class. This is especially relevant in situations where the base SGM exhibits class imbalance or bias; in such cases, classifier-driven weighted sampling can adjust the output distribution and mitigate these biases.

To implement this, the weight function $w(\rvx)$ is defined using the classifier’s logit output. Images that the classifier is more confident belong to the target class (e.g., “man’’) are assigned larger weights, thereby steering the sampling process toward that class through the proposed weighted sampling mechanism.

\paragraph{Dataset.} 
The CelebA dataset \cite{liu2015celebafaceattributes} is utilized for training the classifier in our experiments. For the score function used in the generation process, we employ a pretrained score model based on DDPM \cite{ho2020ddpm}. This pretrained model is obtained from Huggingface\footnote{\url{https://huggingface.co/google/ddpm-celebahq-256}} and is capable of generating RGB images at a resolution of $256 \times 256$ pixels.

\paragraph{Weight function.} 
The design of the weight function begins with training a classifier based on the ResNet18 architecture using the CelebA dataset \cite{liu2015celebafaceattributes}. The classifier is specifically trained to predict whether a given instance is classified as ``man'' or not, utilizing the provided labels. The final activation layer of the classifier employs a sigmoid function to produce probabilistic outputs. During training, we use a batch size of 256, a total of 25 epochs, and the Adam optimizer with a learning rate of $10^{-3}$. We denote this classifier as $F_{\text{cl}}$, where $F_{\text{cl}}: \mathbb{R}^d \mapsto \{0, 1\}$.

Assume that the objective of weighted sampling is to increase the probability density of samples classified as ``man.'' To construct the weight function $w(\rvx)$, we adopt an approach similar to that described in Appendix~\ref{appendix_experiments_details_mnist}. Specifically, the weight function is defined as $w(\rvx) = D_{\text{bce}}(F_{\text{cl}}(\rvx), 0)$, where $D_{\text{bce}}$ represents the scaled binary cross-entropy loss function. Intuitively, this formulation indicates that the weight $w(\rvx)$ takes higher values when the instance $\rvx$ is classified as ``man.''

\begin{figure*}[htbp]
    \centering
    \setlength{\tabcolsep}{2pt}  
    \begin{tabular}{c}
        \begin{tabular}{ccccc}
            \includegraphics[width=0.121\columnwidth]{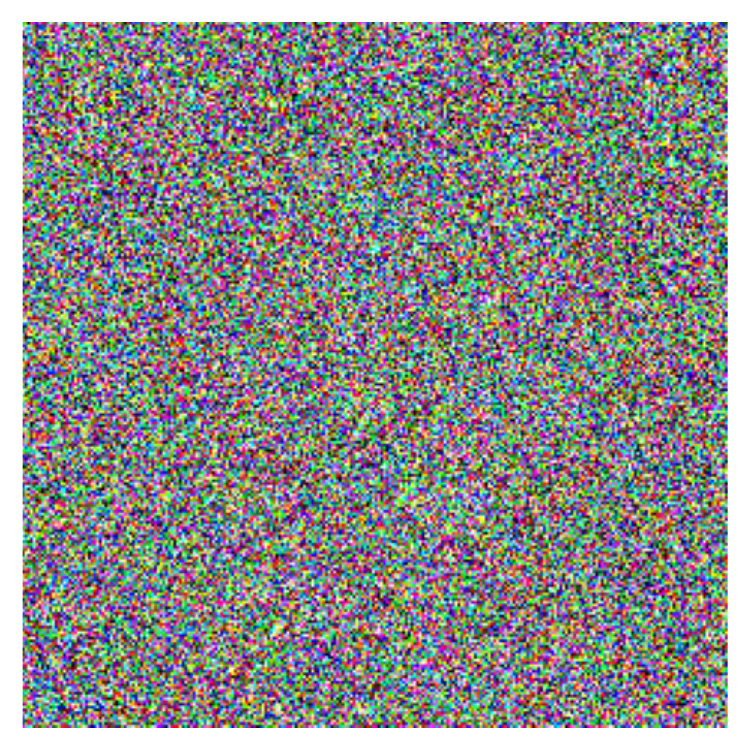}&
            \includegraphics[width=0.121\columnwidth]{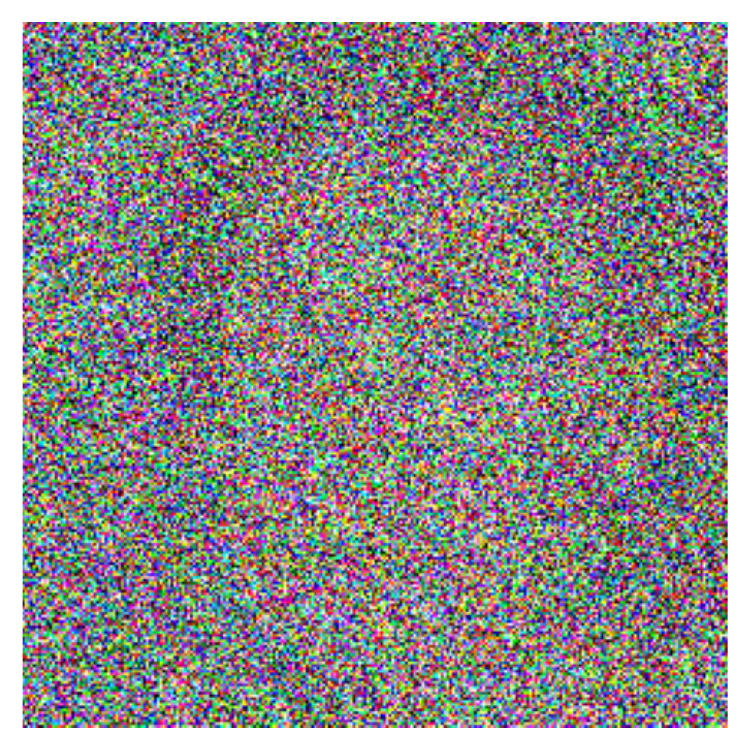}&
            \includegraphics[width=0.121\columnwidth]{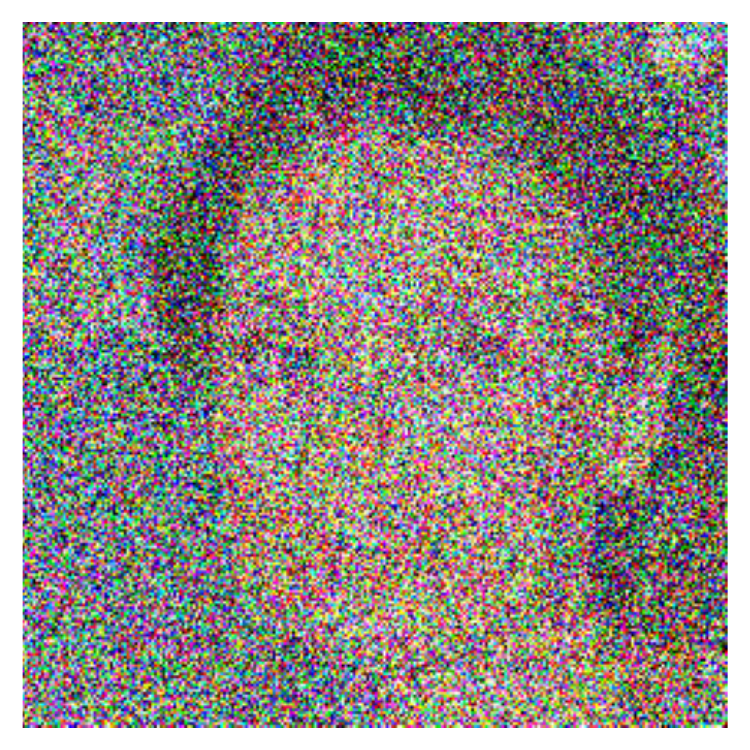}&
            \includegraphics[width=0.121\columnwidth]{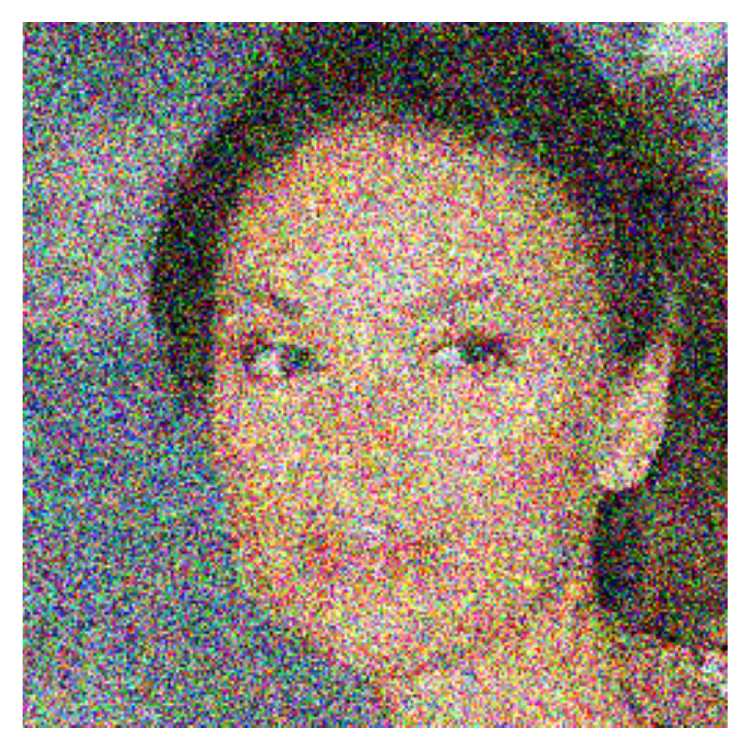} &
            \includegraphics[width=0.121\columnwidth]{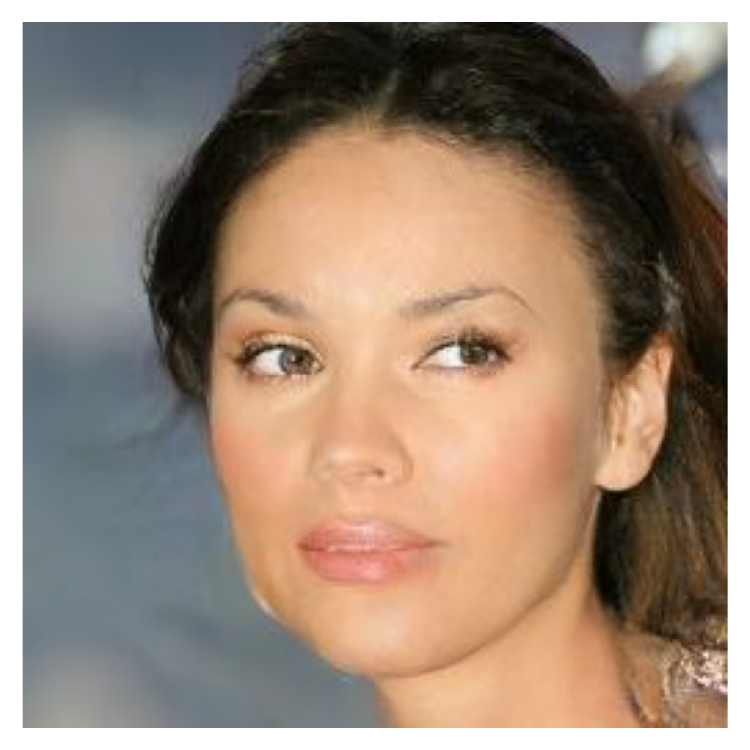} \\
        \end{tabular} \\

        \begin{tabular}{ccccc}
            \includegraphics[width=0.121\columnwidth]{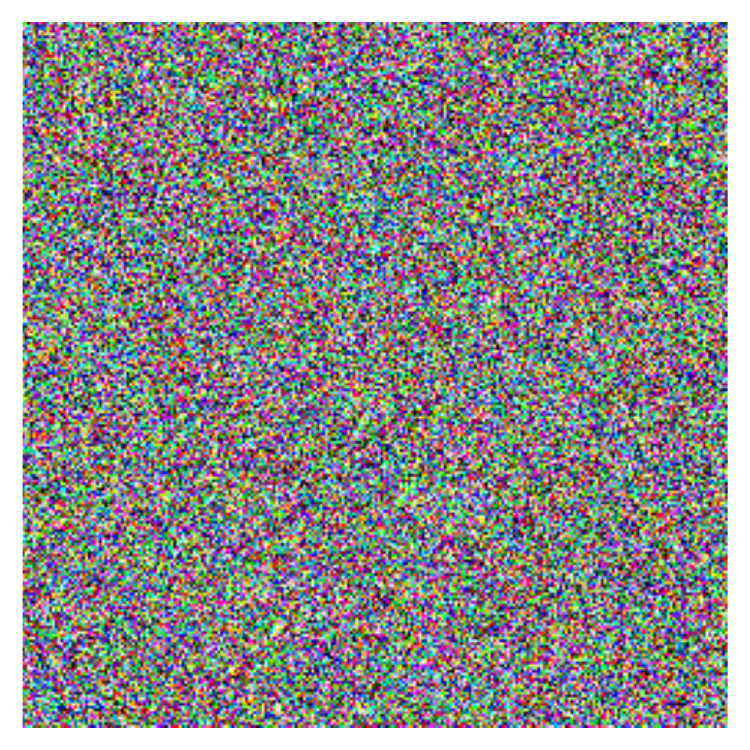}&
            \includegraphics[width=0.121\columnwidth]{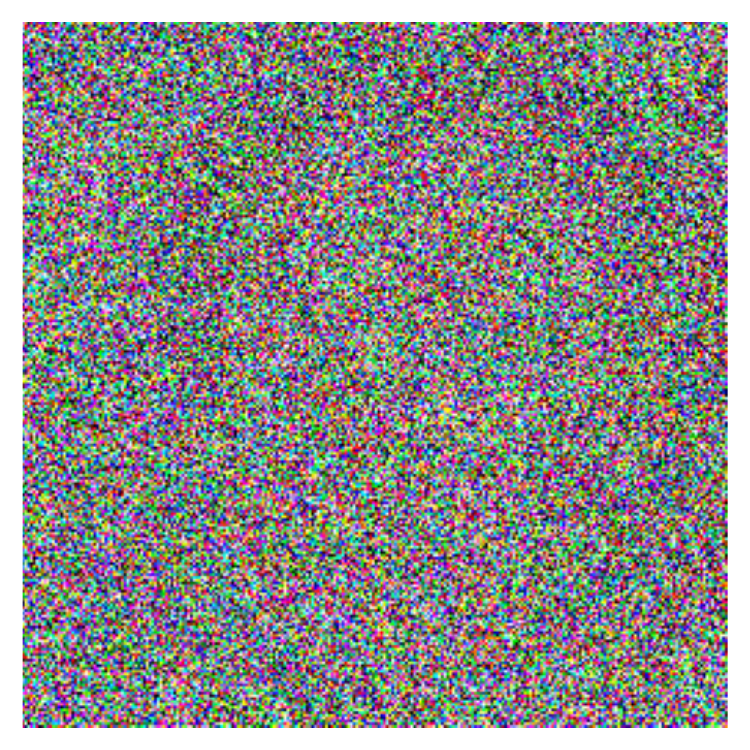}&
            \includegraphics[width=0.121\columnwidth]{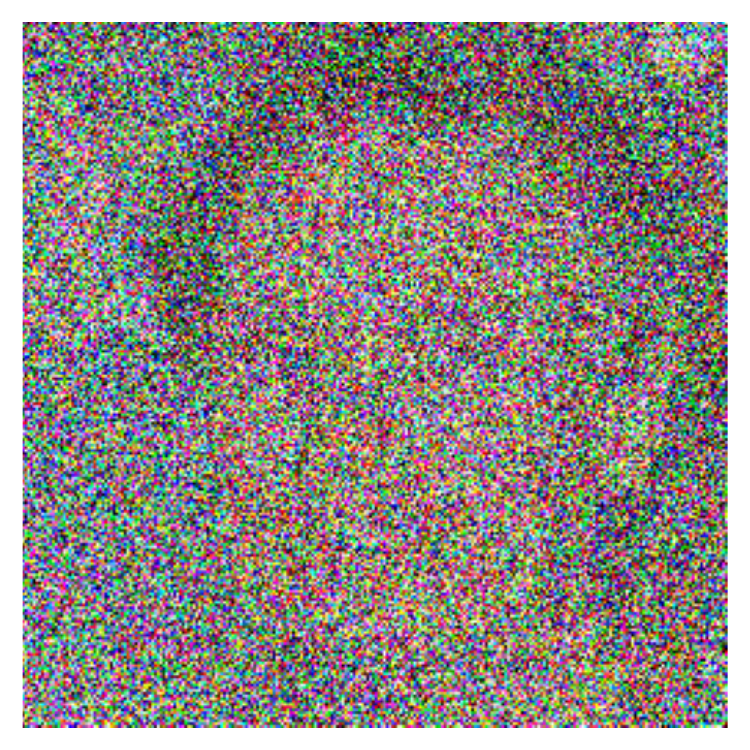}&
            \includegraphics[width=0.121\columnwidth]{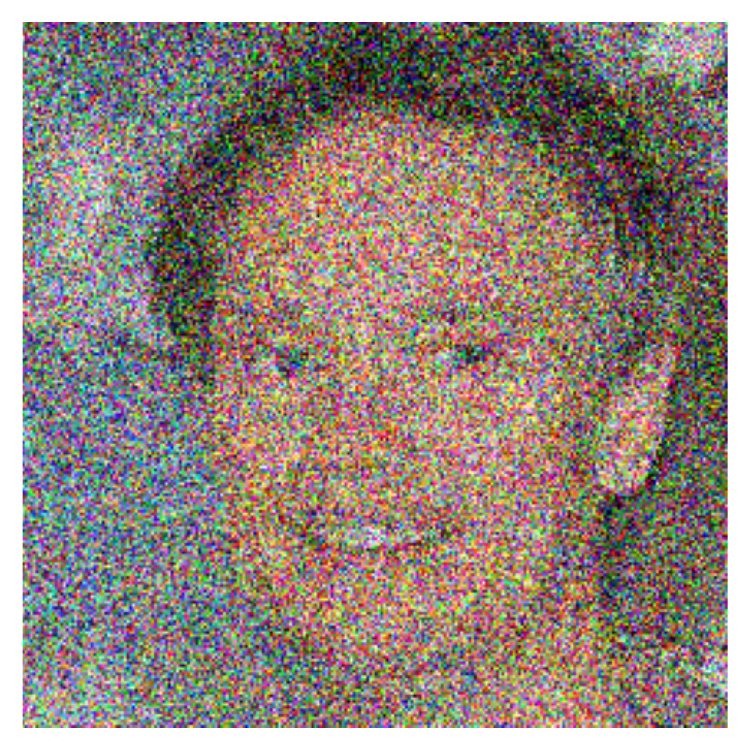} &
            \includegraphics[width=0.121\columnwidth]{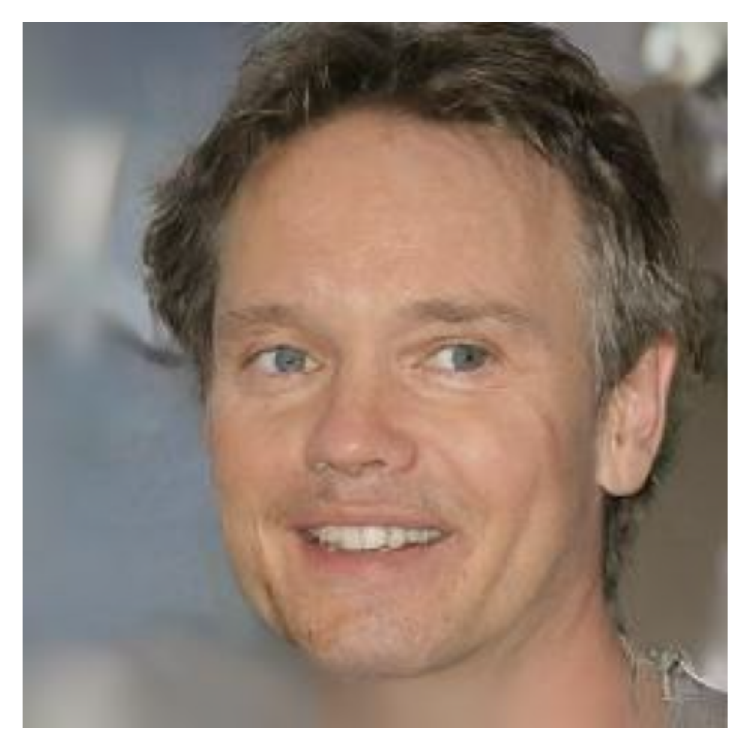} \\
        \end{tabular} \\
        \hline
    \end{tabular}

\caption{
    \textbf{Pairs of Two Diffusion Backward Processes.} \textbf{\emph{Top}}: The standard backward diffusion process for sampling from $p(\rvx)$. \textbf{\emph{Bottom}}: The proposed backward diffusion process for weighted sampling from $q(\rvx)$. The proposed method leverages an externally defined weight function, such as a neural classifier, to prioritize samples with higher weight. This approach can accommodate \emph{any differentiable weight function, i.e., neural classifiers}, for weighted sampling—entirely independent of the base SGM—without requiring additional training.
}

    \label{fig:celeba_generation_process_2}
\end{figure*}

\paragraph{Results.}  
By applying the neural classifier-based weighting function, the proportion of instances classified as ``man'' increased from 35.1\% to 51.2\%. This result empirically validates that our proposed weighted sampling approach can effectively modulate the output distribution by leveraging a classifier that is independently trained from the generative model. Notably, this experiment highlights the scalability of our method, demonstrating its applicability irrespective of the generative model's size.

Figure.~\ref{fig:celeba_generation_process_2} illustrates pairs of two diffusion processes: the top row corresponds to the standard backward diffusion process, while the bottom represents the weighted backward diffusion process achieved using our proposed approach. From left to right, each column depicts discrete timesteps $800, 600, 400, 200, 0$, respectively, out of a total of $1000$. The rightmost column, therefore, represents the final generated samples drawn from the distributions $p(\rvx)$ and $q(\rvx)$.

In Figure~\ref{fig:celeba_generation_process_2}, the class outputs of the two diffusion processes differ and show that the weighted sampling mechanism guides the backward process toward generating samples classified as ``man.'' Importantly, both processes share identical sources of randomness, including the same initial state and identical diffusion process noise, with the only distinction being the difference in score functions.

The results show promise for advancing AI fairness by mitigating class bias introduced by generative models with fast guidance. Additionally, it can be effectively employed for class-specific filtering tasks, such as identifying and removing inappropriate images containing undesirable content. These capabilities underscore the versatility and utility of the proposed method in various practical applications.

This method represents a distinct and efficient alternative to density-estimation-based fairness derivation techniques, such as \cite{kim2024training}, which require extensive retraining to achieve fairness adjustments. We anticipate that our training-free weighted sampling approach will pave the way for new advancements in fairness, a goal of growing importance in the development of machine learning models \cite{bellamy2019ai, trewin2019considerations, john2022reality}.

\subsection{Experiments on MNIST Dataset}
\label{appendix_experiments_details_mnist}
When using a pretrained SGM, the generated samples may inherently exhibit class-wise biases. In certain applications, it is crucial either to mitigate such biases to ensure fairness or to intentionally introduce biases to suppress specific classes, such as filtering out potentially harmful or undesirable categories.

These problems can be naturally formulated within a weighted sampling framework by defining a weight function $w(\rvx)$ that assigns higher values to samples that do not belong to the target class to be suppressed. 

Furthermore, a similar approach can be employed to promote class-wise uniformity in the generated distribution. By assigning higher weights to underrepresented classes, the sampling process can be adjusted to achieve a more balanced class distribution, thereby approximating a uniform class-wise representation.

\paragraph{Dataset.} 
We utilize the MNIST dataset \cite{lecun1998gradient}, which comprises ten classes of handwritten digits. Each sample is normalized to the range $[-1,1]$ for both training the neural classifier and the score function of the original PDF.

\paragraph{Weight function.} 
In this experiment, a neural classifier-based weight function is employed. Specifically, we utilize ResNet18 \cite{he2016deep} as the backbone classifier, with its first layer modified to accommodate the single-channel input structure of the MNIST dataset. The classifier is trained using the Adam optimizer with a learning rate of $1\times 10^{-3}$, a batch size of 256, and for 100 epochs. This configuration yields a neural classifier capable of achieving a classification accuracy of 99.4\% on the MNIST test set. We denote the trained classifier as $F_{\text{cl}}(\rvx)$, where $F_{\text{cl}}: \mathbb{R}^{d} \mapsto \mathbb{R}^{10}$, i.e., the classifier maps each input sample $\rvx$ to a 10-dimensional logit vector. 

The softmax output of the logits is expected to exhibit a high value at the index corresponding to the correct label. Using this neural classifier, we construct a weight function $w(\rvx)$ to selectively downweight specific labels. For instance, consider a scenario where the objective is to downweight the probability of sampling a particular label $l \in \{ 0,1,\ldots, 9\}$. Let $\rve_{l}$ denote a one-hot vector with a value of 1 at the $l$-th position (0-indexed) and zeros elsewhere. Setting $w(\rvx) = D_{\text{ce}}(F_{\text{cl}}(\rvx), \rve_{l})^2$, where $D_{\text{ce}}$ represents the cross-entropy loss, ensures that samples with lower logit values at the $l$-th index are assigned higher importance weights. Consequently, weighted sampling with this weight function increases the likelihood of sampling samples that are less likely to be classified as label $l$.

Assume we wish to downweight class $l \in [9]$. We define the weight function as $w(\rvx) = D_{\text{ce}}( F_{\text{cl}}(\rvx), \rve_{l} )^2$ where $\rve_{l}$ is a one-hot encoded vector with the $l$-th element set to one and all others set to zero, and $D_{\text{ce}}$ represents the cross-entropy loss. This weight function assigns greater significance to samples yielding small logit value for the $l$-th label, thereby effectively downweighting class $l$.

\paragraph{Base score function.}  
The score function of the original PDF, $\nabla_{\rvx} \log p_{t}(\rvx)$ is derived through DDPM training. The score function is implemented using a U-Net-based architecture. The architecture consists of three downsampling blocks and three corresponding upsampling blocks, organized in a symmetrical encoder-decoder configuration. The channel dimensions for the blocks increase progressively, with depths set to (32, 64, 128), respectively. To ensure stable and efficient training, group normalization is applied with four groups per feature map. Furthermore, a padding value of 1 is used during downsampling operations. The model accepts input tensors $\rvx_{t}$ and the diffusion time step $t$, and outputs the denoised sample prediction. 

After obtaining the score function of the original PDF, DDPM sampling is utilized for the implementation of weighted sampling.

\paragraph{Results.}

\begin{figure}[t]
    \centering
    \includegraphics[width=0.6\linewidth]{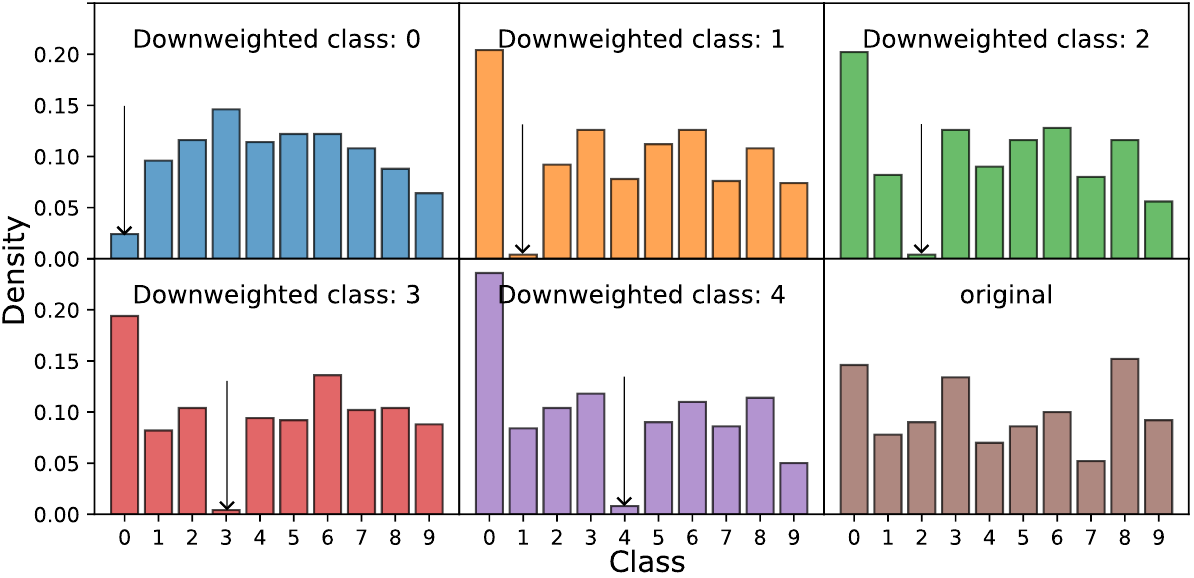}
    \caption{
        Class histograms from the original generative model trained without class labels (bottom right) and from the proposed weighted sampling (first five histograms, left to right, top to bottom) using a neural network classifier-based importance weight. Samples with lower logits for the specified downweighted class are assigned higher importance weights.    
    }
    \label{fig:mnist_histograms}
\end{figure}

In Figure \ref{fig:mnist_histograms}, the first five histograms (from left to right, top to bottom) correspond to $l\!=\!0$ to $l\!=\!4$, while the bottom-right histogram displays results without weighted sampling. These histograms are generated by the neural classifier $F_{\text{cl}}(\rvx)$, by sampling $10^{4}$ instances followed by classification through $F_{\text{cl}}(\rvx)$.
The results demonstrate that the proposed weighted sampling technique effectively attenuates the representation of target classes relative to the original distribution by assigning greater weight to instances classified outside of the target class. Note that this weighted sampling was achieved without additional model training for different values of $l$, relying solely on small modifications to $l$. 

\begin{figure}[t]
\centering
\begin{minipage}[t]{0.45\linewidth}
    \centering
    \includegraphics[width=0.75\linewidth]{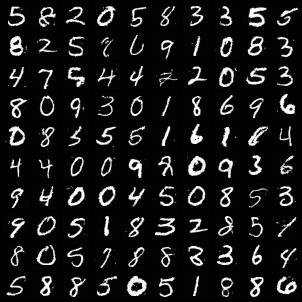}
\end{minipage}%
\hspace{0.1\linewidth}%
\begin{minipage}[t]{0.45\linewidth}
    \centering
    \includegraphics[width=0.75\linewidth]{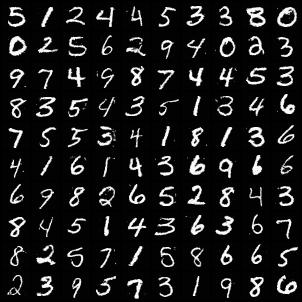}
\end{minipage}
    \caption{
    \textbf{\emph{Left}}: Random samples from $p(\rvx)$.
    \textbf{\emph{Right}}: Random samples from $q(\rvx)$ with \emph{higher importance weights for instances with low values at the 0-th logit} of the neural classifier.  Those instances in identical positions in the two figures share the same randomness (i.e., the initial state and noise for the diffusion backward processes).
    In the original distribution, approximately 15\% of the instances are classified with label 0. Through weighted sampling, the density of instances with label 0 can be significantly reduced. On the right, only 3 out of 100 instances are assigned label 0 under weighted sampling.
    }
    \label{fig:mnist_downweight_0}
\end{figure}
The left image in Figure~\ref{fig:mnist_downweight_0} illustrates 100 samples drawn from the original PDF $p(\rvx)$ using its score function, $\nabla_{\rvx} \log p_{t}(\rvx)$, in conjunction with the DDPM sampling method. We maintain consistent randomness across experiments, i.e., identical initial states and noise values for the stochastic backward process. 
The right image in Figure~\ref{fig:mnist_downweight_0} shows importance samples generated using the proposed weighted sampling method, wherein the weight function is defined as $w(\mathbf{x}) = D_{\text{ce}}(F_{\text{cl}}(\mathbf{x}), \mathbf{e}_0)^2.$ This weighting mechanism assigns higher importance to samples with lower logit values for class 0.

A key observation from Figure~\ref{fig:mnist_downweight_0} is the marked reduction in the frequency of samples classified as class 0—only 3 out of 100—compared to the original distribution. This result demonstrates the efficacy of the proposed weighted sampling method in selectively reducing the likelihood of sampling instances associated with a specified class. The approach leverages a neural classifier as a main component in defining the weight function, highlighting its flexibility in terms of designing the weight function.

\begin{figure}[t]
\centering
    \begin{tabular}{cc}
    \includegraphics[width=0.24\linewidth]{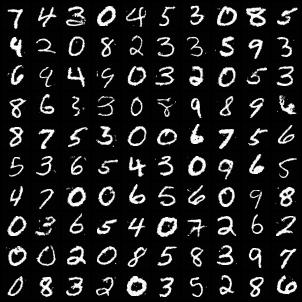}
    \includegraphics[width=0.24\linewidth]{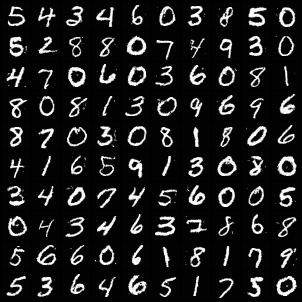}
    \includegraphics[width=0.24\linewidth]{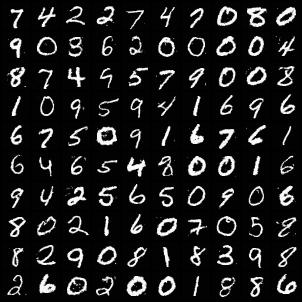}
    \includegraphics[width=0.24\linewidth]{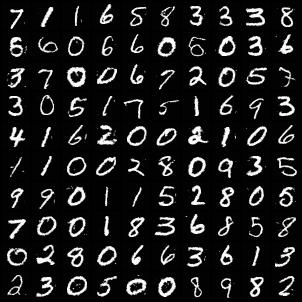}
    \end{tabular}
    \caption{
    \textbf{\emph{From left to right}}: Importance samples generated using the weight function $w(\mathbf{x}) = D_{\text{ce}}(F_{\text{cl}}(\mathbf{x}), \mathbf{e}_l)^2,$ where $l = 1, 2, 3, 4$, respectively. For each case, higher importance weights are assigned to instances with low logit values corresponding to class $l$. For instance, in the first image, no instances classified as class $1$ are present. Similarly, the second image contains only one instance classified as class $2$, while the third and fourth images lack instances classified as classes $3$ and $4$, respectively. This is in contrast to the set of samples from the original distribution shown in Figure~\ref{fig:mnist_downweight_0}.
    }
    \label{fig:mnist_downweight_1234}
\end{figure}
In Figure~\ref{fig:mnist_downweight_1234}, we illustrate the outcomes of employing different weight functions, denoted as $w(\rvx) = D_{\text{ce}}(F_{\text{cl}}(\rvx), \rve_{1})^2$, $w(\rvx) = D_{\text{ce}}(F_{\text{cl}}(\rvx), \rve_{2})^2$, $w(\rvx) = D_{\text{ce}}(F_{\text{cl}}(\rvx), \rve_{3})^2$, $w(\rvx) = D_{\text{ce}}(F_{\text{cl}}(\rvx), \rve_{4})^2$, from left to right, respectively. These functions assign higher importance weights to instances with lower logit values for the respective classes $l=1,2,3,4$. Consistent with Figure \ref{fig:mnist_downweight_0}, the examples demonstrate that the proportion of instances classified into each specified class is less than 3\% over 100 generated samples which show the neural classifier driven weight function can be utilized to downweight specific set of instances' sampling probability.

\paragraph{Weighted sampling towards unbiased class probability}

\begin{figure}[h]
\centering
    \begin{tabular}{cc}
    \includegraphics[width=0.57\linewidth]{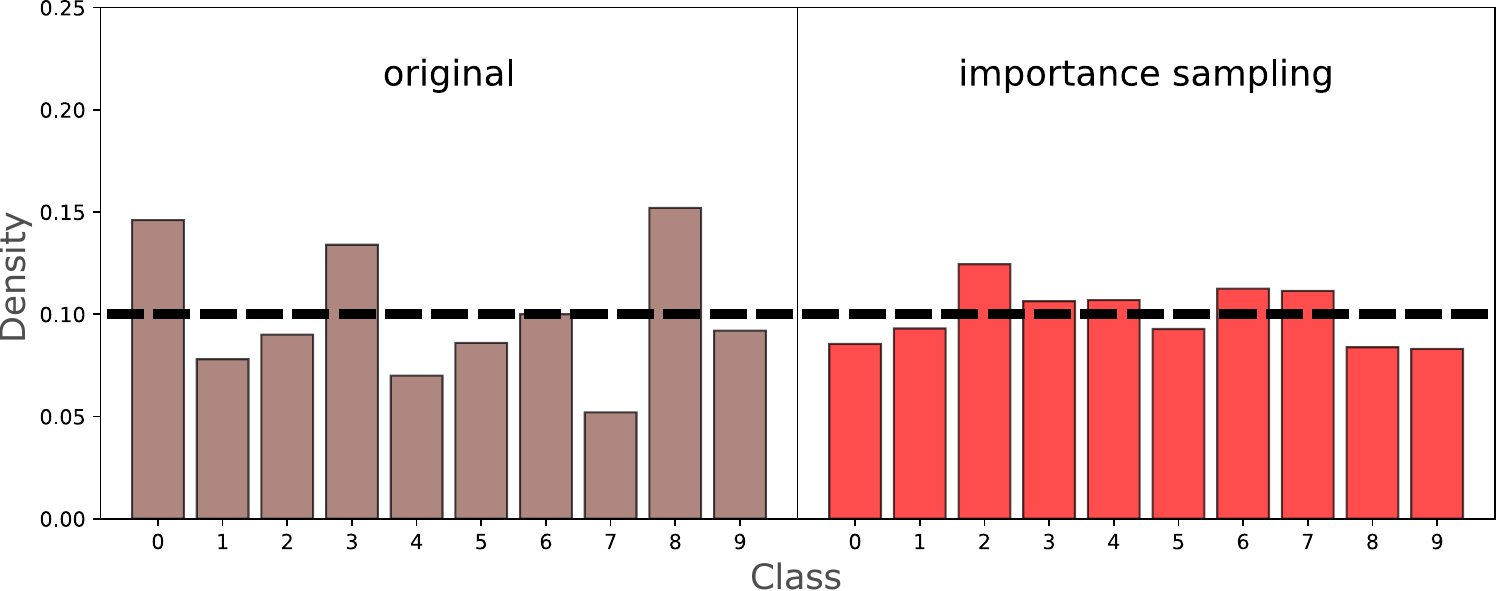}
    \end{tabular}
\caption{
\textbf{\emph{Left}}: Class probability distribution from the original score-based generative model. \textbf{\emph{Right}}: Class probability distribution from weighted sampling. The bias across classes is significantly reduced, with the density variance narrowing from $10^{-3}$ to $1.9 \times 10^{-4}$, demonstrating the effectiveness of the proposed method in balancing the class densities.
}
    \label{fig:mnist_uniform}
\end{figure}

The weighted sampling methodology can also be employed to mitigate bias and enhance class-wise fairness. For instance, consider a scenario where the objective is to control the class probabilities to achieve a uniform distribution. Let $\rvp \in [0,1]^{10}$ represent the vector of class probabilities, where the $l$-th element, $[\rvp]_c$, corresponds to the probability density of class $l$ from the original generative model. These probabilities can be readily obtained using a pretrained score-based generative model in conjunction with a classifier, as illustrated in the histogram at the bottom right of Figure~\ref{fig:mnist_histograms}, by generating samples and classifying them via the classifier.

An intuitive approach to achieve class-wise uniformity is to define a weight function $w(\rvx)$ such that if $\rvx$ is classified as class $l$, the importance weight is given by $1/[\rvp]_c$, where $[\rvp]_c$ is the $l$-th element of the vector $\rvp$. By employing this weighting mechanism, the effective importance weight for each class is equalized, thereby leading to a class-wise distribution with significantly improved uniformity.

To implement this approach, we utilize the Gumbel-softmax function $F_G$, which processes the logit outputs of the classifier to approximate the behavior of an argmax function. This allows us to reweight the logits by the inverse of $\rvp$. Specifically, the weight function is expressed as:
$w(\rvx) = F_G(F_{\text{cl}}(\rvx))^{\top} \left(\frac{1}{\rvp}\right),$
where $F_{\text{cl}}(\rvx)$ denotes the logit output of the classifier.

In Figure~\ref{fig:mnist_uniform}, we present the class probability density distribution generated by the original score-based generative model on the left and the class probability density obtained via weighted sampling using $w(\mathbf{x}) = F_G(F_{\text{cl}}(\mathbf{x}))^{\top} \left(\frac{1}{\mathbf{p}}\right)$ 
on the right. Notably, the bias in the class probability density across the classes is significantly reduced, with the class density variance decreasing from $10^{-3}$ to $1.9 \times 10^{-4}$. This demonstrates that control over class density can be effectively achieved by simply modifying the weight function.

\subsection{Experiments on Stable Cascade}
\label{appendix_experiments_details_stablecascade}
In this section, we explore various weight functions that emphasize different image attributes, such as color and frequency components. By leveraging these functions, our framework provides an additional layer of control over sampling in foundation diffusion models beyond text-prompt conditioning. Specifically, we examine how our weighted sampling method enables controlled variation in the generated distribution, even when the text prompt lacks explicit color, frequency, or stylistic attributes.

It should be noted that the examples provided in this section should not be interpreted as image transformations. The objective of weighted sampling is to adjust the distribution while remaining within the original support set, rather than altering individual images.

\paragraph{Model.} We utilize StableCascade model as our foundation SGM \cite{pernias2024wurstchen}. This model was trained on a curated subset of LAION-5B. We adopt its default DDIM-based sampling process \cite{song2021ddim} and integrate our method with the weight functions defined below.

\subsubsection{Color-biased Sampling}
\label{color_biased_sampling}

\paragraph{Weight function.} 
To emphasize specific colors in generated images, we define weight functions that bias the image toward one of the RGB color channels. An image, represented as $\mathbf{x} \in \mathbb{R}^{3 \times h \times w}$, contains red, green, and blue channels. The average intensity for each channel is computed as
\[
I_r(\rvx) \coloneqq \frac{1}{h \cdot w} \sum_{i=1}^h \sum_{j=1}^w [\mathbf{x}]_{1, i, j}, \quad
I_g(\rvx) \coloneqq  \frac{1}{h \cdot w} \sum_{i=1}^h \sum_{j=1}^w [\mathbf{x}]_{2, i, j}, \quad
I_b(\rvx) \coloneqq  \frac{1}{h \cdot w} \sum_{i=1}^h \sum_{j=1}^w [\mathbf{x}]_{3, i, j}.
\] For instance, to enhance red, we use $w_r(\mathbf{x}; \xi) = \exp\left( \xi \cdot \text{sig} \left( I_r(\rvx) - \frac{I_g(\rvx) + I_b(\rvx)}{2} \right) \right),$ where $\text{sig}$ is the sigmoid function, and $\xi$ determines the strength of the bias. Analogously, $w_g(\mathbf{x}; \xi)$ and $w_b(\mathbf{x}; \xi)$ are defined for green and blue channels, respectively, allowing flexible control over the generated image's color emphasis.

\paragraph{Results.}

\begin{figure*}[htbp]
    \centering
    \setlength{\tabcolsep}{2pt}  
    \begin{tabular}{c}
        \textit{$w_r(\mathbf{x} ; \xi)$} \\
        \begin{tabular}{cccccc}
            \includegraphics[width=0.15\textwidth]{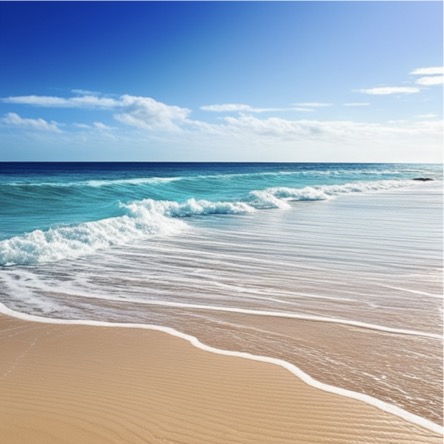} &
            \includegraphics[width=0.15\textwidth]{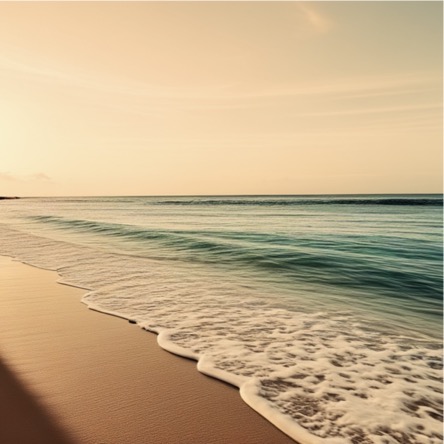} &
            \includegraphics[width=0.15\textwidth]{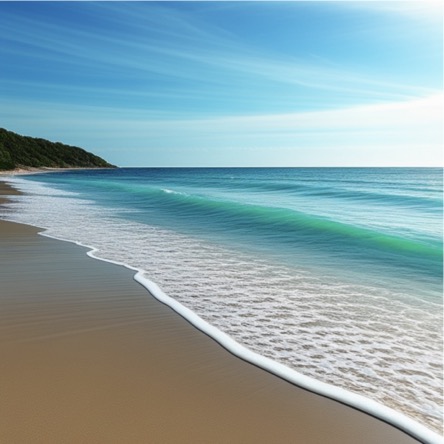} &
            \includegraphics[width=0.15\textwidth]{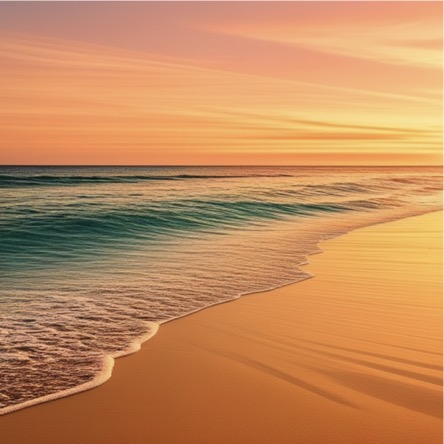} &
            \includegraphics[width=0.15\textwidth]{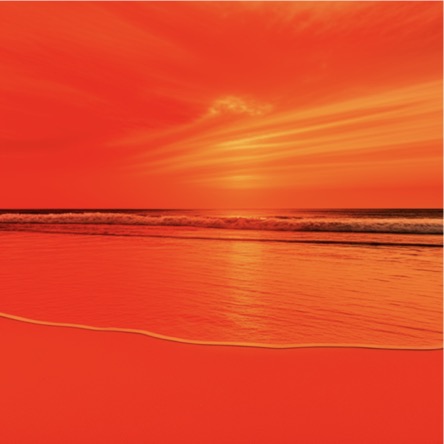} &
            \includegraphics[width=0.15\textwidth]{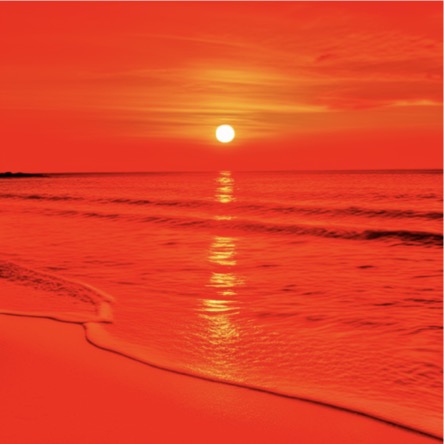} \\
        \end{tabular} \\
        \textit{$w_g(\mathbf{x} ; \xi)$} \\
        \begin{tabular}{cccccc}
            \includegraphics[width=0.15\textwidth]{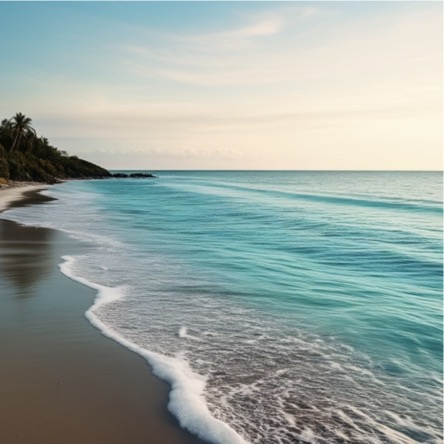} &
            \includegraphics[width=0.15\textwidth]{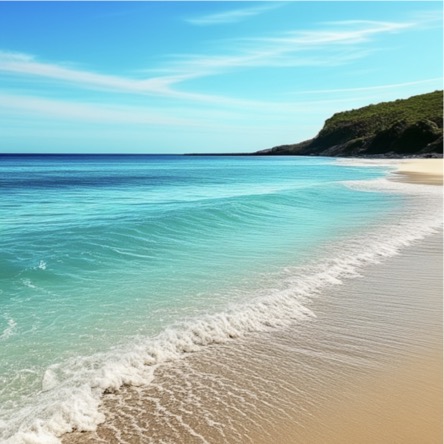} &
            \includegraphics[width=0.15\textwidth]{fig/stable_cascade/ocean_green_30.0.jpg} &
            \includegraphics[width=0.15\textwidth]{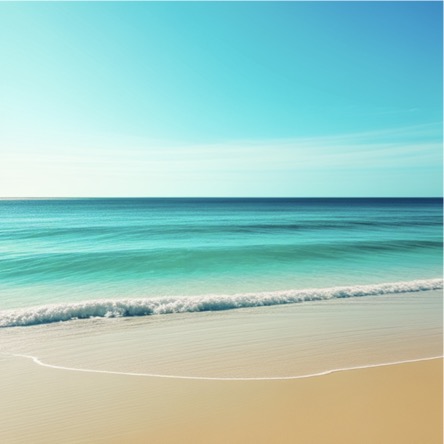} &
            \includegraphics[width=0.15\textwidth]{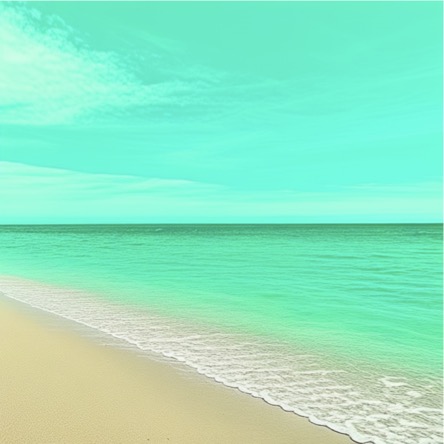} &
            \includegraphics[width=0.15\textwidth]{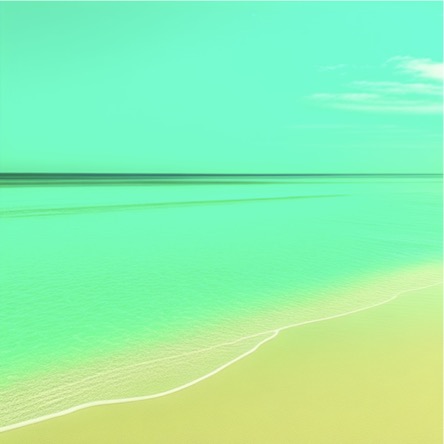} \\
        \end{tabular} \\
        
        \textit{$w_b(\mathbf{x} ; \xi)$} \\
        \begin{tabular}{cccccc}
            \includegraphics[width=0.15\textwidth]{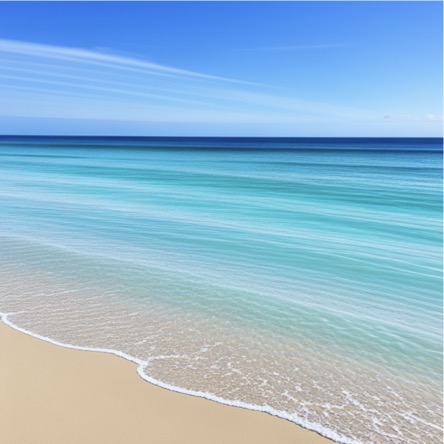} &
            \includegraphics[width=0.15\textwidth]{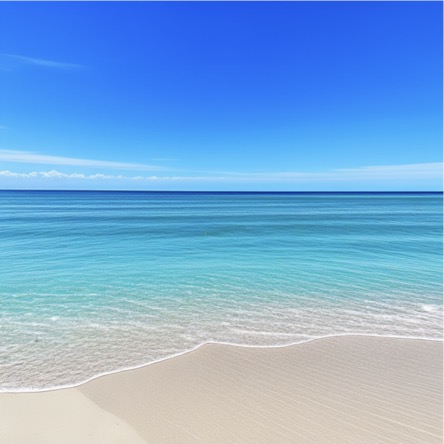} &
            \includegraphics[width=0.15\textwidth]{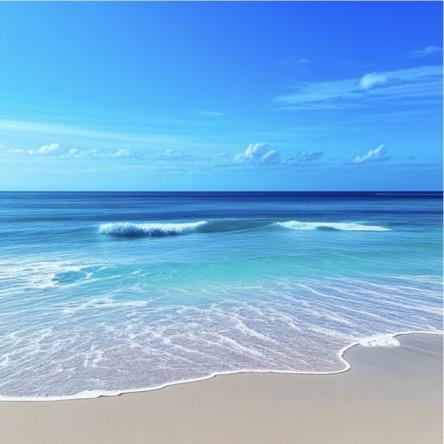} &
            \includegraphics[width=0.15\textwidth]{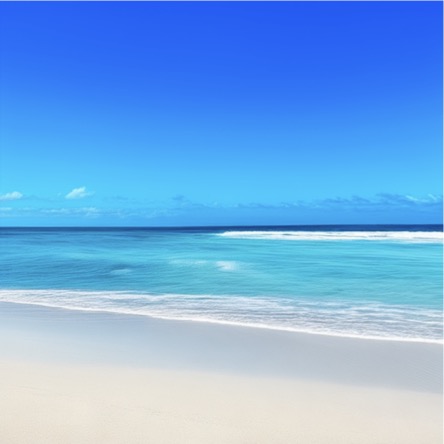} &
            \includegraphics[width=0.15\textwidth]{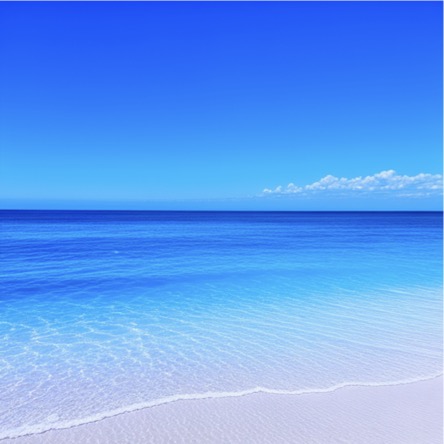} &
            \includegraphics[width=0.15\textwidth]{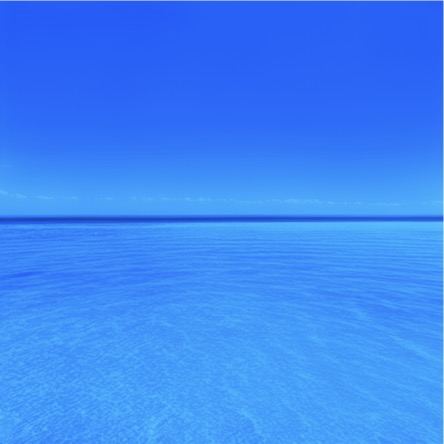} \\
        \end{tabular} \\
    \end{tabular}

    \vspace{0.5em}
    \begin{minipage}{0.9\textwidth}
        \centering
        \textbf{Low } $\xi$: No color bias \(\longleftarrow\) \hspace{3cm} \(\longrightarrow\) \textbf{High } $\xi$: Emphasis on color bias
    \end{minipage}

    \caption{
        Images generated using the prompt \emph{``An ocean with a beach''} for each RGB color channel. The rows correspond to the Red, Green, and Blue channels, respectively, while the columns represent increasing values of the parameter $\xi = (10, 20, 30, 50, 100, 150)$ from left to right. Lower $\xi$ values produce typical colors, whereas higher $\xi$ values enhance the intensity of the corresponding color channel. Notably, this control is achieved solely through weighted sampling, \emph{\textbf{without} use of any conditional prompts related to the channels or colors}.
    }
    \label{fig:ocean_images}
\end{figure*}

In Figure~\ref{fig:ocean_images}, we present weighted sampling results, where the top row corresponds to $w_{r}$, the second row to $w_{g}$, and the bottom row to $w_{b}$. The weight function is applied with varying values of $\xi \in \{10, 20, 30, 50, 100, 150\}$ from left to right, enabling controlled emphasis on each color channel.

By adjusting $\xi$, we observe that the generated images progressively shift toward the target color channel while maintaining a fixed text prompt, \textit{``An ocean with a beach.''}, \textbf{\emph{without any color-related text conditioning}}. At low $\xi$ values, the images retain natural colors, whereas higher values progressively enhance the target channel, producing visually distinct outputs. This demonstrates the effectiveness of our method in biasing the generated distribution based on externally defined weight functions, independent of textual context.

\subsubsection{Sampling High-Spatial-Frequency Images from
Foundation Diffusion Models}
\label{freq_emphasis}

\paragraph{Weight function.}
To control the frequency content of generated images for each RGB channel, we leverage the 2D Fourier transform to separate high- and low-frequency components. Starting with an image $\mathbf{x} \in \mathbb{R}^{3 \times h \times w}$, its three color channels are processed separately. The frequency representations for each channel are computed as
\[
\mathbf{f}_r(\rvx) \coloneqq \mathcal{F}([\mathbf{x}]_{1, i, j}), \quad 
\mathbf{f}_g(\rvx) \coloneqq \mathcal{F}([\mathbf{x}]_{2, i, j}), \quad 
\mathbf{f}_b(\rvx) \coloneqq \mathcal{F}([\mathbf{x}]_{3, i, j})
\]
where $\mathcal{F}(\cdot)$ represents the 2D Fast Fourier Transform (FFT), and the output is FFT-shifted to center low frequencies. To differentiate between low- and high-frequency components, we define a distance matrix $D(i, j)$ as
\[
D(i, j) \coloneqq \sqrt{(i - h/2)^2 + (j - w/2)^2}
\]
where $D(i, j)$ measures the distance of each frequency component from the center of the FFT-shifted representation. This matrix is shared across all channels. Using this distance matrix, masks are constructed to emphasize high- or low-frequency components with a smooth transition. Specifically, the high-frequency mask is defined as
\[
\mathbf{m}_{\text{high}}(i, j) \coloneqq \frac{1}{1 + \exp\left(-\frac{D(i, j) - r}{s}\right)}
\]
where $r$ is the cutoff radius (e.g., $r = \frac{\min(h, w)}{4}$) and $s$ controls the smoothness of the transition. This mask assigns higher values to frequency components farther from the center, emphasizing high frequencies. Similarly, the low-frequency mask is defined as
\[
\mathbf{m}_{\text{low}}(i, j) \coloneqq \frac{1}{1 + \exp\left(\frac{D(i, j) - r}{s}\right)}
\]
which emphasizes components closer to the center (low frequencies).
To quantify the emphasis on specific frequency ranges, we compute the relative strength of high- versus low-frequency components by directly applying the selected mask to the sum of the frequency representations of each channel using element-wise multiplication ($\odot$):

\[
w(\mathbf{x}; \xi) = \exp\left( \xi \cdot \frac{\sum_{i,j} \left| \left( \mathbf{f}_r(\mathbf{x}) + \mathbf{f}_g(\mathbf{x}) + \mathbf{f}_b(\mathbf{x}) \right) \odot \mathbf{m}_{\text{high}} \right|_{i,j}}{\sum_{i,j} \left| \left( \mathbf{f}_r(\mathbf{x}) + \mathbf{f}_g(\mathbf{x}) + \mathbf{f}_b(\mathbf{x}) \right) \odot \mathbf{m}_{\text{low}} \right|_{i,j}} \right).
\]
where \( \xi \) controls the bias. For \( \xi > 0 \), high-frequency components are emphasized, enhancing textures and details. For \( \xi = 0 \), there is neutral sampling without frequency emphasis. For \( \xi < 0 \), low-frequency components are emphasized, enhancing smoothness and gradients. This unified weight function \( w(\mathbf{x}; \xi) \) allows a single formulation for both high- and low-frequency emphasis based on the value of \(\xi\).

\begin{figure*}[htbp]
    \centering
    \begin{tabular}{ccc}
        \begin{tabular}{c}
            \includegraphics[width=0.27\textwidth]{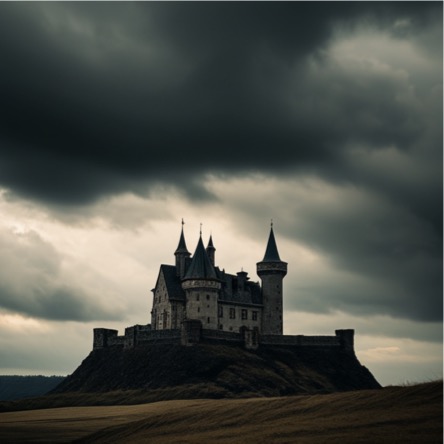} \\
            \includegraphics[width=0.27\textwidth]{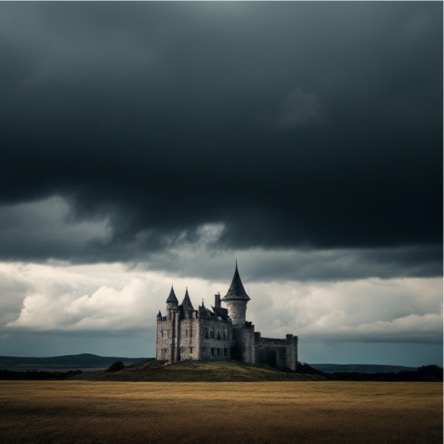} \\
            \includegraphics[width=0.27\textwidth]{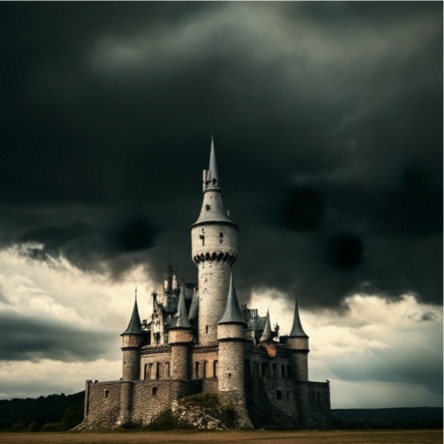} \\
            \(\leftarrow\) \\
            \textit{$w(\cdot ; \xi = -30.0)$}
        \end{tabular}
        &
        \begin{tabular}{c}
            \includegraphics[width=0.27\textwidth]{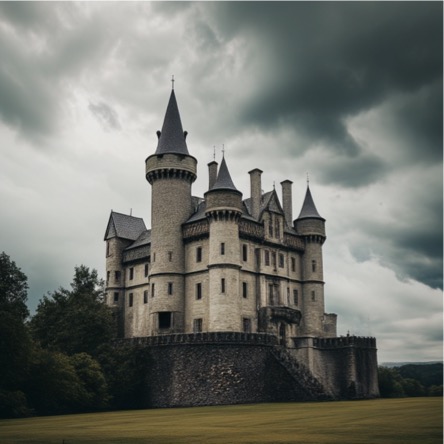} \\
            \includegraphics[width=0.27\textwidth]{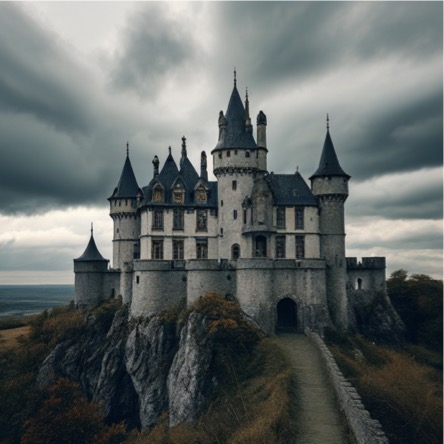} \\
            \includegraphics[width=0.27\textwidth]{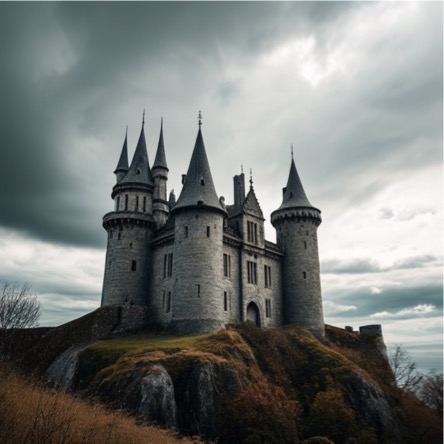} \\
            \\
            \textit{$w(\cdot ; \xi = 0.0)$}
        \end{tabular}
        &
        \begin{tabular}{c}
            \includegraphics[width=0.27\textwidth]{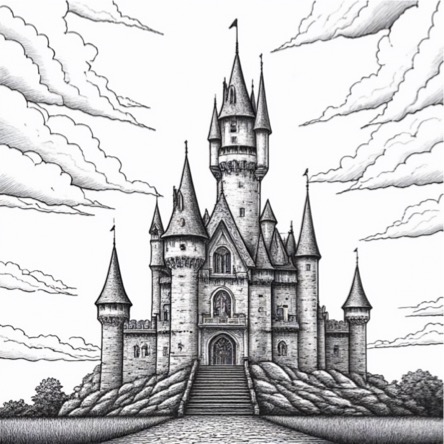} \\
            \includegraphics[width=0.27\textwidth]{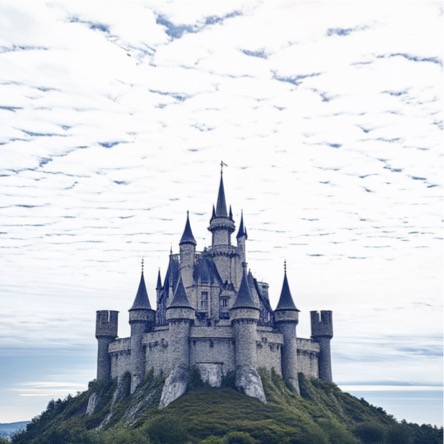} \\
            \includegraphics[width=0.27\textwidth]{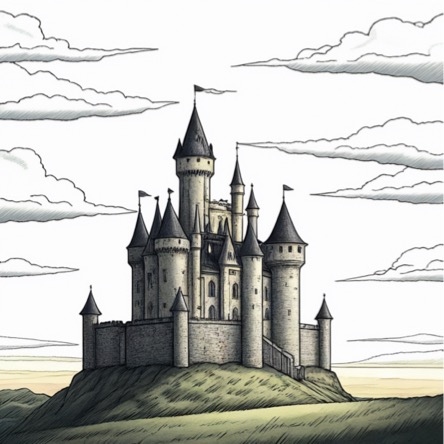} \\
            \(\rightarrow\) \\
            \textit{$w(\cdot ; \xi = 7.0)$}
        \end{tabular}
    \end{tabular}
    \caption{ \textbf{\emph{Left}:} Images generated with the prompt \emph{``A castle under a cloudy sky,''} emphasizing low-frequency details (\(w(\cdot ; \xi = -30.0)\)). These images exhibit smoother gradients and softer transitions, highlighting subtle atmospheric effects and giving a calm, natural feel. \textbf{\emph{Center}:} Neutral images without frequency bias (\(w(\cdot ; \xi = 0.0)\)), showcasing the original generative capabilities of the model with diverse interpretations of the prompt. \textbf{\emph{Right}:} Images generated with the same prompt, emphasizing high-frequency details (\(w(\cdot ; \xi = 7.0)\)). These images display sharper textures and vivid contrasts, producing a slightly stylized, cartoon-like appearance.
        This comparison demonstrates how adjusting the emphasis on high or low frequencies can yield a wide variety of results \textbf{\emph{without}} prompt condition related to the feature.
    }
    \label{fig:castle_frequency_bias}
\end{figure*}

\begin{figure*}[htbp]
    \centering
    \makebox[0.95\textwidth][c]{ 
    \begin{tabular}{ccc}
        \includegraphics[width=0.25\textwidth]{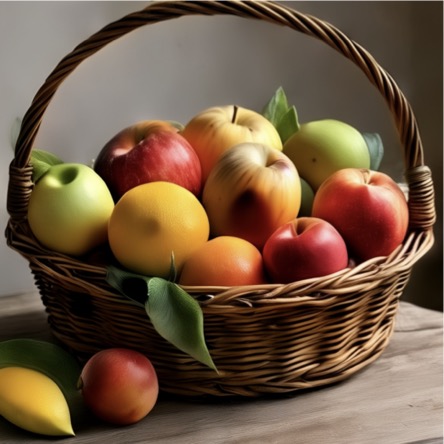} &
        \includegraphics[width=0.25\textwidth]{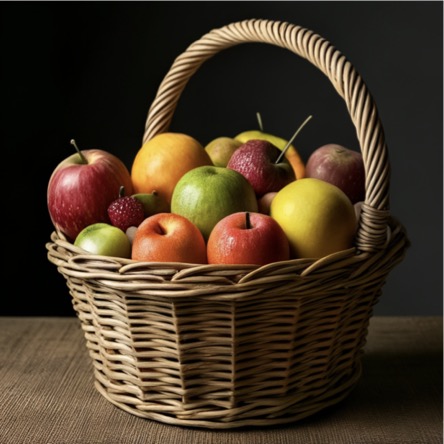} &
        \includegraphics[width=0.25\textwidth]{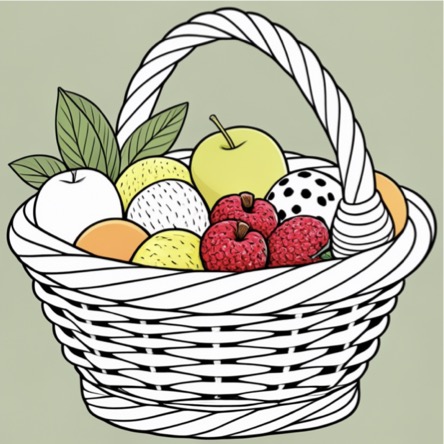} \\

        \includegraphics[width=0.25\textwidth]{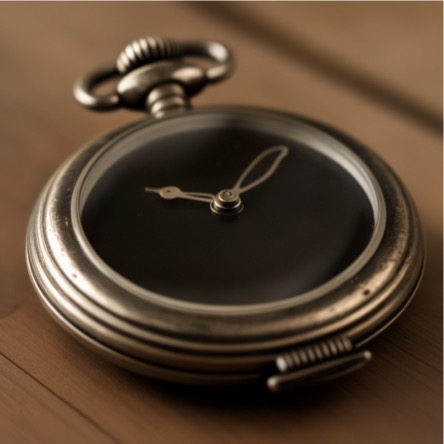} &
        \includegraphics[width=0.25\textwidth]{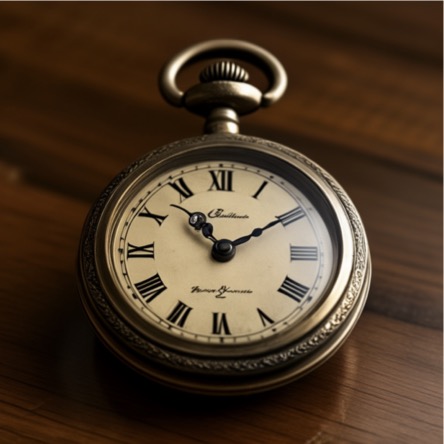} &
        \includegraphics[width=0.25\textwidth]{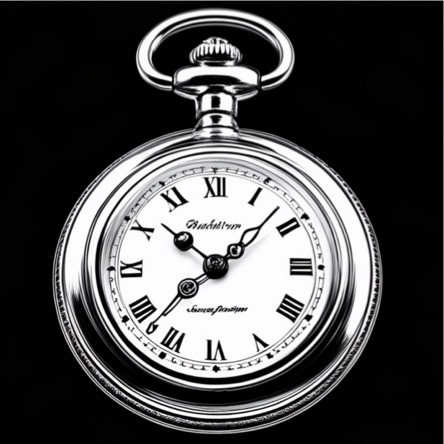} \\

        \includegraphics[width=0.25\textwidth]{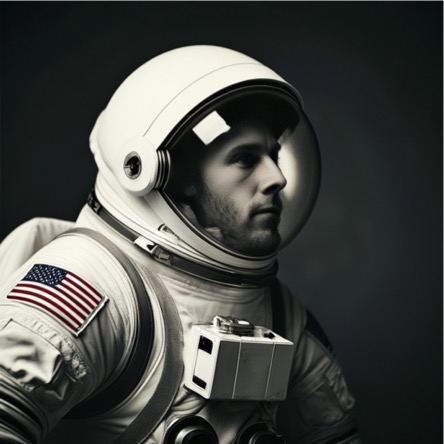} &
        \includegraphics[width=0.25\textwidth]{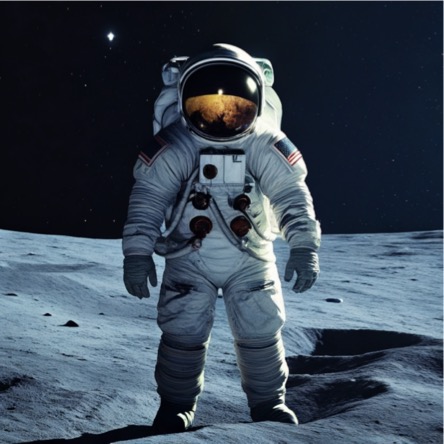} &
        \includegraphics[width=0.25\textwidth]{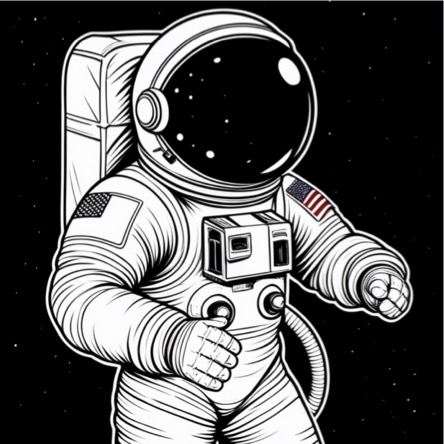} \\
    \end{tabular}
    } 
    \vspace{0.5em}
    \begin{minipage}{0.9\textwidth}
        \centering
 \(\longleftarrow\) \hspace{8.0cm} \(\longrightarrow\) \\
        \textit{$w(\cdot ; \xi = -30.0)$} \hspace{1.4cm} \textit{$w(\cdot ; \xi = 0.0)$} \hspace{2.0cm}
        \textit{$w(\cdot ; \xi = 7.0)$}
    \end{minipage}
    \caption{
        \textbf{\emph{Left}:} Images generated with low-frequency emphasis (\( \xi = -30.0 \)). \textbf{\emph{Center}:} Neutral images without frequency bias (\( \xi = 0.0 \)). \textbf{\emph{Right}:} Images generated with high-frequency emphasis (\( \xi = 7.0 \)). Each row corresponds to a different prompt: \emph{``A woven basket filled with fruit,''} \emph{``An old-fashioned pocket watch,''} and \emph{``An astronaut,''} respectively. 
    }
    \label{fig:object_frequency_bias}
\end{figure*}

\paragraph{Results.} 

Figure~\ref{fig:castle_frequency_bias} presents images generated with different random seeds for each $\xi \in \{-30.0, 0.0, 7.0\}$, corresponding to the left, middle, and right columns, respectively. The prompt \emph{``A castle under a cloudy sky''} was used, which does not contain explicit stylistic descriptors.

For $\xi < 0$, where images with dominant low-frequency components receive higher importance, the generated images exhibit more prominent cloud structures while the castle appears smaller. This observation is consistent with the fact that cloud formations typically contain gradual pixel variations, which contribute predominantly to low-frequency components. Conversely, for $\xi > 0$, where high-frequency components are emphasized, the images display sharper edges, more pronounced outlines, and increased contrast with simplified color transitions.

Figure~\ref{fig:object_frequency_bias} further illustrates the impact of weighted sampling using different text prompts: \emph{``A woven basket filled with fruit''} (top row), \emph{``An old-fashioned pocket watch''} (middle row), and \emph{``An astronaut''} (bottom row). For each prompt, weighted sampling is performed with $\xi \in \{-30.0, 0.0, 7.0\}$, corresponding to the left, middle, and right columns, respectively. Higher values of $\xi$ assign greater importance to samples with dominant high-frequency components.

A similar trend is observed in Figure~\ref{fig:object_frequency_bias}, where images generated with low-frequency emphasis exhibit blurred or simplified details. For instance, in the case of the pocket watch, low-frequency emphasis results in an image without visible tick marks, reflecting the predominance of smooth, low-frequency components. In contrast, high-frequency emphasis enhances fine details, producing images with well-defined edges and a pen-illustration-like appearance.

\begin{figure*}[htbp]
    \centering
        \includegraphics[width=0.9\textwidth]{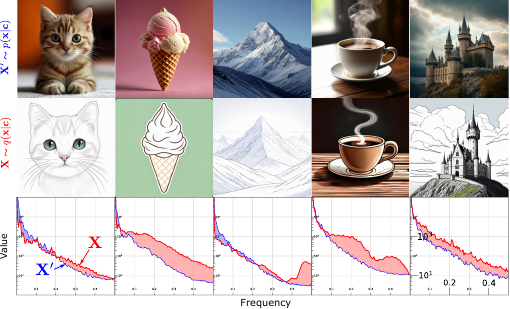} 
    \caption{Comparison of generated samples: (top) images synthesized by the original diffusion model, (middle) images obtained via our weighted sampling process, and (bottom) corresponding frequency component analysis.}
    \label{fig:merged_stable_cascade_for_appendix}
\end{figure*}

In Figure \ref{fig:merged_stable_cascade_for_appendix}, the top row shows five randomly generated samples from the original diffusion backward process using the prompts: \emph{``An adorable cat,'' ``An ice cream cone,'' ``A snowy mountain,'' ``A steaming cup of coffee,''} and \emph{``A castle under a cloudy sky.''} The middle row illustrates results produced by our weighted sampling method, which prioritizes instances with higher high-frequency components, following the same methodology as employed in the preceding examples in this section. The bottom row presents the frequency component analysis obtained via the Fourier transform.

The results indicate that our weighted sampling method effectively enhances high-frequency components, leading to the generation of images with more pronounced edges, often resembling a pen illustration style. The frequency component plots in the bottom row depict the red-shaded regions, which represent the increased magnitude of high-frequency components induced by our weighted sampling approach. As evident across all samples, our method consistently amplifies these components, demonstrating that the weighted sampling technique directs the foundational diffusion model to generate samples enriched with high-frequency details through an externally defined importance weighting function.

Crucially, the images produced via our weighted sampling method exhibit stylistic modifications {without any explicit textual conditioning related to the appearance or characteristics of the generated images}. The transformation is achieved purely through the externally defined frequency emphasis function, while maintaining the exact same textual prompts as those used in the baseline sample generation. 
As such, the proposed approach opens the door to incorporating more sophisticated, task-specific importance functions, enabling controllable and flexible guidance without the need for prompt engineering or additional model training.

\end{document}